\PassOptionsToPackage{numbers}{natbib}
\PassOptionsToPackage{sort}{natbib}

\documentclass{article}

\usepackage{geometry}
\usepackage[utf8]{inputenc} 
\usepackage[T1]{fontenc}   
\usepackage{fullpage}
\usepackage[numbers]{natbib}
\usepackage{floatrow}

\usepackage{blindtext}

\date{}

\usepackage{array}
\usepackage{arydshln}
\usepackage{amsmath}
\usepackage{amssymb}
\usepackage{mathtools}
\usepackage{amsthm}

\usepackage{hyperref} 
\usepackage{cleveref}
\newcommand{\comment}[1]{}
\usepackage[T1]{fontenc}    
\usepackage{url}            
\usepackage{booktabs}       
\usepackage{amsfonts}       
\usepackage{nicefrac}       
\usepackage{microtype}      
\usepackage{algorithm}
\usepackage{algorithmic}
\usepackage{algorithmic}
\usepackage{algorithmic}
\usepackage{algorithmic}
\usepackage{lipsum}
\usepackage{enumitem}
\usepackage{amsmath, amssymb, amsthm}
\usepackage{wrapfig}
\crefname{equation}{}{}
\Crefname{equation}{}{}

\usepackage[utf8]{inputenc}
\usepackage{mathtools}
\usepackage{subcaption}
\usepackage{graphicx}
\usepackage[font=small,labelfont=bf]{caption}
\usepackage{bbm}
\usepackage{array}
\usepackage{arydshln}

\usepackage{multirow}

\usepackage{url}            
\usepackage{booktabs}       
\usepackage{amsfonts}       
\usepackage{nicefrac}       
\usepackage{xcolor}
\usepackage{microtype}      
\usepackage{amssymb,amsmath, amsthm}
\usepackage[utf8]{inputenc}
\usepackage{mathtools}
\usepackage{subcaption}
\usepackage{tikz}
\usetikzlibrary{calc}
\usepackage{graphicx}
\usepackage{caption}
\usepackage{bbm}
\usepackage{bm}
\usepackage{array}
\usepackage{arydshln}
\usepackage{xspace}

\newtheorem{theorem}{Theorem}[section]
\newtheorem{lemma}{Lemma}[section]
\newtheorem{assumption}{Assumption}[section]

\usepackage{algorithmic}
\usepackage[colorinlistoftodos,prependcaption]{todonotes}

\newcommand{\DiJ}[1]{\textcolor{orange}{DJ: #1}}

\newcommand{\YC}[1]{\todo[color=blue!5, inline]{ Yae Jee: #1} \index{Yae Jee: !#1}}

\newcommand\yj[1]{{\color{blue}\textbf{YJ: }{#1} }}

\definecolor{gr}{rgb}{0.25, 0.25, 0.25}

\newcommand{\sigm}{\sigma}
\newcommand{\sgn}{\sigma_g}

\newcommand{\ec}{M}

\newcommand{\ic}{k}

\newcommand{\incfl}{\textsc{MaxFL}\xspace}
\newcommand{\cnst}{\rho}

\newcommand{\gf}{F}

\newcommand{\sgf}{\widetilde{\gf}}
\newcommand{\vpr}{\textsc{GM-Appeal}\xspace}
\newcommand{\lff}{f}
\newcommand{\lf}{F_\ic}

\newcommand{\sett}{\mathcal{S}^{(t,0)}}
\newcommand{\gb}{\mathbf{g}}

\newcommand{\hb}{\mathbf{h}}
\newcommand{\hbl}{\overline{\mathbf{h}}}

\newcommand{\wb}{\mathbf{w}}

\newcommand{\bdat}{\mathcal{B}_\ic}

\newcommand{\defeq}{\vcentcolon=}

\newcommand{\ti}{(t,0)}

\newcommand{\sgrad}{\mathbf{g}_\ic}

\newcommand{\lr}{\eta}
\newcommand{\lrl}{\eta_l}
\newcommand{\lrg}{\eta_g}
\newcommand{\lrt}{\widetilde{\eta}}

\newcommand{\expt}{\mathbb{E}}

\newcolumntype{?}{!{\vrule width 1pt}}

\newcommand{\dtoprule}{\specialrule{1pt}{0pt}{0.7pt}%
            \specialrule{0.3pt}{0pt}{\belowrulesep}%
            }
\newcommand{\dbottomrule}{\specialrule{0.3pt}{0pt}{0.7pt}%
            \specialrule{1pt}{0pt}{0.1pt}%
            }
\DeclareMathOperator*{\argmin}{argmin} 
\DeclareMathOperator*{\argmax}{argmax} 

\setlist[enumerate]{label=\roman*),
                    leftmargin=2em, 
                    }

\theoremstyle{plain}

\newtheorem*{thm*}{Theorem}
\newtheorem{defn}{Definition}

\newcommand{\E}[1]{\mathbb{E}\left[{#1}\right]}
\newcommand{\Prob}[1]{\mathbb{P}\left({#1}\right)}

\newcommand{\brac}[1]{\left({#1}\right)}

\graphicspath{{Figures/}}

\crefname{equation}{}{}
\Crefname{equation}{}{}
\crefname{thm}{theorem}{theorems}
\Crefname{thm}{Theorem}{Theorems}
\crefname{clm}{claim}{claims}
\Crefname{clm}{Claim}{Claims}
\Crefname{coro}{Corollary}{Corollaries}
\Crefname{lem}{Lemma}{Lemmas}
\Crefname{sec}{Section}{Sections}
\crefname{app}{appendix}{appendices}
\Crefname{app}{Appendix}{Appendices}
\crefname{prop}{proposition}{propositions}
\Crefname{prop}{Proposition}{Propositions}
\Crefname{propty}{Property}{Properties}
\crefname{figure}{fig.}{figures}
\Crefname{figure}{Fig.}{Figures}
\crefname{defn}{definition}{definitions}
\Crefname{defn}{Definition}{Definitions}
\crefname{fact}{fact}{facts}
\Crefname{fact}{Fact}{Facts}
\crefname{appendix}{appendix}{appendices}
\Crefname{appendix}{Appendix}{Appendices}
\crefname{algo}{algorithm}{algorithms}
\Crefname{algo}{Algorithm}{Algorithms}
\crefname{algorithm}{algorithm}{algorithms}
\Crefname{algorithm}{Algorithm}{Algorithms}
\crefname{tbl}{table}{table}
\Crefname{tbl}{Table}{Table}
\crefname{table}{table}{table}
\Crefname{table}{Table}{Table}
\crefname{algorithm}{algorithm}{algorithms}
\Crefname{algorithm}{Algorithm}{Algorithms}

\crefname{conj}{conjecture}{conjectures}
\Crefname{conj}{Conjecture}{Conjectures}
\crefname{obs}{observation}{observations}
\Crefname{obs}{Observation}{Observations}

\newcommand{\hsg}{\widehat{\gamma}_G}

\newcommand{\htheta}{\widehat{\theta}}

\newcommand{\bx}{{\bf x}}

\newcommand{\by}{{\bf y}}

\newcommand{\bw}{{\bf w}}

\newcommand{\set}{{\mathcal{S}}}

\title{\textbf{Maximizing Global Model Appeal in Federated Learning}}

\author{
Yae Jee Cho \\
\small Carnegie Mellon University\\
\small \texttt{\href{mailto:yaejeec@andrew.cmu.edu}{yaejeec@andrew.cmu.edu}} \\
\and
Divyansh Jhunjhunwala \\
\small Carnegie Mellon University\\
\small \texttt{\href{mailto:djhunjhu@andrew.cmu.edu}{djhunjhu@andrew.cmu.edu}} \\
\and
Tian Li \\
\small Carnegie Mellon University\\
\small \texttt{\href{mailto:tianli@cmu.edu}{tianli@cmu.edu}} \\
\and
Virginia Smith \\
\small Carnegie Mellon University\\
\small \texttt{\href{mailto:smithv@cmu.edu}{smithv@cmu.edu}} \\
\and
Gauri Joshi \\
\small Carnegie Mellon University\\
\small \texttt{\href{mailto:gaurij@andrew.cmu.edu}{gaurij@andrew.cmu.edu}}
}

\begin{document}
\maketitle
%

%


\begin{abstract}

Federated learning typically considers collaboratively training a global model using local data at edge clients. Clients may have their own individual requirements, such as having a minimal training loss threshold, which they expect to be met by the global model. However, due to client heterogeneity, the global model may not meet each client's requirements, and only a small subset may find the global model {appealing}. 
In this work, we explore the problem of the  global model lacking \textit{appeal} to the clients due to not being able to satisfy local requirements. 
We propose \incfl, which aims to maximize the number of clients that find the global model appealing. We show that having a high global model appeal is important to maintain an adequate pool of clients for training, and can directly improve the test accuracy on both seen and unseen clients. We provide convergence guarantees for \incfl and show that \incfl achieves a $22$-$40\%$ and $18$-$50\%$ test accuracy improvement for the training clients and unseen clients respectively, compared to a wide range of FL modeling approaches, including those that tackle data heterogeneity, aim to incentivize clients, and learn personalized or fair models. 
\end{abstract}

\vspace{-0.3em}
\section{Introduction} \label{sec:intro}
\vspace{-0.3em}

Federated learning (FL) is a distributed learning framework that considers training a machine learning model using a network of clients (e.g., mobile phones, hospitals), without directly sharing client data with a central server~\cite{mcmahan2017communication}. FL is typically performed by aggregating clients' updates over multiple communication rounds to produce a global model~\cite{kairouz2019advances}.
In turn, each client may have its own requirement that it expects to be met by the resulting global model. 
For example, clients such as hospitals or edge-devices may expect that the global model \textit{at least} performs better than a local model trained in isolation on the client's limited local data before contributing to FL training. Unforunately, due to heterogeneity across the clients, the global model may fail to meet all the clients' requirements~\cite{yu2020salvaging}. 

\begin{wrapfigure}[17]{!r}{.5\textwidth}\vspace{-1.5em}
{\includegraphics[width=1\textwidth]{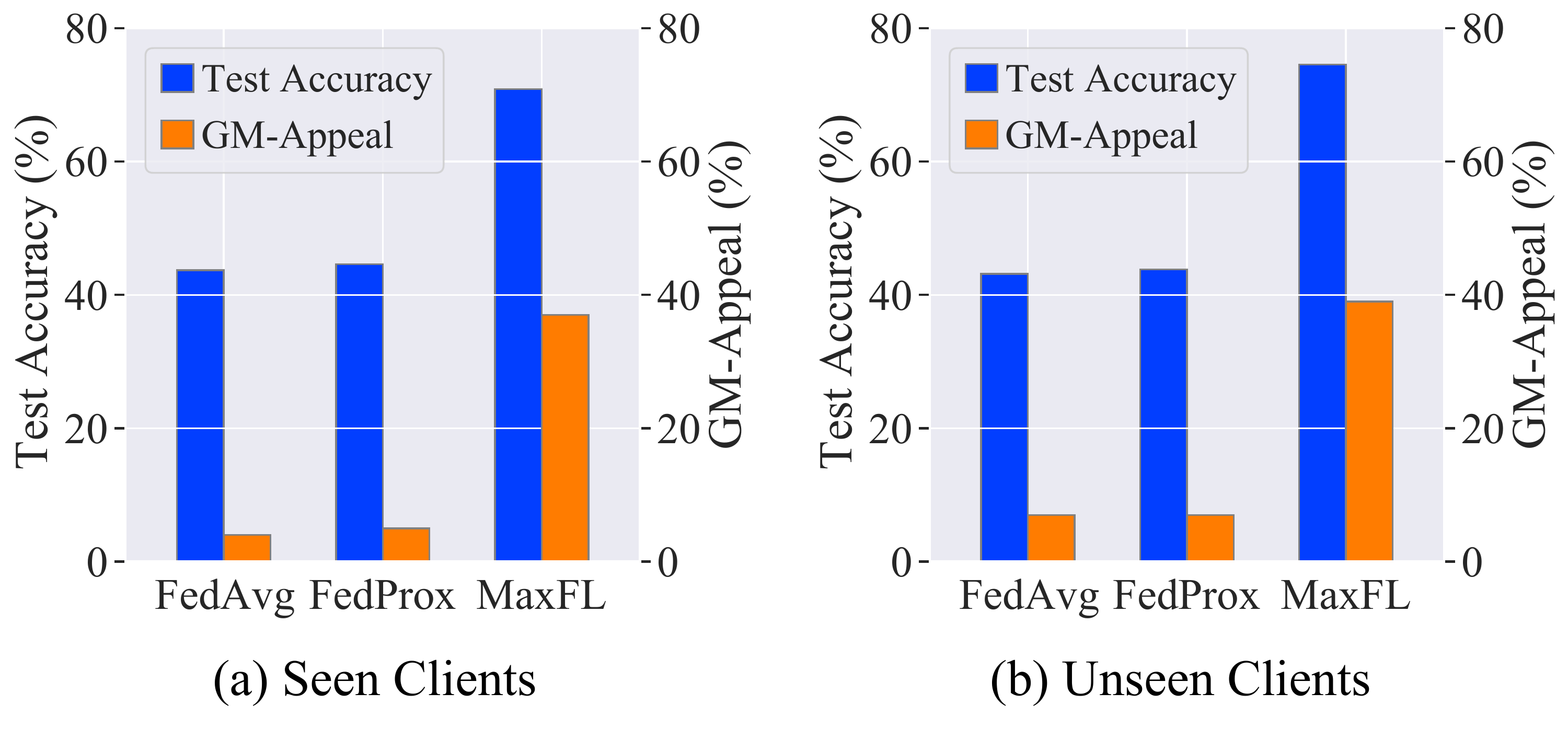}}
 \caption{Test accuracy and global model appeal (GM-Appeal) of the global model for FMNIST dataset with the seen clients that have participated during training and unseen clients that have not (more details in \Cref{tab:testacc}). A higher GM-Appeal results in a higher test accuracy for both types of clients due to the server having a larger pool of clients to select from. \incfl, which aims to maximize GM-Appeal, results in the highest test accuracy compared to the other baselines that do not consider GM-Appeal.} \label{fig:moti}   
\end{wrapfigure}

In this work, we say that {a global model is \textit{appealing} to a client} if it satisfies the client's specified requirement, such as incurring at most some max training loss. Subsequently, we define the number of clients which find the global model appealing as \textit{global model appeal} (formally defined in \Cref{def1}). We find that having a high global model appeal is critical to maintain a large pool of clients to select from for training, and for gathering additional willingly participating clients. This is especially true in the light of clients possibly opting out of FL due to the significant costs associated with training (e.g., computational overhead, privacy risks, logistical challenges). With a larger pool of clients to select from, i.e., with a higher global model appeal, a server can not only improve privacy-utility trade-offs~\cite{mcmahan2017learning}, but can also improve the test accuracy of the seen clients, and produce a global model that generalizes better at inference to new unseen clients (see \Cref{fig:moti} and \Cref{tab:testacc}).

In this work, we seek to understand: (1) What strategies exist to maximize global model appeal in federated settings, and (2) What  benefits exist when maximizing this notion relative to other common federated modeling approaches.
Our key contributions are summarized as follows: \vspace{-0.5em}
\begin{itemize}[leftmargin=*]
    \item We introduce the notion of global model appeal (referred to as \vpr), which is the fraction of clients that have their local requirements met by the global model. \vspace{-0.5em}
    \item We show that having a high global model appeal is imperative for better test accuracy on seen clients as well as better generalization to new unseen clients. \vspace{-0.5em}
   \item We propose \incfl, a novel framework that maximizes global model appeal via a client requirement-aware weighted aggregation of client updates.  \vspace{-0.5em}
\end{itemize}

\begin{itemize}[leftmargin=*] \vspace{-1.7em}
   \item We provide convergence guarantees for our \incfl solver which allows partial client participation, is applicable to non-convex objectives, and is stateless (does not require clients maintain local parameters during training). 
    \item We empirically validate the performance of \incfl with experiments where i) clients can flexibly opt-out of training, ii) there are new incoming (unseen) clients, and iii) there are Byzantine clients participating in training. 
    \item We show that \incfl improves the global model appeal greatly and thus achieves a $22$-$40\%$ and $18$-$50\%$ test accuracy improvement for the seen clients and unseen clients respectively, compared to a wide range of FL methods, including those that tackle data heterogeneity, aim for variance reduction or incentivizing clients, or provide personalization or fairness. 
\end{itemize} 
 \vspace{-0.5em}

To the best of our knowledge, the notion of global model appeal has not been explored previously in FL. Our work is the first to consider such a notion,  show its importance in FL, and then propose an objective to train a global model that can maximize the number of clients whose individual requirements are satisfied. We provide a more detailed review of prior work and related areas of fairness, personalization and client incentives in \Cref{sec:related_work}.
\vspace{-0.3em}
\section{Problem Formulation}\vspace{-0.3em}
\label{sec:pf}

We consider a setup where $M$ clients are connected to a central server to collaboratively train a global model. For each client $\ic\in [M]$, its true loss function is given by $f_\ic(\wb)=\expt_{\xi\sim\mathcal{D}_\ic}[\ell(\wb,\xi)]$ where $\mathcal{D}_\ic$ is the true data distribution of client $\ic$, and $\ell(\wb, \xi)$ is the composite loss function for the model $\wb\in\mathbb{R}^{d}$ for data sample $\xi$. In practice, each client only has access to its local training dataset $\bdat$ with $|\bdat|=N_k$ data samples sampled from $\mathcal{D}_\ic$. Client $k$'s empirical loss function is $F_\ic(\wb)=\frac{1}{|\bdat|}\sum_{\xi \in \bdat} \ell(\wb,\xi)$. While some of the take-aways of our work (e.g., improving performance on unseen clients) are more specific to cross-device applications, our general setup and method are applicable to both cross-device and cross-silo FL.



\textbf{Defining Global Model Appeal.} Each client's natural aim is to find a model that minimizes its true local loss $f_k(\wb)$. Clients can have different thresholds of how small this loss should be, and we denote such self-defined threshold for each client as $\cnst_k,~k\in[M]$. For instance, each client can perform solo training on its local dataset $\mathcal{B}_k$ to obtain an approximate local model $\widehat{\wb}_\ic$ and have its threshold to be the true loss from this local model, i.e., $\cnst_k=f_k(\widehat{\wb}_\ic)$\footnote{In practice, the client may have held-out data used for calculating the true loss $f_k(\cdot)$ or use its training data as a proxy.}. 
Based on these client requirements, we provide the formal definition of global model appeal below: 

\begin{defn}[Global Model Appeal]
A global model $\wb$ is said to be appealing to client $k\in[M]$ if $f_k(\wb) < \rho_k$, i.e., the global model $\wb$ yields a smaller local true loss than the self-defined threshold of the client. Global model appeal (GM-appeal) is then defined as the fraction of clients to which the global model is appealing: \vspace{-0.5em}
\begin{align}
\vpr=\frac{1}{M}\sum_{k=1}^M\mathbb{I}\{f_k(\wb)<\rho_k\} \label{eq:gmas}
\end{align}
where $\mathbb{I}$ is the indicator function.
\label{def1}
\end{defn}
Our \vpr metric measures the fraction of clients that find the global model appealing. Note that \vpr only looks at whether the global model satisfies the clients' requirements or not instead of looking at the gap between $f_k(\wb)$ and $\rho_k$. 
Another variation of \Cref{eq:gmas} could be to measure the margin $\sum_k\max\{\rho_k-f_k(\wb), 0\}$, but this does not capture the motivation behind our work which is for the server to maximize the number of clients that find the global model appealing. To the best of our knowledge, a similar indicator-based metric has not been explored previously in the FL literature.


\subsection{Why \vpr is Important in FL}
Before presenting our proposed objective \incfl that maximizes GM-Appeal, we first elaborate on the significance of the \vpr metric in FL, including how it affects the test performance of the training (seen) clients and the generalization performance on new (unseen) clients at inference. 

\vpr measures how many clients' requirements are satisfied by the global model. Thus, it gauges important characteristics of the global model such as how many clients are likely to dropout with the current global model or how many new incoming clients will likely be satisfied with the current global model. Ultimately, a high global model appeal leads to a larger pool of clients for the server to select from. The standard FL objective~\cite{mcmahan2017communication} does not consider whether the global model satisfies the clients' requirements, and implicitly assumes that the server will have a large number of clients to select from. However, this may not necessarily be true if clients are allowed to dropout when they find the global model unappealing. 

Acquiring a larger pool of clients by improving global model appeal is imperative to improve test accuracy performance of the global model to the seen clients as well as for improving the generalization performance of the global model on the unseen clients at inference (see \Cref{fig:moti} and \Cref{tab:testacc}). In fact, we find that  other baselines such as those that aim to tackle data heterogeneity, improve fairness, or provide personalization all have low GM-Appeal, leading to a large number of clients opting out. Due to this, the global model is trained on just a few limited data points, resulting in poor performance. On the other hand, our proposed \incfl which aims to maximize \vpr is able to retain a large number of clients for training, resulting in a better global model. Moreover, we show that a high GM-appeal for the current set of training (seen) clients also leads to having a high GM-appeal for the unseen clients and also better test accuracy on these new unseen clients at inference. 

\subsection{Proposed \incfl Objective}\vspace{-0.3em} \label{sec:proposed}
In the previous section, we have shown that having a high global model appeal is imperative to achieve both good test accuracy and generalization performance. In this section, we introduce \incfl~whose aim is to train a global model that maximizes \vpr. A na\"ive approach is to find the global model that can directly maximize \vpr defined in \Cref{eq:gmas} as follows:
\begin{flalign}
\begin{aligned}
  \argmax_{\wb} \vpr 
  = \argmin_{\wb}\sum_{k=1}^M\text{sign}(f_k(\wb)-\rho_k).
  \label{eq:max_min_vpr}
\end{aligned}
\end{flalign}
where $\text{sign}(x) = 1$ if $x\geq 0$ and $0$ otherwise. There are two immediate difficulties in minimizing \cref{eq:max_min_vpr}. First, clients may not know their true data distribution $\mathcal{D}_k$ to compute $f_k(\wb)-\rho_k$. Second, the sign function makes the objective nondifferentiable and limits the use of common gradient-based methods. We resolve these issues by proposing a "proxy" for \cref{eq:max_min_vpr} with the following relaxations. \\
    
\textbf{Replacing the Sign function with the Sigmoid function $\sigma(\cdot)$:} Replacing the non-differentiable 0-1 loss with a smooth differentiable loss is a standard tool used in optimization~\cite{tan201301loss,hamed200801loss}. Given the many candidates (e.g. hinge loss, ReLU, sigmoid), 
    we find that using the sigmoid function 
    is essential for our objective to faithfully approximate the true objective in \cref{eq:max_min_vpr}. We discuss the theoretical implications of using the sigmoid loss in more detail in \Cref{app:theo_mean_est}.
    
\textbf{Replacing $\sigma(f_k(\wb)-\rho_k)$ with  $\sigma(F_k(\wb)-\rho_k)$:} 
As clients do not have access to their true distribution $\mathcal{D}_k$ to compute $f_k(\cdot)$ we propose to use an empirical estimate $\sigma(F_k(\wb)-\rho_k)$. This is again similar to what is done in standard FL where we minimize $F_k(\wb)$ instead of $f_k(\wb)$ at client $k$. Note that the global model $\wb$ is trained on the data of all clients, making it unlikely to overfit to the local data of any particular client, leading to $f_k(\wb) \approx F_k(\wb)$, also shown empirically in \Cref{app:er}.

With the Sigmoid relaxation, we present our proposed \incfl objective:
\begin{align}
\begin{aligned}
    \text{\incfl Obj.:}~\min_{\mathbf{w}}\widetilde{F}(\wb)=
    \min_{\mathbf{w}}\frac{1}{M}\sum_{i=1}^M\sgf_i(\wb),~\text{where}~\sgf_i(\wb)\defeq\sigm(F_i(\mathbf{w})-\rho_i). 
\label{eq:prop_obj}
\end{aligned}
\end{align}
Our empirical results in \Cref{sec:exp} support our intuition of these relaxations and  demonstrate that minimizing our proposed objective leads to a  higher \vpr than the standard FL objective. We also provide a theoretical analysis in \Cref{app:theo_mean_est} to illustrate how our \incfl objective behaves differently from the standard FL objective to maximize global model appeal for mean estimation. 

\section{Proposed \incfl Solver} \label{sec:solver} 
In this section, we present our \incfl~objective's solver. The \incfl algorithm enjoys the following properties, which make it a good candidate for real-world applications of cross-device FL: i) uses the same local SGD procedure as in standard FedAvg, ii) allows partial client participation, and iii) is stateless. By stateless, we mean that clients do not carry varying local parameters throughout training rounds, preventing issues from stale parameters~\cite{wang2021field}. 

With the sigmoid approximation of sign loss and for differentiable $F_k(\wb)$, our objective $\widetilde{F}(\wb)$ in \cref{eq:prop_obj} is differentiable and can be minimized with gradient descent and its variants. Its gradient is given by:
\begin{align}
    \nabla\widetilde{F}(\wb)=\frac{1}{M}\sum_{k=1}^M \underbrace{(1-\sgf_k(\wb))\sgf_k(\wb)}_{\text{aggregating weight}\defeq q_k(\wb)}\nabla F_k(\wb).
    \label{eq:weight}
\end{align}

Observe that $\nabla\widetilde{F}(\wb)$ is a \textbf{weighted aggregate} of the gradients of the clients' empirical losses, similar in spirit to the gradient $\nabla F(\wb)$ in standard FL. The key difference is that in \incfl, the weights $q_k(\wb)\defeq(1-\sgf_k(\wb))\sgf_k(\wb)$ depend on how much the global model appeals to the clients and are dynamically updated based on the current model $\bw$, as we discuss below.

\begin{wrapfigure}{!r}{.35\textwidth} \vspace{-1em}
\centering 
    \includegraphics[width=1\textwidth]{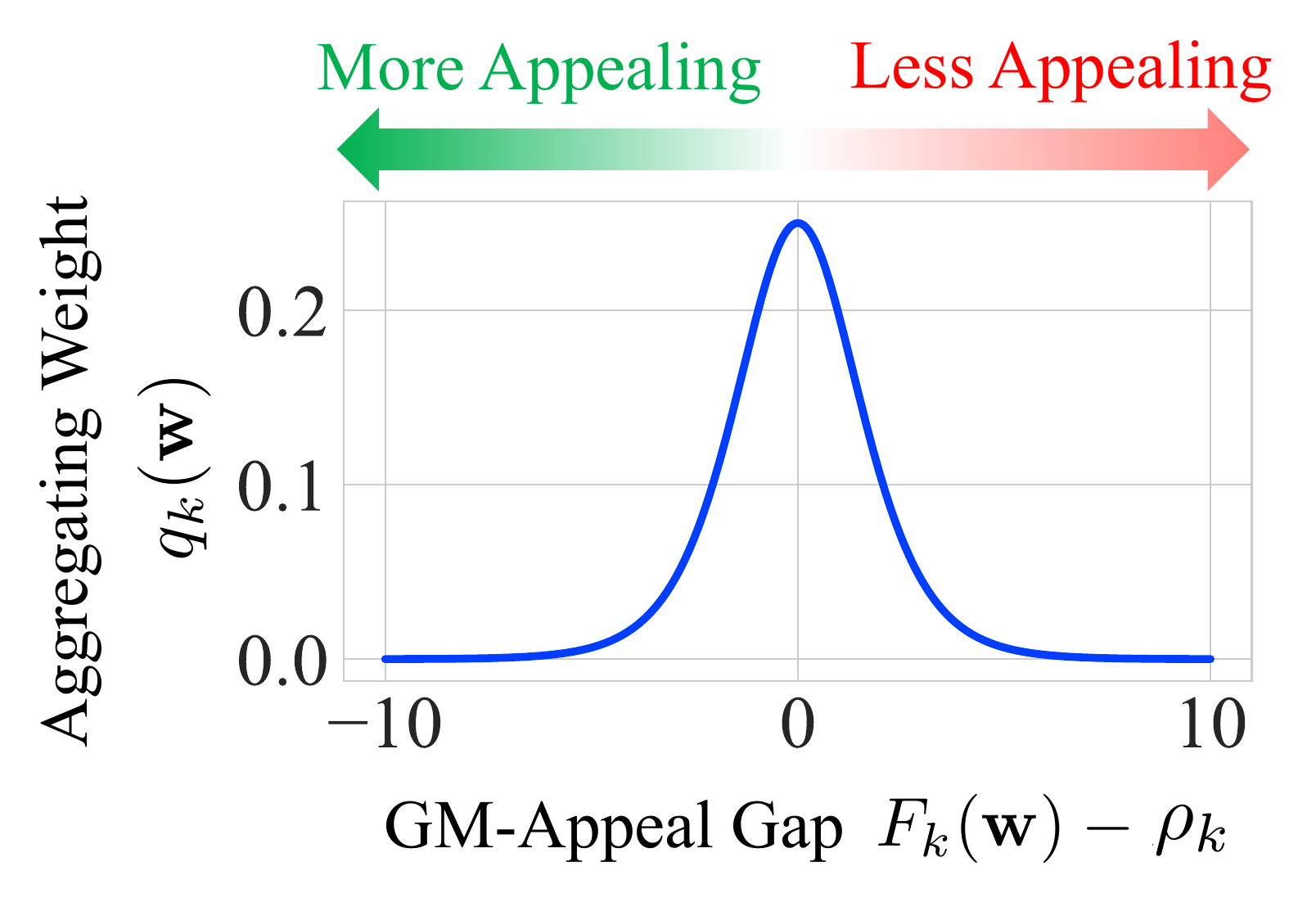}
    \caption{Aggregating weight $q_k(\wb)$ for any client $k$ versus the \vpr gap $F_k(\wb) - \rho_k$. The weight is the highest for the clients which the global model performs similarly to the clients' requirements since this allows it to increase \vpr without sacrificing other clients' requirements.} 
    \label{fig:incent} \vspace{-1em}
\end{wrapfigure}

\textbf{Behavior of the Aggregation Weights $q_k(\wb)$.} 
For a given $\wb$, the aggregation weights $q_k(\wb)$ depend on the \textit{\vpr Gap}, $F_k(\wb) - \rho_k$ (see \Cref{fig:incent}). 
When $F_k(\wb) \ll \rho_k$, the global model $\wb$ sufficiently meets the client's requirement. Therefore, \incfl sets $q_k(\wb) \approx 0$ to focus on the updates of other clients.
Similarly, if $F_k(\wb) \gg \rho_k$, \incfl sets $q_k(\wb) \approx 0$. This is because $F_k(\wb) \gg \rho_k$ implies that the current model $\wb$ is incompatible with the requirement of client $k$ and hence it is better to avoid optimizing for this client at the risk of sacrificing other clients' requirements. \incfl gives the highest weight to clients for which the global model performs similarly to the clients' requirements since this allows it to increase the \vpr without sabotaging other clients' requirements. 

\textbf{A Practical \incfl Solver.} Directly minimizing the \incfl objective using gradient descent can be slow to converge and impractical, as it requires all clients to be available for training. Instead, we propose a practical \incfl algorithm, which uses multiple local updates at each client to speed up convergence as done in standard FL \cite{mcmahan2017communication} and allow partial client availability. 

We use the superscript $(t,r)$ to denote the communication round $t$ and the local iteration index $r$. In each round $t$, the server selects a new set of clients $\mathcal{S}^{(t,0)}$ uniformly at random and sends the most recent global model $\wb^{(t,0)}$ to the clients in $\mathcal{S}^{(t,0)}$. 
Clients in $\mathcal{S}^{(t,0)}$ perform $\tau$ local iterations with a learning rate $\lrl$ to calculate their updates as follows:
\begin{align}
\label{eqn:local_model_update}
\wb_\ic^{(t,r+1)} =\wb_\ic^{(t,r)}-\lrl \gb(\wb_\ic^{(t,r)},\xi_\ic^{(t,r)}),\forall~r \in \{0,...,\tau-1\}
\end{align}
where $\gb(\wb_\ic^{(t,r)},\xi_\ic^{(t,r)})=\frac{1}{b}\sum_{\xi \in \xi_\ic^{(t,r)}}\nabla \lff(\wb_\ic^{(t,r)}, \xi)$ is the stochastic gradient computed using a mini-batch $\xi_\ic^{(t,r)}$ of size $b$ that is randomly sampled from client $\ic$'s local dataset $\mathcal{B}_k$. The weight $q_k(\wb_k^{(t,0)})$ can be computed at each client by calculating the loss over its training data with $\wb_k^{(t,0)}$, which is a simple inference step. 
Clients in $\mathcal{S}^{(t,0)}$ then send their local updates $\Delta \wb_k^{(t,0)}\defeq \wb_{k}^{(t,\tau)}-\wb_k^{(t,0)}$ and weights $q_k(\wb_k^{(t,0)})$ back to the server, which updates the global model as follows:
\begin{flalign}
\label{eqn:global_model_update}
\begin{aligned}
\wb^{(t+1,0)}=\wb^{(t,0)}-\eta^{(t,0)}_g\hspace*{-0.8em}\sum_{k\in\mathcal{S}^{(t,0)}}q_k(\wb^{(t,0)})\Delta \wb_k^{(t,0)}
\end{aligned}
\end{flalign}
where $\eta^{(t,0)}_g=\frac{\eta_g}{\sum_{k\in\mathcal{S}^{(t,0)}}q_k(\wb^{(t,0)})+\epsilon}$ is the adaptive server learning rate with global learning rate $\eta_g$ and $\epsilon>0$. We discuss the reasoning for such a learning rate below.

\comment{denominator is dependent and is the motivation for the normalization we do in }

\setlength{\textfloatsep}{1em}
\begin{algorithm} [t]
\caption{Our Proposed \incfl Solver}\label{algo1}
\renewcommand{\algorithmicloop}{\textbf{Global server do:}}
\begin{algorithmic}[1]
\STATE {\bfseries Input:} mini-batch size $b$, local iteration steps $\tau$, client requirement $\rho_k,~k\in[M]$
\STATE {\bfseries Output:} Global model $\wb^{(T,0)}$
\STATE {\bfseries Initialize:} Global model $\wb^{(0,0)}$
\STATE {\bfseries For ${t=0,...,T-1}$ communication rounds do}:
\STATE \hspace*{1em} {\bfseries Global server do:}\\
\STATE \hspace*{2em} Select $m$ clients for $\mathcal{S}^{(t,0)}$ uniformly at random and send $\wb^{(t,0)}$ to clients in $\mathcal{S}^{(t,0)}$
\STATE \hspace*{1em} {\bfseries Clients $\ic\in\mathcal{S}^{(t,0)}$ in parallel do:}
\STATE \hspace*{2em} {Set $\wb_\ic^{(t,0)}=\wb^{(t,0)}$}, and calculate $q_\ic(\wb_\ic^{(t,0)})=\sigm(F_\ic(\wb_\ic^{(t,0)})- \rho_k)$\\
\STATE {\hspace*{2em} {\bfseries For $r=0,...,\tau-1$ local iterations do:}}\\
\STATE {\hspace*{3em}Update $\wb_\ic^{(t,r+1)}\leftarrow\wb_\ic^{(t,r)}-\lrl\gb(\wb_\ic^{(t,r)},\xi_\ic^{(t,r)})$} \\
\STATE {\hspace*{2em} Send $\Delta \wb_\ic^{(t,0)}=\wb_\ic^{(t,0)}-\wb_\ic^{(t,\tau)}$ and aggregation weight $q_\ic(\wb_\ic^{(t,0)})$ to the server}
\STATE {\hspace*{1em} {\bfseries Global server do:}}\\
\STATE {\hspace*{2em} Update global model with $\wb^{(t+1,0)}=\wb^{(t,0)}-\eta^{(t,0)}_g\sum_{k\in\mathcal{S}^{(t,0)}}q_k(\wb^{(t,0)})\Delta\wb_k^{(t,0)}$}\\
\end{algorithmic} 
\end{algorithm} 

\textbf{Adaptive Server Learning Rate for \incfl.} With $L_c$ continuous and $L_s$ smooth $F_k(\wb),~\forall k \in [M]$ (see \Cref{as1}), the objective $\widetilde{F}(\wb)$ is $\widetilde{L}_s$ smooth where $\widetilde{L}_s =\frac{L_s}{M}\sum_{k=1}^M q_k(\wb) + \frac{L_c}{4}$ (see \Cref{app:the1proof}). Hence, the optimal learning rate $\Tilde{\eta}$ for the \incfl~is given by, $\widetilde{\eta} = 1/\widetilde{L}_s =M\eta/\left({\sum_{k=1}^M q_k(\wb) + \epsilon}\right)$,
where $\eta = \frac{1}{L_s}$ is the optimal learning rate for standard FL and $\epsilon = \frac{ML_c}{4L_s}$ > 0 is a constant. The denominator of the optimal $\widetilde{\eta}$ is proportional to the sum of the aggregation weights $q_k(\wb)$ and acts as a dynamic normalizing factor. Therefore, we propose using an adaptive global learning rate $\eta^{(t,0)}_g=\eta_g/(\sum_{k \in\mathcal{S}^{(t,0)}}q_k(\wb^{(t,0)})+\epsilon)$ with hyperparameters $\eta_g$, $\epsilon$. 



\textbf{Example of $\rho_k$ as $F_k(\widehat{\wb}_k)$ for \incfl.} One intuitive way to set $\rho_k$ for each client is to set it as the training loss value $F_k(\widehat{\wb}_k)$ where $\widehat{\wb}_k$ is a client local model that is solo-trained with a few warm-up local SGD steps on its local data. The loss value only needs to be computed once and saved as a constant beforehand at each client. The number of steps for training $\widehat{\wb}_k$ can be entirely dependent on the personal resources and requirements of the clients. 
For our experiments, we use the same number of warm-up SGD steps (100 iterations) to achieve the local model $\widehat{\wb}_k$ across all algorithms and set $\rho_k=F_k(\widehat{\wb}_k)$ for all our experiments. This gives us a reasonable requirement for each client which is the realistic estimate of the local model accuracy at a client without assuming any significant computation burden at clients. We have also included an ablation study on the effect of the number of local steps to obtain $\widehat{\wb}_k$ in \Cref{app:er}.
\label{para:setting_rho_k}

\textbf{Appeal-based Flexible Client Participation.} It may appear that our \incfl~solver in \Cref{algo1} requires clients to always participate in FL if selected even when the global model does not appeal to them. However, our algorithm can be easily modified to allow clients to participate flexibly during training depending on whether they find the global model appealing or not. For such appeal-based flexible client participation, we assume that clients are available for training if selected only during a few initial training rounds. After these rounds, clients may participate only if they find the global model appealing. We demonstrate this extension of \incfl~with appeal-based flexible client participation in \Cref{tab:testacc} and \Cref{tab:byz}. These experiments show that with flexible client participation, retaining a high global model is even more imperative for the server to achieve good test accuracy and generalization performance. We also show that even after we allow clients to participate flexibly, \incfl retains a significantly higher number of clients that find the global model appealing compared to the other baselines.

\subsection{Convergence Properties of \incfl} \label{sec:conv} \vspace{-0.5em}
In this section we show the convergence guarantees of~\incfl in \Cref{algo1}. Our convergence analysis shows that the gradient norm of our global model goes to zero, and therefore we converge to a stationary point of our objective $\sgf(\wb)$. First, we introduce the assumptions and definitions utilized for our convergence analysis below.



\begin{assumption}[Continuity \& Smoothness of $F_k(\wb),~\forall~k$] The local objective functions $F_1(\wb),~...,F_M(\wb)$, are $L_c$-continuous and $L_s$-smooth for any $\wb$. \label{as1}
\end{assumption} 
\begin{assumption}[Unbiased Stochastic Gradient with Bounded Variance for $F_k(\wb),~\forall~k$]
For the mini-batch $\xi_k$ uniformly sampled at random from $\bdat$, the resulting stochastic gradient is unbiased, i.e., $\expt[\sgrad(\wb_\ic,\xi_k)]=\nabla \lf(\wb_\ic)$. The variance of stochastic gradients is bounded: $\expt[\|\sgrad(\wb_\ic,\xi_k)-\nabla \lf(\wb_\ic)\|^2]\leq\sigma_g^2$ for $k\in[M]$. \label{as2}
\end{assumption}
\begin{assumption}[Bounded Dissimilarity of $F(\wb)$] There exists $\beta^2\geq1,~\kappa^2\geq0$ such that $\frac{1}{M}\sum_{i=1}^M\|\nabla F_i(\wb)\|^2\leq\beta^2\|\frac{1}{M}\sum_{i=1}^M\nabla F_i(\wb)\|^2+\kappa^2$ for any $\wb$. \label{as4}
\end{assumption}
\Cref{as1}-\ref{as4} are standard assumptions used in the optimization literature~\cite{stich2018local,karimireddy2019scaffold,bist2020distdis,wang2020tackling}, including the $L_c$-continuity assumption~\cite{shwartz2009sco,riis2021geononconvex}. Note that we do not assume anything for our proposed objective function $\sgf(\wb)$ and only have assumptions over the standard objective function $\gf(\wb)$ to prove the convergence of \incfl over $\sgf(\wb)$ in \Cref{the1}. 
\begin{theorem}[Convergence to the \incfl Objective $\widetilde{F}(\wb)$] Under \Cref{as1}-\ref{as4}, suppose the server uniformly selects $m$ out of $M$ clients without replacement in each round of \Cref{algo1}. With $\eta_l = \frac{1}{\sqrt{T}\tau},~\eta_g = \sqrt{\tau m}$, for a sufficiently large $T$ we have: 
\begin{flalign}
\begin{aligned}
\min_{t\in[T]}\expt\left[\left\|\nabla\sgf(\wb^{(t,0)})\right\|^2\right]\leq \mathcal{O}\left(\frac{\sigma_g^2}{\sqrt{m\tau T}}\right)
+\mathcal{O}\left(\frac{\sgn^2}{T \tau }\right)+\mathcal{O}\left(\frac{\sqrt{\tau}}{\sqrt{Tm}}\right)
+\mathcal{O}\left(\frac{\kappa^2+\beta^2}{T}\right)
\end{aligned}
\end{flalign}
where $\mathcal{O}$ subsumes all constants (including $L_s$ and $L_c$). 
\label{the1}
\end{theorem}
\Cref{the1} shows that with a sufficiently large number of communication rounds $T$ we reach a stationary point of our objective function $\widetilde{F}(\wb)$. The proof is deferred to \Cref{app:the1proof} where we also show a version of this theorem that contains the learning rates $\lrg$ and $\lrl$ with the constants.

\comment{
\begin{theorem}[Convergence to the Standard FedAvg Objective $F(\wb)$] Under the same conditions that are used for \Cref{the1}, the optimization error of the proposed algorithm $\incfl$ in \Cref{algo1} with respect to the standard objectvie $\widetilde{F}(\wb)$ is bounded as follows:
\begin{align}
     \min_{t\in[T]}\|\nabla F(\wb^{(t,0)})\|^2
    \leq 2(2\beta^2+1)\rho+8\beta^2 L_c^2+2\kappa^2 \label{eq:the2}
\end{align} \label{the2}
\end{theorem}
where $\rho$ is the optimization error bound of $\incfl$ with respect to $\widetilde{F}(\wb)$ in \Cref{the1}. \Cref{the2} shows that although \incfl aims to optimize $\widetilde{F}(\wb)$, a non-linear combination of $F_k(\wb),k\in[M]$, which is different from $F(\wb)$, the optimization error with respect to $F(\wb)$ is bounded by the bounded dissimilarity parameters $\beta^2\geq 1,\kappa^2\geq 0$ and the $L_c$ continuous property of $F_k(\wb),k\in[M]$.

\DiJ{Not sure if this is relevant with the dependence on $L_c$ and $\kappa^2$. For instance we have a tighter bound using just $L_c$ continuous assumptions.
For any $\wb$,
\begin{align*}
    ||\nabla F(\wb)||^2 = ||\frac{1}{M}\sum_{k=1}^M\nabla F_k(\wb)||^2 \leq \frac{1}{M}\sum_{i=1}^M || \nabla F_k(\wb)||^2 \leq L_c
\end{align*}
Therefore $\min_{t\in[T]}\|\nabla F(\wb^{(t,0)})\|^2 \leq L_c$
}
}

\section{Related Work}
\label{sec:related_work}
To the best of our knowledge, the notion of \vpr and the proposal to maximize it while considering flexible client participation have not appeared before in the previous literature. Previous works have focused on the notion of satisfying clients' personal requirements from a game-theoretic lens or designing strategies specifically to prevent client dropout, including the use of personalization, which have their limitations, as we discuss below. 
\subsection{Incentivizing Clients and Preventing drop-out}
A recent line of work in game theory models FL as a coalition of self-interested agents and studies how clients can optimally satisfy their individual incentives defined differently from our goal. 
Instead of training a single global model, \cite{donahue2021modelshare, donahue2021optstab} consider the problem where each client tries to find the best possible coalition of clients to federate with to minimize its own error. 
\cite{blum2021onefone} consider an orthogonal setting where each client aims to satisfy its constraint of low expected error while simultaneously trying to minimize the number of samples it contributes to FL.
While these works establish useful insights for simple linear tasks, it is difficult to extend these to practical non-convex machine learning tasks.
In contrast to these works, in \incfl we aim to directly maximize the \textit{number} of satisfied clients using a global model. 
This perspective alleviates some of the analysis complexities occurring in game-theoretic formulations and allows us to consider general non-convex objective functions. 

A separate line of work looks at how to prevent and deal with client drop-out in FL. \cite{wang2022friends} introduce a notion of `friendship' among clients and proposes to use friends' local update as a substitute for the update of dropped-out clients. \cite{gu2021fast} propose to use previous updates of dropped-out clients as a substitute for their current updates. Both algorithms are stateful. Another line of work \cite{ han2022tokenized, kang2019incentive, zhang2021faithful} aims to incentivize clients to contribute resources for FL and promote long-term participation by providing monetary compensation for their contributions, determined using game-theoretic tools.  These techniques are orthogonal to \incfl's formulation and can be combined if needed to further incentivize clients.

\subsection{Personalized and Fair Federated Learning}
Personalized federated learning (PFL) methods are able to increase performance by training multiple related models across the network~\cite[e.g.,][]{smith2017federated}.
In contrast to PFL, \incfl focuses on the more challenging goal of training a \textit{single} global model that can maximize the number of clients for which the global model outperforms their local model.
%
This is because, unlike PFL which may require additional training on new clients for personalization, \incfl's global model can be used by new clients without additional training (see \Cref{tab:unseen}). Also, \incfl is stateless, in that clients do not carry varying local parameters throughout training rounds as in many popular personalized FL methods \cite{smith2017federated,dinh2020pfedme,fallah2020personalized,li2021ditto}, preventing parameter staleness problems which can be exacerbated by partial client participation~\cite{wang2021field}. Furthermore, \incfl is orthogonal to and can be combined with PFL methods. We demonstrate this in \Cref{tab:local}, where we show results for \incfl~jointly used with personalization via fine-tuning~\cite{jiang2019improving}. 
We compare \incfl$+$Fine-tuning with another well known PFL method PerFedAvg~\cite{fallah2020personalized} and show that \incfl appeals to a significantly higher number of clients than the baseline.

Finally, another related area is fair FL, where a common goal is to train a global model whose accuracy has less variance across the client population than standard FedAvg~\cite{li2019fair,mohri2019agnostic}. A side benefit of these methods is that they can improve global model appeal for the worst performing clients. However, the downside is that the performance of the global model may be degraded for the best performing clients, thus making it unappealing for them to participate. We show in \Cref{app:er} that fair FL methods are indeed not effective in increasing \vpr.

\comment{
\begin{figure}[!t]
\centering
\begin{subfigure}{0.23\textwidth}
\centering
\includegraphics[width=1\textwidth]{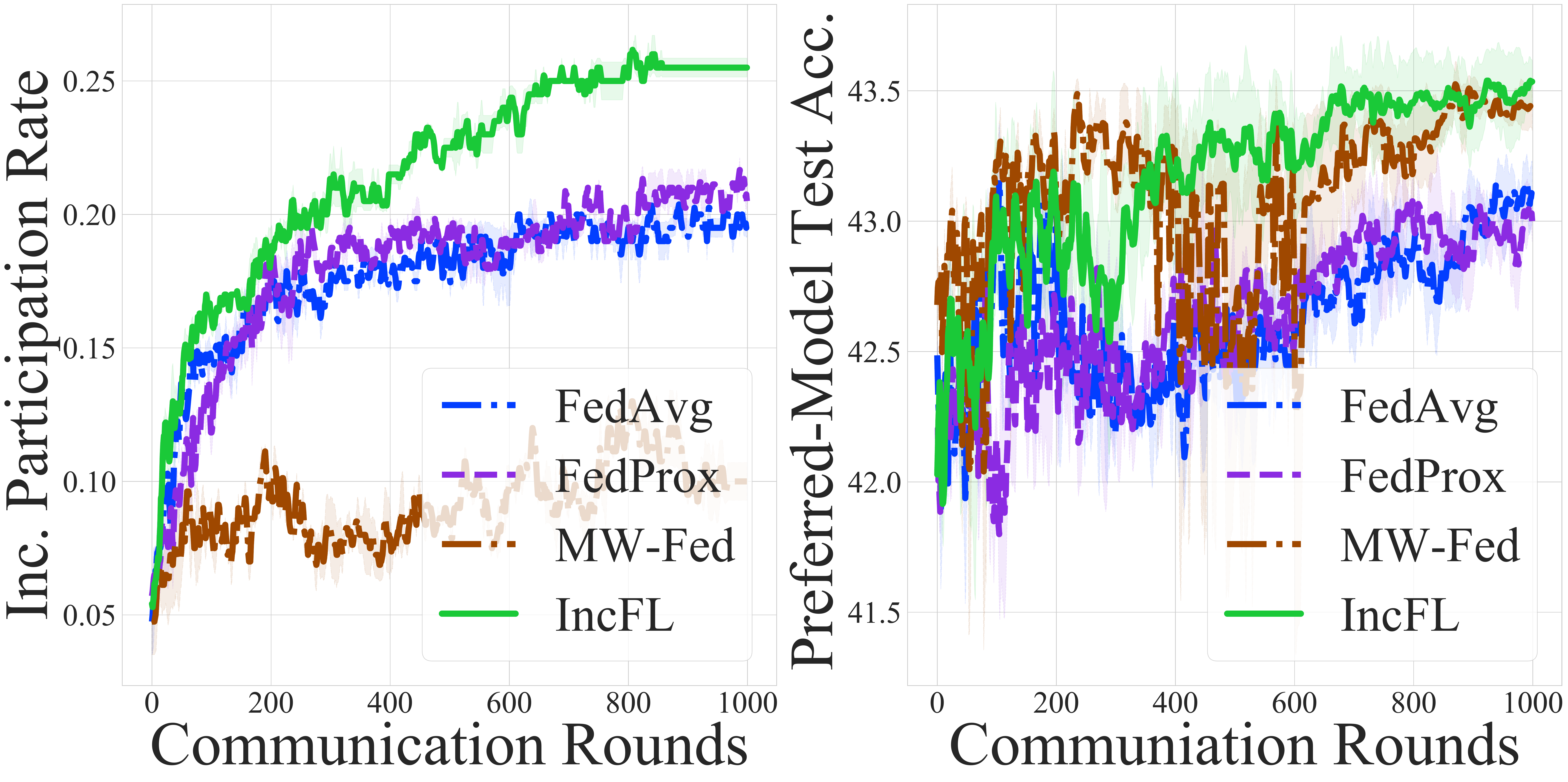} \caption{Training Data Ratio $40\%$}
\end{subfigure} \hfill
\begin{subfigure}{0.23\textwidth}
\centering
\includegraphics[width=1\textwidth]{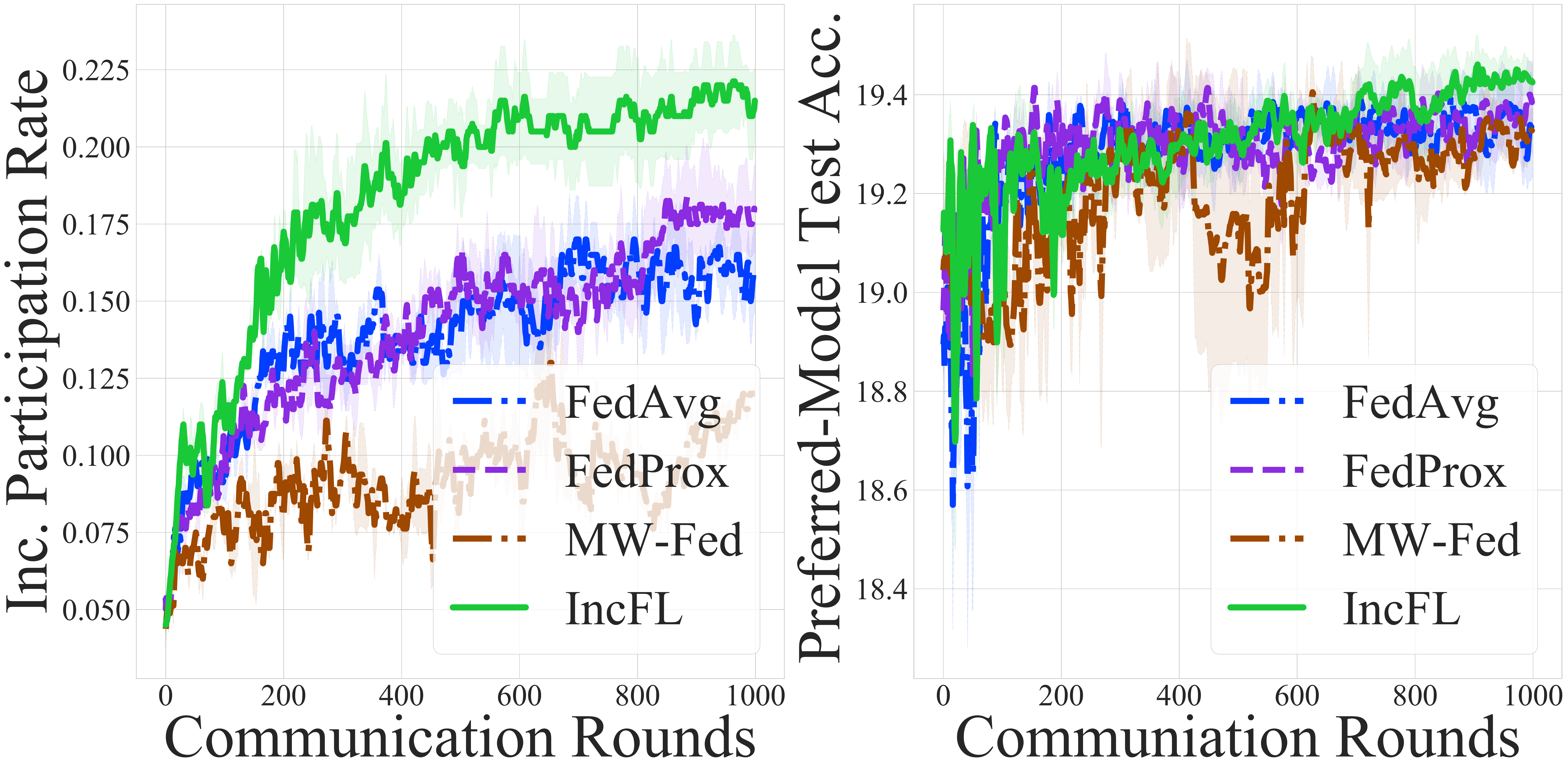} \caption{Training Data Ratio $80\%$}
\end{subfigure} 
 \caption{Incentivized participation rate (\text{GM-Appeal}), i.e., the fraction of incentivized clients, and preferred-model test accuracy evaluated on the test data for the synthetic data with the training clients. \incfl improves on both \text{GM-Appeal} and preferred-model test accuracy for both smaller ($40\%$) and larger ($80\%$) training data ratios. The \text{GM-Appeal} improvement of \incfl is larger for the smaller training data ratio since the solo-trained local model of a client may not sufficiently generalize well to its test data.  \yj{only show GM-Appeal for this figure for the goal of showing the effect of training data ratio to \vpr}
 } \label{fig:syn}  \vspace{-1em}
\end{figure}
}

\renewcommand{\arraystretch}{1.2}
\begin{table*}[!t]  \centering
\setlength\tabcolsep{0.3pt}
\begin{tabular}{@{}l cc cc cc cc@{}} \dtoprule 
 & \multicolumn{4}{c}{\small{Seen Clients}} & \multicolumn{4}{c}{\small{Unseen Clients}} \\ 
 \cmidrule(lr){2-5} \cmidrule(lr){6-9}
 & \multicolumn{2}{c}{\small{FMNIST}} & \multicolumn{2}{c}{\small{EMNIST}} &\multicolumn{2}{c}{\small{FMNIST}} & \multicolumn{2}{c}{\small{EMNIST}} \\   \cmidrule(lr){2-3} \cmidrule(lr){4-5} \cmidrule(lr){6-7} \cmidrule(lr){8-9}
  & {\small{Test Acc.}} & {\small{\vpr}} &  {\small{Test Acc.}} & {\small{\vpr}} &{\small{Test Acc.}} & \small{{\vpr}} & \small{Test Acc.} & \small{\vpr}  \\ \hdashline
 \small{FedAvg} & $43.70{\scriptstyle (\pm 0.02)}$ & $0.04{\scriptstyle (\pm 0.0)}$ & $35.15{\scriptstyle (\pm 0.51)}$ & $0.02{\scriptstyle (\pm 0.01)}$ &  $43.14{\scriptstyle (\pm 0.23)}$ & $0.07{\scriptstyle (\pm 0.01)}$ & $37.14{\scriptstyle (\pm 0.10)}$ & $0.06{\scriptstyle (\pm 0.0)}$  \\  
 \small{FedProx}  &  $44.59{\scriptstyle (\pm 1.94)}$ & $0.05{\scriptstyle (\pm 0.01)}$ & $34.06{\scriptstyle (\pm 1.21)}$ & $0.004{\scriptstyle (\pm 0.0)}$ & $43.80{\scriptstyle (\pm 1.67)}$ & $0.07{\scriptstyle (\pm 0.01)}$ & $36.82{\scriptstyle (\pm 0.22)}$ & $0.008{\scriptstyle (\pm 0.0)}$ \\  
  \small{Scaffold}  &  $39.90{\scriptstyle (\pm 0.59)}$ & $0.0{\scriptstyle (\pm 0.0)}$ & $34.78{\scriptstyle (\pm 2.05)}$ & $0.0{\scriptstyle (\pm 0.0)}$& $39.24{\scriptstyle (\pm 0.68)}$ & $0.01{\scriptstyle (\pm 0.0)}$  &$34.19{\scriptstyle (\pm 1.25)}$ & $0.004{\scriptstyle (\pm 0.0)}$ \\  \hdashline
   \small{PerFedAvg}  &  $46.62{\scriptstyle (\pm 1.0)}$ & $0.05{\scriptstyle (\pm 0.0)}$ & $34.78{\scriptstyle (\pm 1.05)}$ & $0.003{\scriptstyle (\pm 0.0)}$ & $46.00{\scriptstyle (\pm 0.87)}$ & $0.07{\scriptstyle (\pm 0.0)}$  & $36.92{\scriptstyle (\pm 0.51)}$ & $0.008{\scriptstyle (\pm 0.0)}$\\ 
\small{qFFL} &  $29.92{\scriptstyle (\pm 3.13)}$ & $0.0{\scriptstyle (\pm 0.0)}$ & $15.95{\scriptstyle (\pm 3.02)}$ & $0.0{\scriptstyle (\pm 0.0)}$ & $19.63{\scriptstyle (\pm 2.17)}$ & $0.0{\scriptstyle (\pm 0.0)}$ & $5.41{\scriptstyle (\pm 0.52)}$ & $0.0{\scriptstyle (\pm 0.0)}$\\ \hdashline
\small{MW-Fed} &  $44.41{\scriptstyle (\pm 2.38)}$ & $0.04{\scriptstyle (\pm 0.0)}$
          &$30.44{\scriptstyle (\pm 3.07)}$ & $0.01{\scriptstyle (\pm 0.0)}$ &$43.46{\scriptstyle (\pm 2.15)}$ & $0.06{\scriptstyle (\pm 0.0)}$ &  $36.54{\scriptstyle (\pm 0.40)}$ & $0.01{\scriptstyle (\pm 0.0)}$\\
\small{\incfl} &  $\mathbf{70.86}{\scriptstyle (\pm 2.18)}$ & $\mathbf{0.37}{\scriptstyle (\pm 0.05)}$ &$\mathbf{57.34}{\scriptstyle (\pm 1.41)}$ & $\mathbf{0.25}{\scriptstyle (\pm 0.03)}$ & $\mathbf{74.53{\scriptstyle (\pm 0.50)}}$ & $\mathbf{0.39{\scriptstyle (\pm 0.07)}}$ &$\mathbf{55.62}{\scriptstyle (\pm 0.86)}$ & $\mathbf{0.31}{\scriptstyle (\pm 0.03)}$ \\
 \dbottomrule  \end{tabular} \vspace{-0.5em} \caption{Avg. test accuracy and GM-Appeal where we train for 200 communication rounds. At the 10th communication round, we let clients flexibly opt-out or opt-in depending on whether the global model has met their requirements. We report the final avg. test accuracy and GM-Appeal at the 200th communication round. \label{tab:testacc} }
\end{table*}

\section{Experiments} \label{sec:exp} 

\paragraph{Datasets and Model.} We evaluate~\incfl in three different settings: image classification for non-iid partitioned (i) FMNIST~\cite{xiao2017fmnist}, (ii) EMNIST with 62 labels~\cite{cohen2017emnist}, and (iii) sentiment analysis for (iv) Sent140~\cite{alec2009sent}with a MLP. For FMNIST, EMNIST, and Sent140 dataset, we consider 100, 500, and 308 clients in total that are used for training where we select 5 and 10 clients uniformly at random per round for FMNIST and EMNIST, Sent140 respectively. These clients are active at some point in training the global model and we call them \textbf{`seen clients'}. We also sample the \textbf{`unseen clients'} from the same distribution from which we generate the seen clients, with 619 clients for Sent140, 100 clients for FMNIST, and 500 for EMNIST. These unseen clients represent new incoming clients that have not been seen before during the training rounds of FL to evaluate the generalization performance at inference. 
Further details of the experimental settings are deferred to \Cref{app:ed}. 
\vspace{-0.5em}
\paragraph{Baselines.} We compare \incfl with numerous well-known FL algorithms such as standard FedAvg~\cite{mcmahan2017communication}; 
 FedProx \cite{sahu2019federated} which aims to tackle data heterogeneity; SCAFFOLD~\cite{karimireddy2019scaffold} which aims for variance-reduction; PerFedAvg \cite{fallah2020personalized} which facilitates personalization; MW-Fed \cite{blum2021onefone} which incentivizes client participation; and qFFL which facilitates fairness~\cite{li2019fair}. For all algorithms, we set $\rho_k$ to be the same, i.e., $\rho_k=F_k(\widehat{\wb}_k)$, where $\widehat{\wb}_k$ is obtained by running a few warm-up local SGD steps on client $k$'s data as outlined in \cref{para:setting_rho_k}.
 We do this to ensure a fair comparison across baselines. We perform grid search for hyperparameter tuning for all different baselines and choose the best performing ones.

\comment{
\renewcommand{\arraystretch}{1.1}
\begin{table}[!t]  \centering
\setlength\tabcolsep{0.7pt} \small
\begin{tabular}{@{}l||cc||cc@{}} \dtoprule 
 & \multicolumn{2}{c}{Seen Clients} & \multicolumn{2}{c}{Unseen Clients} \\ 
 \cmidrule(lr){2-3} \cmidrule(lr){4-5}
  & {Test Acc.} & {\vpr} &  {Test Acc.} & {\vpr}  \\ 
 FedAvg & $26.41{\scriptstyle (\pm 0.45)}$ & $0.01{\scriptstyle (\pm 0.01)}$  &  $25.57{\scriptstyle (\pm 0.07)}$ & $0.01{\scriptstyle (\pm 0.0)}$  \\  
 FedProx  &  $27.48{\scriptstyle (\pm 1.68)}$ & $0.01{\scriptstyle (\pm 0.0)}$  & $26.67{\scriptstyle (\pm 1.38)}$ & $0.01{\scriptstyle (\pm 0.0)}$  \\  
  Scaffold  &  $26.92{\scriptstyle (\pm 1.43)}$ & $0.01{\scriptstyle (\pm 0.0)}$ & $26.34{\scriptstyle (\pm 1.40)}$ & $0.01{\scriptstyle (\pm 0.0)}$   \\  
   PerFedAvg  &  $26.74{\scriptstyle (\pm 1.17)}$ & $0.01{\scriptstyle (\pm 0.0)}$  & $26.10{\scriptstyle (\pm 1.35)}$ & $0.01{\scriptstyle (\pm 0.0)}$  \\ 
     pFedme  &  $14.90{\scriptstyle (\pm 3.51)}$ & $0.0{\scriptstyle (\pm 0.0)}$  &$14.92{\scriptstyle (\pm 4.93)}$ & $0.0{\scriptstyle (\pm 0.0)}$   \\ 
          MW-Fed &  $26.00{\scriptstyle (\pm 2.89)}$ & $0.01{\scriptstyle (\pm 0.0)}$ & $25.32{\scriptstyle (\pm 2.52)}$ & $0.01{\scriptstyle (\pm 0.0)}$\\
            qFFL &  $14.71{\scriptstyle (\pm 3.48)}$ & $0.0{\scriptstyle (\pm 0.0)}$  & $19.63{\scriptstyle (\pm 2.17)}$ & $0.0{\scriptstyle (\pm 0.0)}$ \\
           \incfl &  $\mathbf{74.48}{\scriptstyle (\pm 1.35)}$ & $\mathbf{0.33}{\scriptstyle (\pm 0.05)}$  & $\mathbf{72.85{\scriptstyle (\pm 0.98)}}$ & $\mathbf{0.40{\scriptstyle (\pm 0.0)}}$\\
 \dbottomrule  \end{tabular} \vspace{-0.5em} \caption{Avg. test accuracy and GM-Appeal across 100 for FMNIST where we train for 200 communication rounds and select 3 clients each round. At the 5th communication round, we let clients flexibly opt-out or opt-in depending on whether the global model has met their requirements. We report the final avg. test accuracy and GM-Appeal at the 200th communication round. \label{tab:testacc2} }
\end{table}}
 
\renewcommand{\arraystretch}{1.2}
\begin{table*}[!t]  \centering \small
\setlength\tabcolsep{1pt}
\begin{tabular}{@{}l cc cc cc cc@{}} \dtoprule 
&  \multicolumn{8}{c}{Seen Clients} \\  
 \cmidrule(lr){2-9}
 & \multicolumn{4}{c}{FMNIST} & \multicolumn{4}{c}{EMNIST} \\ 
 \cmidrule(lr){2-5} \cmidrule(lr){6-9}
 & \multicolumn{2}{c}{Byz=0.1} & \multicolumn{2}{c}{Byz=0.05} &\multicolumn{2}{c}{Byz=0.1} & \multicolumn{2}{c}{Byz=0.05} \\   \cmidrule(lr){2-3} \cmidrule(lr){4-5} \cmidrule(lr){6-7} \cmidrule(lr){8-9}
  & {Test Acc.} & {\vpr} &  {Test Acc.} & {\vpr} &{Test Acc.} & {\vpr} &{Test Acc.} & {\vpr}  \\ \hdashline
 MW-Fed & $17.24~{\scriptstyle (\pm 2.35)}$ & $0.01~{\scriptstyle (\pm 0.0)}$ & $21.28~{\scriptstyle (\pm 1.79)}$ & $0.02~{\scriptstyle (\pm 0.0)}$ & $15.83~{\scriptstyle (\pm 1.52)}$ & $0.004~{\scriptstyle (\pm 0.0)}$  &$22.22~{\scriptstyle (\pm 0.63)}$ & $0.008~{\scriptstyle (\pm 0.001)}$  \\  
 \incfl  &  $\mathbf{69.42}~{\scriptstyle (\pm 2.87)}$ & $\mathbf{0.35}~{\scriptstyle (\pm 0.05)}$ & $\mathbf{70.60}~{\scriptstyle (\pm 2.76)}$ & $\mathbf{0.42}~{\scriptstyle (\pm 0.03)}$ & $\mathbf{52.74}~{\scriptstyle (\pm 0.44)}$ & $\mathbf{0.20}~{\scriptstyle (\pm 0.01)}$  & $\mathbf{56.10}~{\scriptstyle (\pm 0.77)}$ & $\mathbf{0.23}~{\scriptstyle (\pm 0.01)}$ \\ \hline \hline  \vspace{-1em}  \\
 ~ &  \multicolumn{8}{c}{Unseen Clients} \\ \cmidrule(lr){2-9}
 ~ & \multicolumn{4}{c}{FMNIST} & \multicolumn{4}{c}{EMNIST} \\ 
 \cmidrule(lr){2-5} \cmidrule(lr){6-9}
 & \multicolumn{2}{c}{Byz=0.1} & \multicolumn{2}{c}{Byz=0.05} &\multicolumn{2}{c}{Byz=0.1} & \multicolumn{2}{c}{Byz=0.05} \\   \cmidrule(lr){2-3} \cmidrule(lr){4-5} \cmidrule(lr){6-7} \cmidrule(lr){8-9}
  & {Test Acc.} & {\vpr} &  {Test Acc.} & {\vpr} &{Test Acc.} & {\vpr} &{Test Acc.} & {\vpr}  \\ \hdashline
 MW-Fed & $18.45~{\scriptstyle (\pm 2.81)}$ & $0.01~{\scriptstyle (\pm 0.0)}$ & $21.91~{\scriptstyle (\pm 3.81)}$ & $0.01~{\scriptstyle (\pm 0.0)}$ &  $17.03~{\scriptstyle (\pm 0.21)}$ & $0.005~{\scriptstyle (\pm 0.0)}$  & $22.23~{\scriptstyle (\pm 0.63)}$ & $0.003~{\scriptstyle (\pm 0.0)}$  \\  
 \incfl  &  $\mathbf{69.75}~{\scriptstyle (\pm 3.66)}$ & $\mathbf{0.39}~{\scriptstyle (\pm 0.01)}$ & $\mathbf{71.11}~{\scriptstyle (\pm 1.47)}$ & $\mathbf{0.46}~{\scriptstyle (\pm 0.01)}$ & $\mathbf{53.82}~{\scriptstyle (\pm 0.09)}$ & $\mathbf{0.26}~{\scriptstyle (\pm 0.02)}$  & $\mathbf{55.10}~{\scriptstyle (\pm 0.78)}$ & $\mathbf{0.28}~{\scriptstyle (\pm 0.01)}$ \\
 \dbottomrule  \end{tabular} \caption{Byzantine clients are included in the total clients where they artificially report large losses to the server and add noise to their gradients. The percentage of the Byzantine clients are denoted as `Byz'. We report the final avg. test accuracy and GM-Appeal across clients where we train for 200 communication rounds. At the 10th communication round, we let clients flexibly opt-out or opt-in depending on whether the global model has met their requirements.  \label{tab:byz} }
\end{table*}

\paragraph{Evaluation Metrics: \vpr, Average Test Accuracy, and Preferred-model Test Accuracy.} 
We evaluate \incfl and other methods with three key metrics: 1) \vpr, defined in \eqref{eq:gmas}, 2) average test accuracy (avg. test acc.) across clients, and a new metric that we propose called 3) preferred-model test accuracy. Preferred-model test accuracy is the average of the clients' test accuracies computed on either the global model $\wb$ or their solo-trained local model $\widehat{\wb}_k$, whichever one satisfies the client's requirement. We belive that average test accuracy is a more server-oriented metric as it assumes that clients will use the global model by default. On the other hand, preferred-model test accuracy is a more client-centric metric that allows clients to select the model which works best, thereby better reflecting their actual satisfaction. 
Ideally, it is desirable for an algorithm to improve all three metrics for both the server and clients to benefit from the algorithm. 

\subsection{Experiment Results}
\paragraph{Average Test Accuracy of Seen Clients \& Unseen Clients.} We first show that we improve the \vpr and thus the average test accuracy performance for the `seen clients' used during the training of the global model. In \Cref{tab:testacc}, we show the average test accuracy across clients where we let clients flexibly join or drop-out depending on whether the global model is appealing after $5\%$ of communication rounds of mandatory participation. We show that \incfl achieves the highest \vpr than other baselines for both FMNIST and EMNIST by $0.32$-$0.39$ and $0.23$-$0.31$ improvement, respectively. Since \incfl~is able to retain a larger pool of clients due to having a higher \vpr, it therefore trains from selecting from a more larger client pool, leading to the highest average test accuracy compared to the baselines by $22$-$40\%$ and $18$-$50\%$ improvement respectively for the seen and unseen clients. Since the other baselines do not consider the notion of \vpr entirely, it fails in preventing client dropouts leading to poor performance. Note that we do not use any of the `unseen clients' during training and only calculate the \vpr and test accuracy via inference with the global model trained with the `seen clients'.


\newfloatcommand{capbtabbox}{table}[][0.47\textwidth]
\begin{figure}[!t]
\begin{floatrow}
\ffigbox{\includegraphics[width=0.5\textwidth]{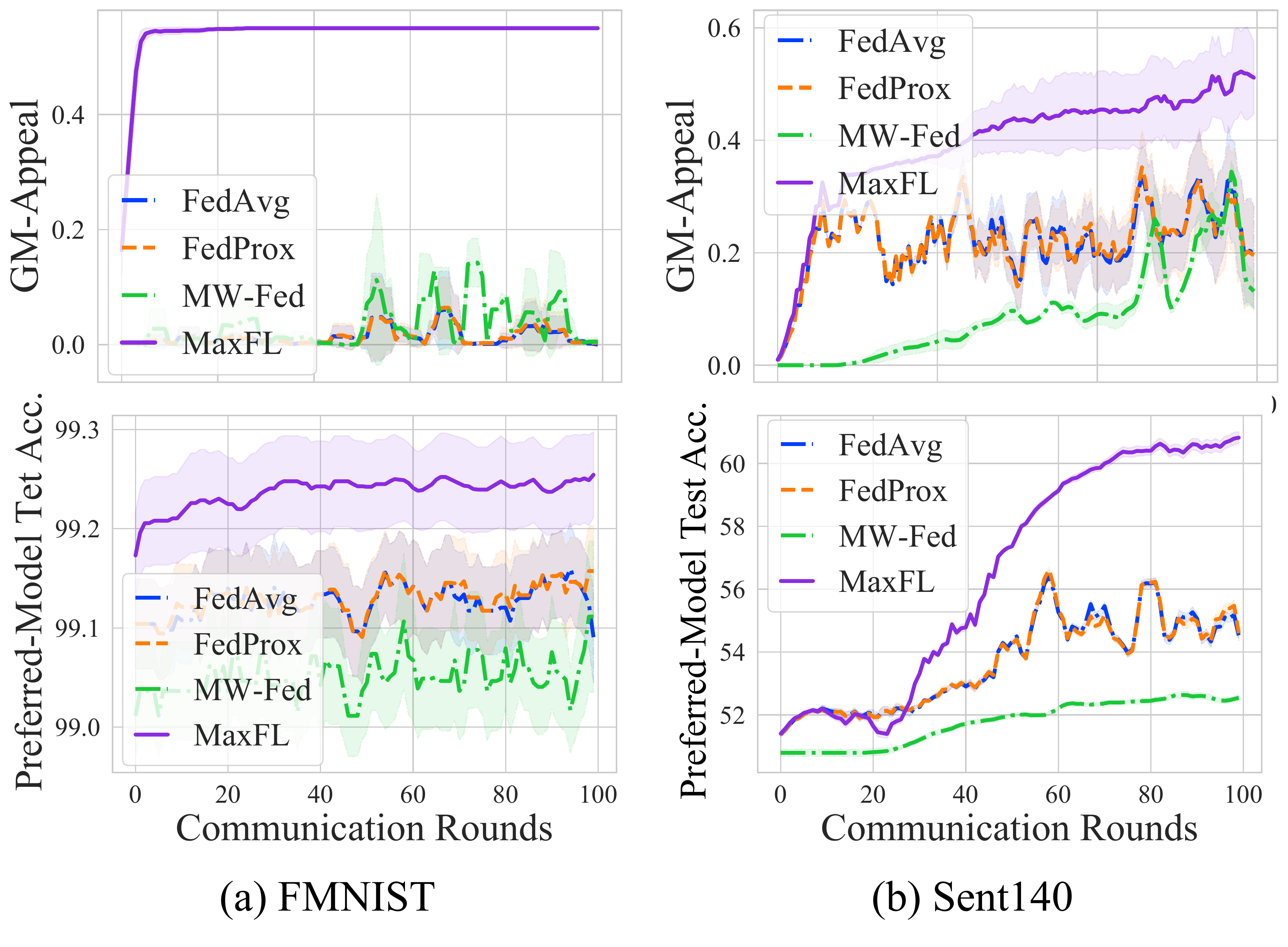}}
{\caption{\vpr (\textit{upper}) and preferred-model test accuracy (\textit{lower}) for the seen clients. For both datasets, the preferred-model test accuracy and \vpr is significantly higher for $\incfl$. Therefore, by using \incfl, clients can also benefit from choosing either the local or global model for best performance, while the server also gains a large number of clients to select from.} \label{fig:dnn}}
\capbtabbox{\renewcommand{\arraystretch}{1.7} \setlength\tabcolsep{0.05pt} \small \vspace{-6em}
\begin{tabular}{@{}l cc cc@{}} 
\dtoprule 
& \multicolumn{2}{c}{\vpr} & \multicolumn{2}{c}{Preferred-Model Test Acc.} \\ \cmidrule(lr){2-3} \cmidrule(lr){4-5}
& {FMNIST} &  {Sent140} & {FMNIST}  & {Sent140} \\ \hdashline
 FedAvg & $0.08{\scriptstyle (\pm 0.01)}$  & $0.37{\scriptstyle (\pm 0.07)}$ & $98.53{\scriptstyle (\pm 0.13)}$  & $57.05{\scriptstyle (\pm 1.44)}$ \\  
   FedProx & $0.07{\scriptstyle (\pm 0.01)}$  & $0.37{\scriptstyle (\pm 0.07)}$ & $98.43{\scriptstyle (\pm 0.21)}$  & $57.07{\scriptstyle (\pm 1.42)}$ \\ 
  Scaffold & $0.02{\scriptstyle (\pm 0.01)}$ & $0.03{\scriptstyle (\pm 0.05)}$ & $98.26{\scriptstyle (\pm 0.20)}$   & $51.59{\scriptstyle (\pm 0.11)}$\\ \hdashline
  MW-Fed & $0.05{\scriptstyle (\pm 0.04)}$  & $0.17{\scriptstyle (\pm 0.03)}$ & $98.32{\scriptstyle (\pm 0.13)}$   & $55.57{\scriptstyle (\pm 1.28)}$\\  
\incfl & $\mathbf{0.55}{\scriptstyle (\pm 0.0)}$ & $\mathbf{0.43}{\scriptstyle (\pm 0.05)}$ &  $\mathbf{98.83}{\scriptstyle (\pm 0.06)}$  & $\mathbf{57.16}{\scriptstyle (\pm 1.35)}$\\
 \dbottomrule  
\end{tabular} }
{\vspace{2em}\caption{\vpr and preferred-model test accuracy of the final global models for the unseen clients' test data that were not active during the training. \incfl improves the \vpr of the new incoming unseen clients by at least $47\%$ for FMNIST, and $6\%$ for Sent140 \label{tab:unseen} and achieves the same or higher preferred-model test
accuracy compared to that of all baselines.}}
\end{floatrow}
\end{figure}

\textbf{Robustness of \incfl Against Byzantine Clients.} One may think that \incfl~may be perceptible to attacks from Byzantine clients that intentionally send a greater \vpr gap to the server to gain a higher aggregation weight. To show \incfl's robustness against such attacks we show in \Cref{tab:byz} the performance of \incfl with Byzantine clients attacks which send higher losses to gain higher weights and then send Gaussian noise mixed gradients to the server. We compare with the MW-Fed baseline~\cite{blum2021onefone} which aims for incentivizing client participation by clients sending higher weights to the server and performing more local updates. In \Cref{tab:byz} we see that for both high and low byzantine client ratios, \incfl achieves only $1$-$5\%$ lower test accuracy for seen and unseen clients compared to the case where there are no Byzantine clients in \Cref{tab:testacc}. This is due to our objective (\Cref{eq:prop_obj}) giving lower weight to those clients that give a too high \vpr gap (see \Cref{fig:incent}). Hence while MW-Fed is susceptible to Byzantine attacks, \incfl disregards these clients that send artificially high \vpr gaps. \vspace{-1em}

\paragraph{Preferred-model Test Accuracy: Clients' Perspective.} In \Cref{fig:dnn} and \Cref{tab:unseen} we show the \vpr and preferred-model test accuracy for the seen and unseen clients respectively. Recall that a high preferred-model test accuracy implies that the client has a higher chance in satisfying its requirement by choosing between the global or solo-trained local model, whichever performs better. First, in \Cref{fig:dnn} we show that as the \vpr increases across the communication round, preferred-model test accuracy also increases. Among the other baselines, \incfl achieves the highest final \vpr and preferred-model test accuracy. This indicates that \incfl provides a win-win situation for both the server and the clients, since the clients have the highest accuracy by choosing the better model between the global model $\wb$ and the local model $\widehat{\wb}_k$, and the server has the highest fraction of participating clients. Similarly, in \Cref{tab:unseen}, we show that \incfl achieves the highest \vpr and preferred-model test accuracy. Although the preferred-model test accuracy improvement compared to the other baselines may appear small, showing that \incfl is able to maintain a high preferred-model test accuracy while also achieving a high \vpr implies that it does not sabotage the benefit of clients while also bringing the server more clients to select from.

\begin{wraptable}{!r}{0.48\textwidth} \centering \small
\setlength\tabcolsep{0.5pt} \vspace{-2em}
\begin{tabular}{@{}l cc cc@{}} 
\dtoprule 
& \multicolumn{2}{c}{Seen Clients} & \multicolumn{2}{c}{Unseen Clients} \\ \cmidrule(lr){2-3} \cmidrule(lr){4-5}
& {FMNIST} &  {Sent140} & {FMNIST} &  {Sent140} \\ \hdashline
 FedAvg & $0.38{\scriptstyle (\pm 0.06)}$ & $0.25{\scriptstyle (\pm 0.09)}$ & $0.39{\scriptstyle (\pm 0.06)}$  & $0.42{\scriptstyle (\pm 0.06)}$ \\  
   FedProx & $0.40{\scriptstyle (\pm 0.07)}$  & $0.26{\scriptstyle (\pm 0.09)}$ & $0.41{\scriptstyle (\pm 0.07)}$ &  $0.43{\scriptstyle (\pm 0.12)}$ \\  
  Scaffold & $0.02{\scriptstyle (\pm 0.02)}$ & $0.16{\scriptstyle (\pm 0.22)}$ & $0.03{\scriptstyle (\pm 0.02)}$  & $0.07{\scriptstyle (\pm 0.01)}$\\   \hdashline
  PerFedAvg & $0.45{\scriptstyle (\pm 0.05)}$  & $0.24{\scriptstyle (\pm 0.10)}$ & $0.46{\scriptstyle (\pm 0.06)}$ & $0.47{\scriptstyle (\pm 0.06)}$\\  
  MW-Fed & $0.28{\scriptstyle (\pm 0.07)}$  & $0.08{\scriptstyle (\pm 0.01)}$ & $0.39{\scriptstyle (\pm 0.04)}$   & $0.20{\scriptstyle (\pm 0.01)}$\\ 
\incfl & $\mathbf{0.55}{\scriptstyle (\pm 0.01)}$  & $\mathbf{0.36}{\scriptstyle (\pm 0.05)}$ &  $\mathbf{0.56}{\scriptstyle (\pm 0.01)}$  & $\mathbf{0.55}{\scriptstyle (\pm 0.01)}$\\
 \dbottomrule
\end{tabular}
\caption{\vpr of locally-tuned models with 5 local steps from the final global models for seen clients and unseen clients. Both for clients that are active during training and unseen test clients, \incfl increases the fraction of clients that find the global model appealing by at least $10\%$ as compared to all baselines. 
\vspace{-2.5em}} 
\label{tab:local} 
\end{wraptable}

\paragraph{Local Tuning for Personalization.} Personalized FL methods can be used to fine-tune the global model at each client before comparing it with the client's locally trained model. \incfl can be combined with these methods by simply allowing clients to perform some fine-tuning iterations before computing the aggregation weights in Step 7 of \Cref{algo1}. 
Both for clients that are active during training and unseen test clients, we show in \Cref{tab:local} that \incfl increases the \vpr by at least $10\%$ compared to all baselines. For FMNIST and Sent140, the improvement in \vpr over other methods is up to $27\%,~28\%$ respectively for active clients and $17\%,~4\%$ respectively for unseen clients.


\newpage
\section{Concluding Remarks} \vspace{-0.3em}

In this work, we aim to understand whether the global model can maximize the number of clients whose requirements are satisfied in FL, defining this via the novel notion of \vpr. We show that when  participating clients drop out or new clients do not join due to finding the global model less appealing, the test accuracy for the current training (seen) clients and generalization performance to the new unseen clients can suffer significantly. 
We show that our proposed \incfl framework (with convergence guarantees) which aims to maximize the global model appeal, is able to retain many clients for training and thus achieves a high average test accuracy across the participating clients and also across the new incoming clients. Moreover, we show additional benefits of \incfl such as being robust against Byzantine clients and improving the preferred-model test accuracy for the clients. A similar notion of global model appeal has not been properly examined before, and we expect our work to open up new research directions in understanding the role played by the server in preventing client dropout and recruiting new clients by finding a global model that can satisfy as many clients as possible. 


\bibliography{dist_sgd}
\bibliographystyle{unsrt}

\clearpage
\newpage
\onecolumn
\appendix
\section{Toy Example: Mean Estimation for \incfl} 
\label{app:mean_est}

\subsection{Maximizing \text{GM-Appeal} in Mean Estimation: Theoretical Analysis} \vspace{-0.3em}
\label{app:theo_mean_est}
We consider a setup with $M=2$ clients where the true loss function at each client is given by $f_k(w) = (w-\theta_k)^2$. In practice, clients only have $N_k$ samples drawn from the distribution given by $ e_{k,j} \sim \mathcal{N}(\theta_k, \nu^2), \; \forall j \in [N_k]$. We further assume that the empirical loss function at each client is given by $ F_k(w) = (w-\widehat{\theta}_k)^2 + (\widehat{\theta}_k - \theta_k)^2$ where $\widehat{\theta}_k$ is the empirical mean, $\widehat{\theta}_k= \frac{1}{|\mathcal{B}_k|}\sum_{j=1}^{N_{k}}e_{k,j}$.  
It is easy to see that the minimizer of $F_k(w)$ is the empirical mean  $\widehat{\theta}_k$. Thus, we set the solo-trained model at each client as $\widehat{w}_k = \widehat{\theta}_k$ and the loss threshold requirement at a client as $\rho_k = F_k(\widehat{w}_k) =  (\widehat{\theta}_k-\theta_k)^2$. 


\textbf{\text{GM-Appeal} for Standard FL Model Decreases Exponentially with Heterogeneity.} For simplicity let us assume $N_1 = N_2 = N$. Let $\gamma^2 = \nu^2/N$ be the variance of the local empirical means and $\gamma_G^2 = ((\theta_1-\theta_2)/2)^2 > 0$ be a measure of heterogeneity between the true means. The standard FL objective will always set the FL model to be the average of the local empirical means (i.e. $w = (\widehat{\theta}_1 + \widehat{\theta}_2)/2$) and does not take into account the heterogeneity among the clients. As a result, the \text{GM-Appeal} of the global model decreases \textit{exponentially} as $\gamma_G^2$ increases.
\begin{lemma}
The expected \text{GM-Appeal} of the standard FL model is upper bounded by $2\exp\left(-\frac{\gamma_G^2}{5\gamma^2} \right)$, where the expectation is taken over the randomness in the local datasets $\mathcal{B}_1,\mathcal{B}_2$.
\end{lemma}


\textbf{Maximizing \text{GM-Appeal} with Relaxed Objective.} 
We now explicitly maximize the \text{GM-Appeal} for this setting by solving for a relaxed version of the objective in \cref{eq:max_min_vpr} as proposed earlier. We replace the true loss $f_k(\cdot)$ by the empirical loss $F_k(\cdot)$ and replace the 0-1 (sign) loss with a differentiable approximation $h(\cdot)$.

We first show that setting $h(\cdot)$ to be a standard convex surrogate for the 0-1 loss (e.g. log loss, exponential loss, ReLU) leads to our new objective behaving the same as the standard FL objective.

\begin{lemma} 
Let $h$ be any function that is convex, twice differentiable, and strictly increasing in $[0,\infty)$. Then our relaxed objective is strictly convex and has a unique minimizer at $w^* = \left(\frac{\widehat{\theta}_1 + \widehat{\theta}_2}{2}\right)$.  \label{the-mean}
\end{lemma}





\begin{figure*}[!h]
\centering
\begin{subfigure}{0.3\textwidth}
\centering
\includegraphics[width=1\textwidth]{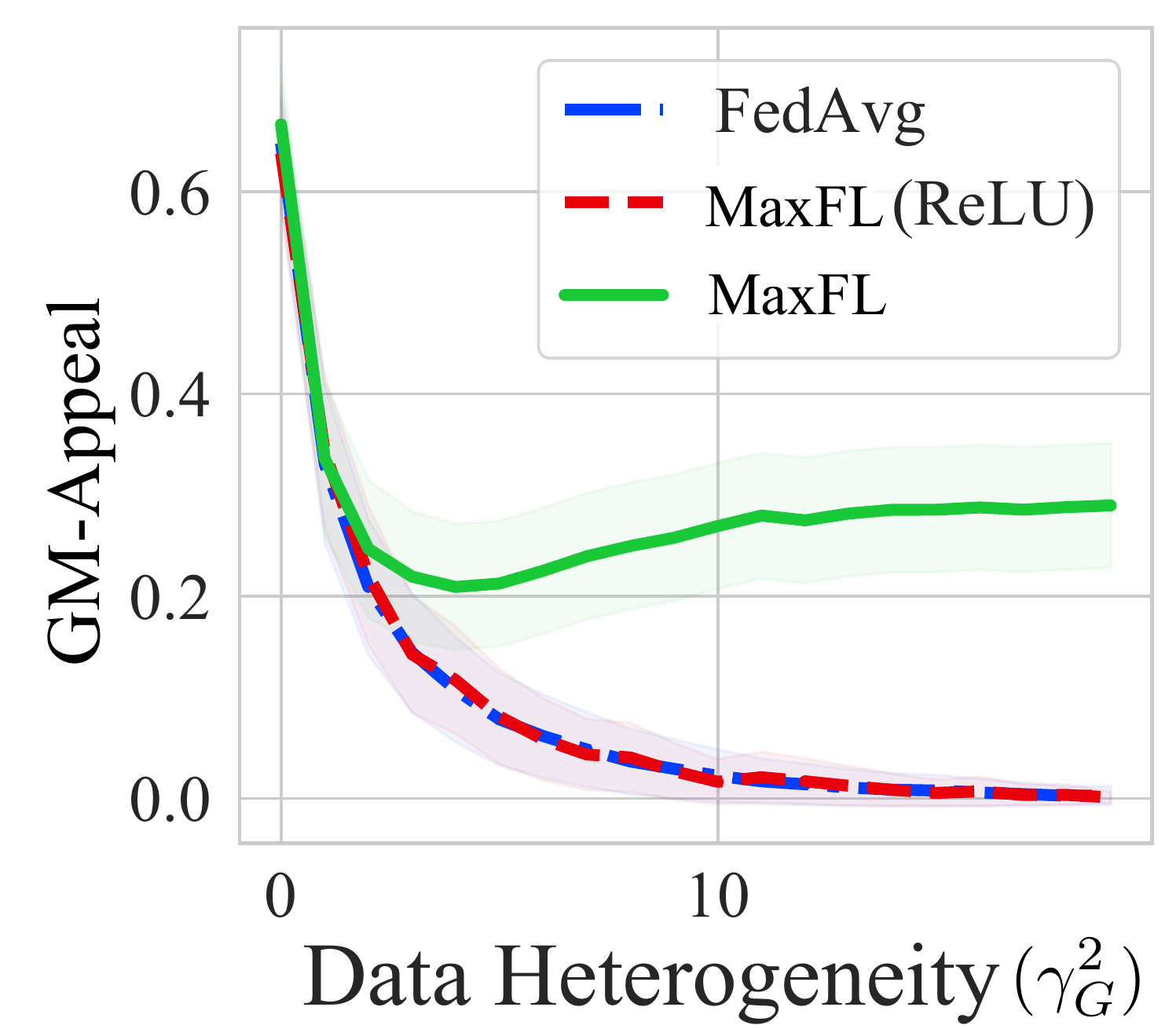}\vspace{-0.5em} \caption{}
\label{fig-1a}
\end{subfigure} 
\begin{subfigure}{0.55\textwidth}
\centering
\includegraphics[width=1\textwidth]{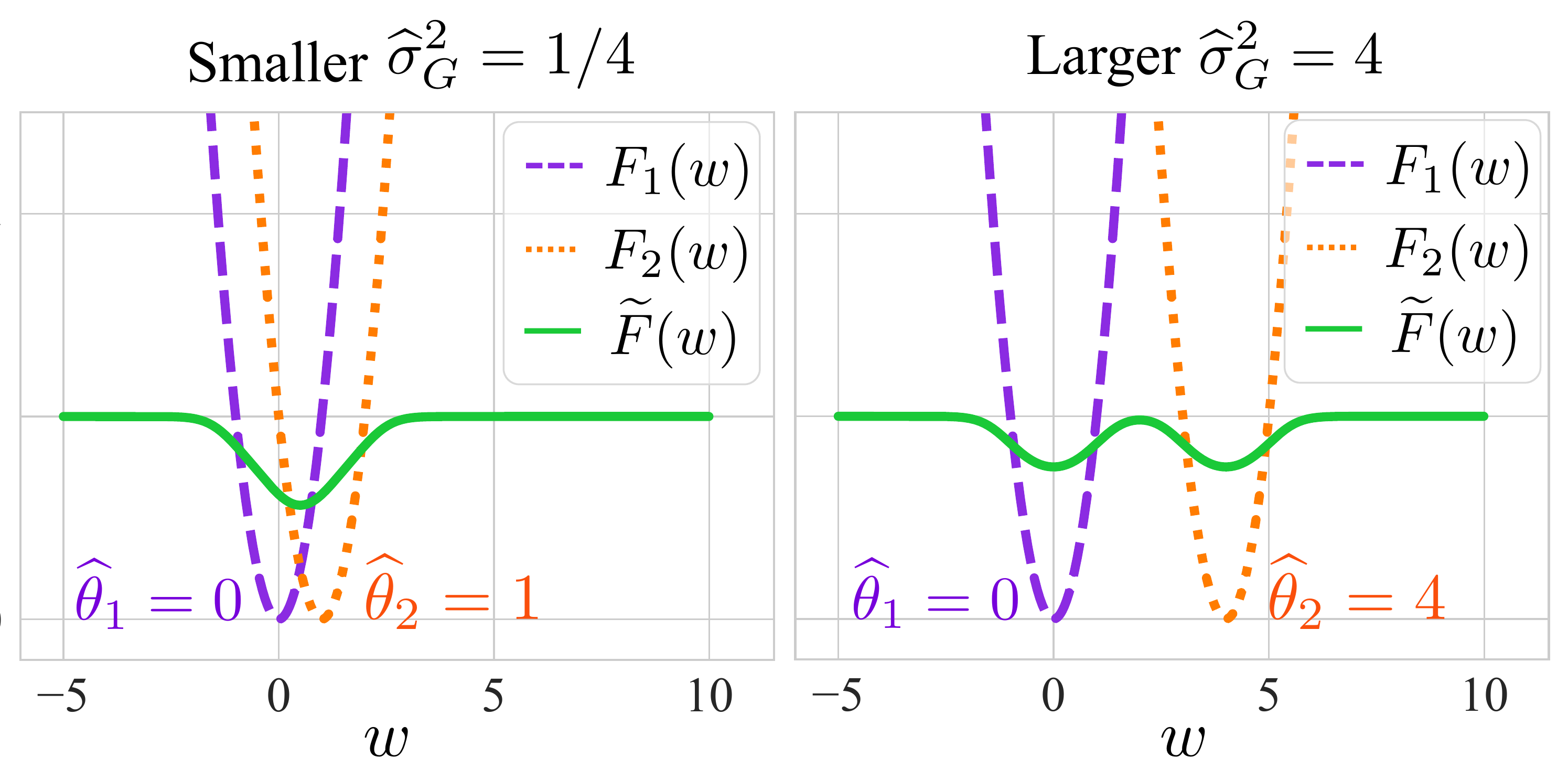}\vspace{-0.2em} \caption{}
\end{subfigure} \vspace{-0.5em}
 \caption{Results for the two client mean estimation; (a): \text{GM-Appeal} for FedAvg decays exponentially while \text{GM-Appeal} for \incfl is lower bounded by a constant. Replacing the sigmoid approximation with ReLU approximation in \incfl leads to the same solution as FedAvg; (b): \incfl adapts to the heterogeneity of the problem---for small heterogeneity it encourages collaboration by having a single global minima, for large heterogeneity it encourages separation by having far away local minimas.} \label{fig:mean}
\end{figure*}
 
\textbf{Maximizing the \incfl Objective Leads to Increased \text{GM-Appeal}.} Based on \Cref{the-mean}, we see that we need nonconvexity in $h(\cdot)$ for the objective to behave differently than standard FL. We set $h(x) = \sigma(x)= \frac{\exp(x)}{1+\exp(x)}$, as proposed in our \incfl objective in \eqref{eq:prop_obj}. We find that the \incfl  objective \textit{adapts} to the empirical heterogeneity parameter $\widehat{\gamma}_G^2 = \left(\frac{\widehat{\theta}_2 -\widehat{\theta}_1}{2} \right)^2$.
If $\widehat{\gamma}_G^2 < 1$ (small data heterogeneity), the objective encourages collaboration by setting the global model to be the average of the local models. 
On the other hand, if $\widehat{\gamma}_G^2 > 2$ (large data heterogeneity), the objective encourages \textit{separation} by setting the global model close to either the local model of the first client or the local model of the second client (see \Cref{fig:mean}). 
Based on this observation, we have the following theorem.
\begin{theorem}
Let $w$ be a local minima of the \incfl objective. The expected \text{GM-Appeal} using $w$ is lower bounded by $\frac{1}{16}\exp\left(-\frac{1}{\gamma^2}\right)$where the expectation is over the randomness in the local dataset $\mathcal{B}_1,\mathcal{B}_2$.
\end{theorem}

Note that our result above is independent of the heterogeneity parameter $\gamma_G^2$. Therefore even with $\gamma_G^2 \gg 0$, \incfl will keep incentivizing atleast one client by adapting its objective accordingly. Additional discussion and proof details can be found in \Cref{app:theo_mean_est_prof}. 

\begin{figure*}[!t]
\centering
\begin{subfigure}{0.3\textwidth}
\centering
\includegraphics[width=1\textwidth]{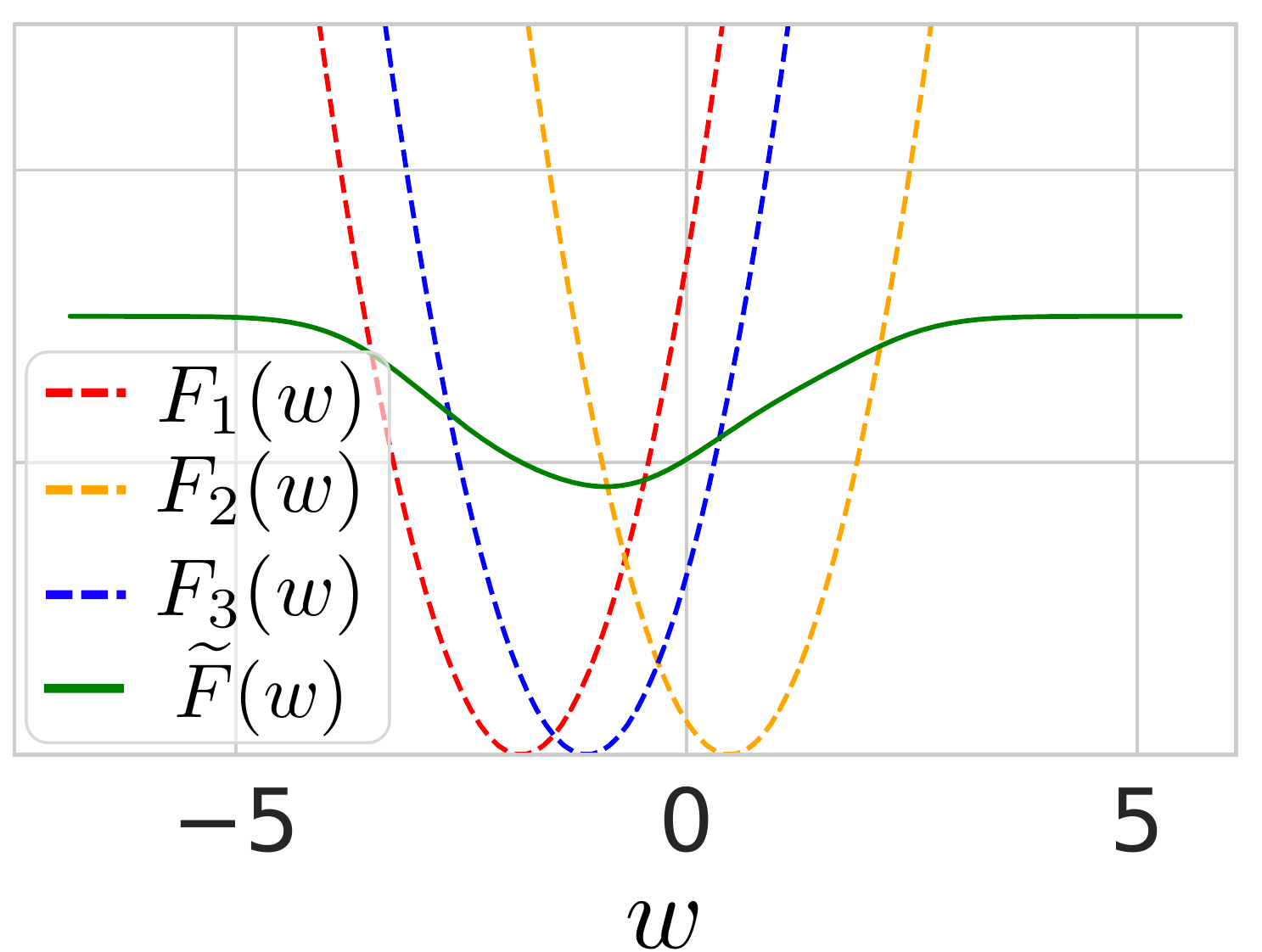}\vspace{-0.5em} \caption{}
\label{fig-11a}
\end{subfigure} 
\begin{subfigure}{0.3\textwidth}
\centering
\includegraphics[width=1\textwidth]{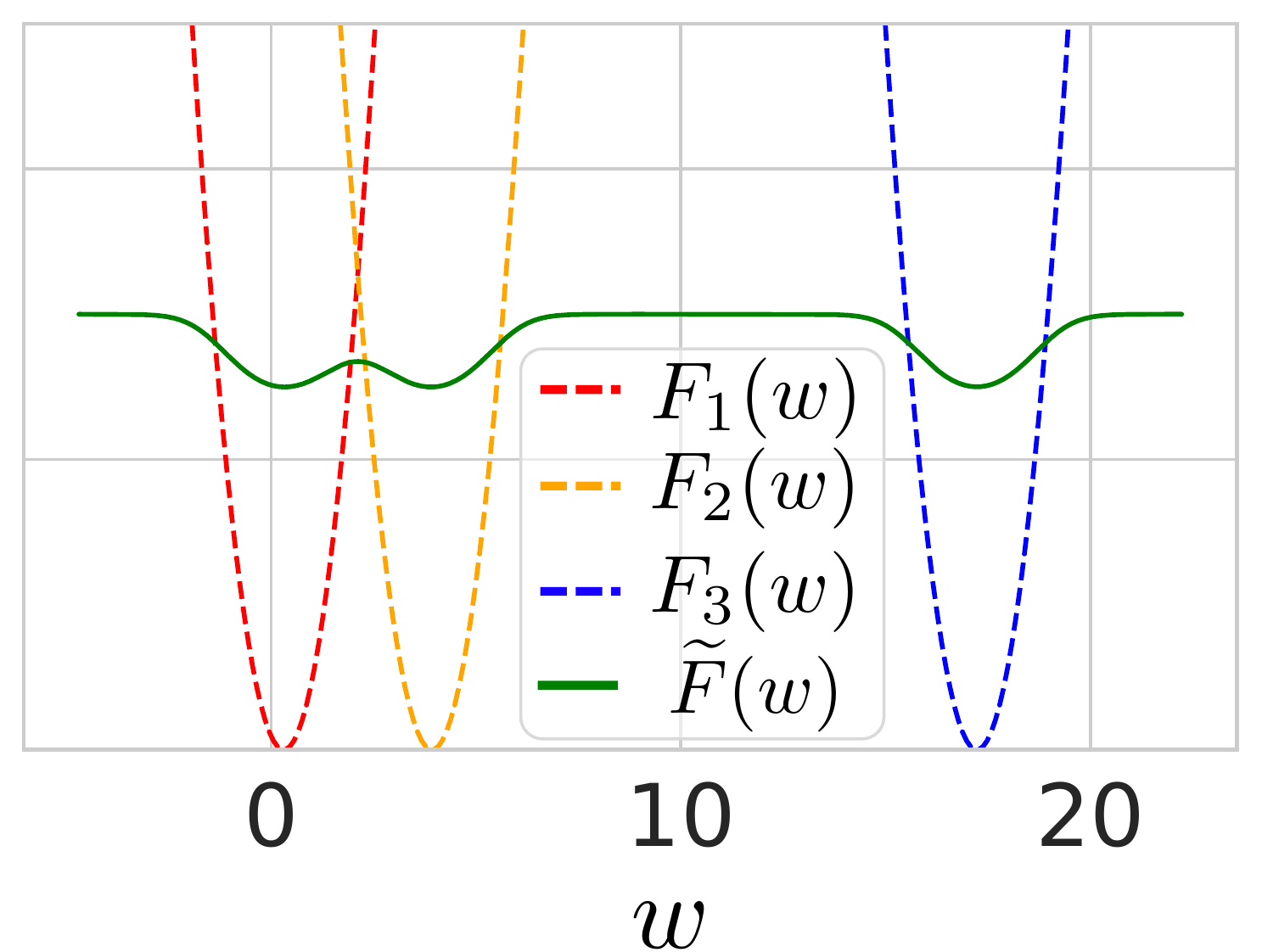}\vspace{-0.5em} \caption{}
\label{fig-12a}
\end{subfigure} 
\begin{subfigure}{0.3\textwidth}
\centering
\includegraphics[width=1\textwidth]{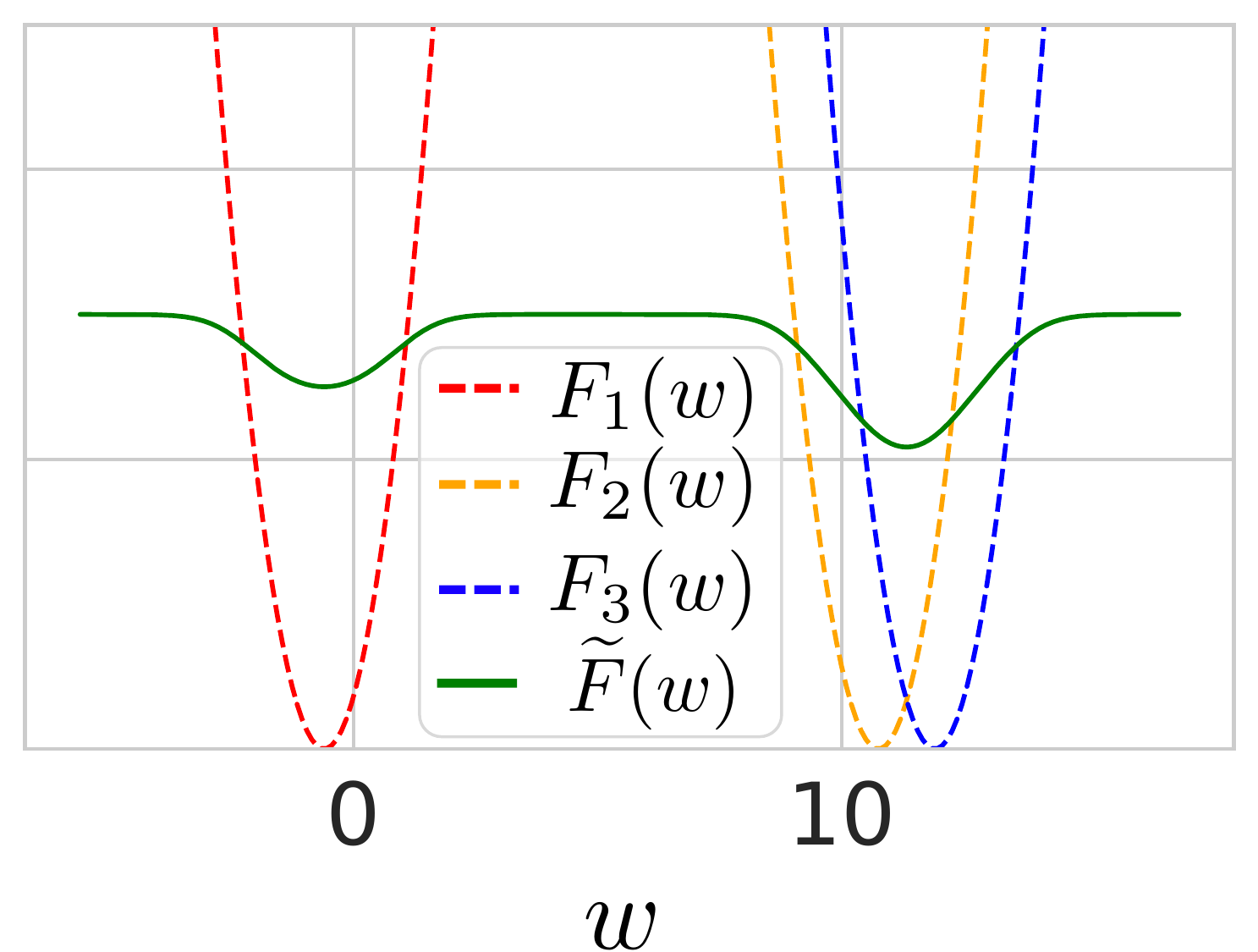}\vspace{-0.5em} \caption{}
\label{fig-13a}
\end{subfigure} 
 \caption{Results for the three client mean estimation; (a): case 1 when the true mean across clients are close to amongst each other where \incfl's optimal solution is identical to that of FedAvg; (b): case 2 when the true mean across clients are all different from each other where \incfl's optimal solution ensures that at least one of the clients will be satisfied with \incfl's global model (unlike FedAvg); (c) case 3 when two clients' true means are close to each other while the other client has a different mean. \incfl in this case, is able to ensure that the two clients satisfied while FedAvg is not able to make any client satisfied.} \label{fig:mean2} 
\end{figure*}

\textbf{Mean Estimation with 3 Clients with \incfl.}
We further examine the property of \incfl to satisfy clients with a 3 clients toy example which is an extension from what we have shown for 2 clients. Reusing the notation from the 2 client example, where $\theta_i$ is the true mean at client $i$ and $\hat{\theta}_i \sim \mathcal{N}(\theta_i,1)$ is the empirical mean of a client, our analysis can be divided into the following cases for the 3 client example (see \Cref{fig:mean2}):
\begin{itemize}[leftmargin=0.35cm]
\item Case 1: $\theta_1 \approx \theta_2 \approx \theta_3$: 
This case captures the setting where the data at the clients is almost i.i.d. In this case, it makes sense for clients to collaborate together and therefore \incfl's optimal solution will be the average of local empirical means (same as FedAvg).

\item Case 2: $\theta_1  \neq \theta_2 \neq \theta_3$: This case captures the setting where the data at clients is completely disparate. In this case, none of the clients benefit from collaborating and therefore \incfl's optimal solution will be the local model of one of the clients. This ensures at least one of the clients will still be satisfied with the \incfl global model unlike FedAvg.

\item Case 3: $\theta_1 \approx \theta_2 \neq \theta_3$: The most interesting case happens when data at two of the clients is similar but the data at the third client is different. Without loss of generality we assume that data at clients 1 and 2 is similar and client 3 is different. In this case, although client 1 and 2 benefit from federating, FedAvg is unable to leverage that due to the heterogeneity at client 3. \incfl, on the other hand, will set the optimal solution to be the average of the local models of just client 1 and client 2. This ensures clients 1 and 2 are satisfied with the global model, thus maximizing the \vpr. 
\end{itemize}

The behavior of \incfl in the three client setup clearly highlights the non-trivialness of our proposed \incfl's formulation.

\subsection{Proof for Theoretical Analysis in \Cref{app:theo_mean_est}} \label{app:theo_mean_est_prof}
Recall the setup discussed in \Cref{app:theo_mean_est}. 
We additionally define the following quantities
\begin{align}
    \gamma^2 \defeq \frac{\nu^2}{N} ; \hspace{10pt} \gamma^2_G = \left(\frac{\theta_2 - \theta_1}{2} \right)^2;
\end{align}

Note that the distribution of the empirical means itself follows a normal distribution following the linear additivity of independent normal random variables. 
\begin{align}
    \widehat{\theta}_1 \sim \mathcal{N} (\theta_1,\gamma^2); \hspace{10pt} \widehat{\theta}_2 \sim \mathcal{N}(\theta_2, \gamma^2) \label{mean_est_fact_1}
\end{align}

\textbf{Lemma A.1 }
\textit{The expected \text{GM-Appeal} of the standard FL model is upper bounded by $2\exp\left(-\frac{\gamma_G^2}{5\gamma^2} \right)$, where the expectation is taken over the randomness in the local datasets $\mathcal{B}_1,\mathcal{B}_2$.}

\textbf{Proof.}

The standard FL model is given by,
\begin{align}
    w = \frac{\widehat{\theta}_1 + \widehat{\theta}_2}{2}
\end{align}

Therefore the expected \text{GM-Appeal} is,

\begin{align}
    &\E{\frac{\mathbb{I}\{ (w-\theta_1)^2 < (\htheta_1-\theta_1)^2\} + \mathbb{I}\{ (w-\theta_2)^2 < (\htheta_2-\theta_2)^2\}}{2}}\\
    & = \frac{1}{2}\left[ \underbrace{\Prob{(w-\theta_1)^2 <(\htheta_1-\theta_1)^2}}_{T_1} + \underbrace{\Prob{(w-\theta_2)^2 <(\htheta_2-\theta_2)^2}}_{T_2}  \right]
\end{align}

Next we bound $T_1$ and $T_2$. 
\begin{align}
    T_1 &= \Prob{(w-\theta_1)^2 <(\htheta_1-\theta_1)^2}\\
    & = \Prob{\brac{\frac{\htheta_1 + \htheta_2}{2}-\theta_1}^2 < (\htheta_1-\theta_1)^2}\\
    & = \Prob{\brac{\frac{\htheta_2-\htheta_1}{2}}^2 + 2\brac{\frac{\htheta_2-\htheta_1}{2}}(\htheta_1-\theta_1) < 0}\\
    & = \Prob{\left\{\brac{\frac{\htheta_2-\htheta_1}{2}}^2 + 2\brac{\frac{\htheta_2-\htheta_1}{2}}(\htheta_1-\theta_1) < 0\right\} \cap \left\{\htheta_2 > \htheta_1 \right\}} \nonumber\\
    & \hspace{10pt} + \Prob{\left\{\brac{\frac{\htheta_2-\htheta_1}{2}}^2 + 2\brac{\frac{\htheta_2-\htheta_1}{2}}(\htheta_1-\theta_1) < 0\right\} \cap \left\{\htheta_2 \leq \htheta_1 \right\}} \label{lem-3-1-1}\\
    & = \Prob{\left\{\brac{\frac{\htheta_2-\htheta_1}{2}} + 2(\htheta_1-\theta_1) < 0\right\} \cap \left\{\htheta_2 > \htheta_1 \right\}} \nonumber\\
    & \hspace{10pt} + \Prob{\left\{\brac{\frac{\htheta_2-\htheta_1}{2}}^2 + 2\brac{\frac{\htheta_2-\htheta_1}{2}}(\htheta_1-\theta_1) < 0\right\} \cap \left\{\htheta_2 \leq \htheta_1 \right\}}
\end{align}
\begin{align}
    & \leq \Prob{\brac{\frac{\htheta_2-\htheta_1}{2}} + 2(\htheta_1-\theta_1) < 0} + \Prob{\htheta_2-\htheta_1 \leq 0} \label{lem-3.1-2}\\
    & = \Prob{Z_1 < 0} + \Prob{Z_2 \leq 0} \hspace{5pt} \text{ where } Z_1 \sim \mathcal{N}\left(\gamma_G,\frac{5}{2}\gamma^2\right) ,Z_2 \sim \mathcal{N}\left(2\gamma_G,2\gamma^2\right) \label{lem-3.1-3}\\
    & \leq \exp\brac{-\frac{\gamma_G^2}{5\gamma^2}} + \exp\brac{-\frac{\gamma_G^2}{\gamma^2}}\label{lem-3.1-4}\\
    & \leq 2\exp\brac{-\frac{\gamma_G^2}{5\gamma^2}}
\end{align}
where \cref{lem-3-1-1} uses $\Prob{A} = \Prob{A \cap B} + \Prob{A \cap B^{\complement}}$, \cref{lem-3.1-2} uses $\Prob{A \cap B} \leq \Prob{A}$, \cref{lem-3.1-3} uses \cref{mean_est_fact_1} and linear additivity of independent normal random variables, \cref{lem-3.1-4} uses a Chernoff bound. 

We can similarly bound $T_2$ to get $T_2 \leq 2\exp\brac{-\frac{\gamma_G^2}{5\gamma^2}}$. Thus the expected \text{GM-Appeal} of the standard FL model is upper bounded by $2\exp\brac{-\frac{\gamma_G^2}{5\gamma^2}}$.

\textbf{Lemma A.2 }
\textit{Let $h$ be any function that is convex, twice differentiable, and strictly increasing in $[0,\infty)$. Then our relaxed objective is strictly convex and has a unique minimizer at $w^* = \left(\frac{\widehat{\theta}_1 + \widehat{\theta}_2}{2}\right)$.}

\textbf{Proof.}

Let us denote our relaxed objective by $v(w)$. Then $v(w)$ can be written as,

\begin{align}
    v(w) & = \frac{1}{2}\left[h\left(F_1(w)-F(\widehat{w}_1)\right) +  h\left(F_2(w)-F(\widehat{w}_2)\right)\right]= \underbrace{\frac{1}{2}h\left((w-\htheta_1)^2\right)}_{v_1(w)} +  \underbrace{\frac{1}{2}h\left((w-\htheta_2)^2\right)}_{v_2(w)}\\
\end{align}

We first prove that $v_1(w)$ is strictly convex. Let $\lambda \in (0,1)$ and $(w_1 ,w_2)$ be any pair of points in $\mathbb{R}^2$ such that $w_1 \neq w_2$. We have,
\begin{align}
    v_1(\lambda w_1 + (1-\lambda)w_2) &= \frac{1}{2}h\left((\lambda (w_1-\htheta_1) + (1-\lambda)(w_2-\htheta_1))^2\right) \\
    & < \frac{1}{2} h\left(\lambda(w_1-\htheta_1)^2 + (1-\lambda)(w_2-\htheta_1)^2\right) \label{lem-3.2-1}\\
    & \leq  \frac{\lambda}{2} h\left((w_1-\htheta_1)^2\right)+ \frac{1-\lambda}{2}h\left((w_2-\htheta_1)^2\right) \label{lem-3.2-2}\\
    & = \lambda v_1(w_1) + (1-\lambda)v_1(w_2)
\end{align}
where \cref{lem-3.2-1} follows from the strict convexity of $f(w)=w^2$ and the fact that $h(w)$ is strictly increasing in the range $[0,\infty)$, \cref{lem-3.2-2} follows from the convexity of $h(w)$.

This completes the proof that $v_1(w)$ is strictly convex. We can similarly prove that $v_2(w)$ is stricly convex and hence $v(w)$ is strictly convex since summation of strictly convex functions is strictly convex. 

Also note that,
\begin{align}
    \nabla v(w) = \nabla h\left( (w-\htheta_1)^2\right)(w-\htheta_1) + \nabla h\left( (w-\htheta_2)^2\right)(w-\htheta_2) 
\end{align}
It is easy to see that $\nabla v(w)= 0$ at $w = \left(\frac{\widehat{\theta}_1 + \widehat{\theta}_2}{2}\right) $. Since $v(w)$ is strictly convex this implies that $w^*= \left(\frac{\widehat{\theta}_1 + \widehat{\theta}_2}{2}\right)$ will be a unique global minimizer. This completes the proof. 

\textbf{Proof of Theorem A.1}

Before stating the proof of Theorem A.1 we first state some intermediate results that will be used in the proof. 

The \incfl objective can be written as,
\begin{align}
    v(w) = \frac{1}{2} \sigma\brac{(w-\htheta_1)^2} +\frac{1}{2} \sigma\brac{(w-\htheta_2)^2}
\end{align}
where $\sigma(w) = 1/(1+\exp(-w))$.

We additionally define the following quantities,
\begin{align}
    i \defeq \argmin \left\{\htheta_1,\htheta_2\right\}; \hspace{5pt} j \defeq \argmax \left\{\htheta_1,\htheta_2\right\}; \hspace{5pt} \hsg \defeq \frac{\htheta_j-\htheta_i}{2}
\end{align}

Let $q(w) = \sigma(w)(1-\sigma(w))$. The gradient of $v(w)$ is given as,
\begin{align}
    \nabla v(w) = q\left((w-\htheta_1)^2 \right)(w-\htheta_1) + q\left((w-\htheta_2)^2 \right)(w-\htheta_2)
\end{align}

\textbf{Lemma A.3 }
\label{lem-3.3}
\textit{For $\hsg > 2$, $w = \left(\frac{\widehat{\theta}_1 + \widehat{\theta}_2}{2}\right)$ will be a local maxima of the \incfl objective.}

It is easy to see that $w = \left(\frac{\widehat{\theta}_1 + \widehat{\theta}_2}{2}\right)$ will always be a stationary point of $\nabla v(w)$. Our goal is to determine whether it will be a local minima or a local maxima. To do so, we calculate the hessian of $v(w)$ as follows. Let $f(w) = 2\sigma(w)(1-\sigma(w))(1-2\sigma(w))$. Then,
\begin{align}
    \nabla^2 v(w) = \underbrace{f\left((w-\htheta_1)^2 \right)(w-\htheta_1)^2+  q\left((w-\htheta_1)^2 \right)}_{h_1(w)}+ \underbrace{f\left((w-\htheta_2)^2 \right)(w-\htheta_2)^2+  q\left((w-\htheta_2)^2 \right)}_{h_2(w)}
\end{align}

Note that $h_1(w) = h_2(w)$ for $w = \left(\frac{\widehat{\theta}_1 + \widehat{\theta}_2}{2}\right)$. Hence it suffices to focus on the condition for which $h_1(w) < 0$ at $w = \left(\frac{\widehat{\theta}_1 + \widehat{\theta}_2}{2}\right)$. We have,

\begin{align}
    h_1\left((\htheta_1 + \htheta_2)/2\right) &=  f(\hsg^2)\hsg^2 + q(\hsg^2)\\
    & = q(\hsg^2)( 2(1-2\sigma(\hsg^2))\hsg^2 + 1)\\
    & <0 \hspace{5pt} \text{ for } \hsg \geq 1.022
\end{align}
where the last inequality follows from the fact that $q(w) > 0$ for all $w \in \mathbb{R}$ and $2(1-2\sigma(w^2))w^2 + 1 < 0$ for $w \geq 1.022$. Thus for $\hsg > 2$, $w = \left(\frac{\widehat{\theta}_1 + \widehat{\theta}_2}{2}\right)$ will be a local maxima of the \incfl objective.


\textbf{Lemma A.4 }
\textit{For $\hsg > 0$, any local minima of $v(w)$ lies in the range $(\htheta_i,\htheta_i+2] \cup [\htheta_j-2,\htheta_j)$.}

Firstly note that since $\hsg > 0$ we have $\htheta_j > \htheta_i$.
Secondly note that since $q(w) > 0 $ for all $w \in \mathbb{R}$, $\nabla v(w) < 0 $ for all $w \leq \htheta_i$ and $\nabla v(w) > 0$ for all $w \geq \htheta_j$. Therefore any root of the function $\nabla  v(w)$ must lie in the range $(\htheta_i,\htheta_j)$. 

\textbf{Case 1: $0 < \hsg \leq 2$}.

In this case, the lemma is trivially satisified since $(\htheta_i,\htheta_j) \subset \left\{(\htheta_i,\htheta_i+2] \cup [\htheta_j-2,\htheta_j)\right\}$.

\textbf{Case 2: $\hsg > 2$}.

Let $x = w - \htheta_i$ and $g(x) = q(x^2)x$. We can write $\nabla v(w)$ as,
\begin{align}
    \nabla v(\htheta_i+x) = g(x) - g(2\hsg-x)
\end{align}

It can be seen that for $x > 2$, $g(x)$ is a decreasing function. For $x \in (2,\hsg)$ we have $x > 2\hsg-x$ which implies $g(x) > g(2\hsg-x)$. Therefore $\nabla v(\htheta_i+x) > 0$ for $x \in (2,\hsg)$. Also $\nabla v(\htheta_i + 2\hsg-x) = - \nabla v(\htheta_i+x)$ and therefore $\nabla v(\htheta_i+x) < 0$ for $x \in (\hsg, 2\hsg-2)$. $\nabla v(\htheta_i+\hsg) = 0$ but this will be a local maxima for $\hsg > 2$ as shown in Lemma A.3. Thus there exists no local minima of  $v(w)$ for $w \in (\htheta_i+2,\htheta_j-2)$

Combining both cases we see that any local minima of $v(w)$ lies in the range $\left\{(\htheta_i,\htheta_i+2] \cup [\htheta_j-2,\htheta_j)\right\}$.

\textbf{Theorem A.1 }
\textit{Let $w$ be a local minima of the \incfl objective. The expected \text{GM-Appeal} using $w$ is lower bounded by $\frac{1}{16}\exp\left(-\frac{1}{\gamma^2}\right)$ where the expectation is over the randomness in the local dataset $\mathcal{B}_1,\mathcal{B}_2$.}

\textbf{Proof.}

The \text{GM-Appeal} can be written as,
\begin{align}
  \frac{1}{2}\left[ \Prob{(w-\theta_i)^2 <(\htheta_i-\theta_i)^2} + \Prob{(w-\theta_j)^2 <(\htheta_j-\theta_j)^2}  \right]  
\end{align}

We focus on the case where $\htheta_2 \neq \htheta_i $ implying $\htheta_j > \htheta_i$ ($\htheta_2 = \htheta_1$ is a zero-probability event and does not affect our proof). Let $w$ be any local minima of the \incfl objective. From Lemma A.4 we know that $w$ will lie in the range $(\htheta_i,\htheta_i+2] \cup [\htheta_j-2,\htheta_j)$

\textbf{Case 1: } $w \in (\htheta_i,\htheta_i+2]$
\begin{align}
    \Prob{(w-\theta_i)^2 <(\htheta_i-\theta_i)^2} &= \Prob{(w-\htheta_i)^2+2(w-\htheta_i)(\htheta_i-\theta_i)<0}\\
    &=  \Prob{(w-\htheta_i)+2(\htheta_i-\theta_i)<0} \label{thm-3.1-0}\\
    & \geq \Prob{2+2(\htheta_i-\theta_i)<0} \label{thm-3.1-1}\\
    & = \Prob{(\htheta_i-\theta_i)<-1}\\
    & \geq \Prob{ \left\{\htheta_1 < \htheta_2\right\} \cap \left\{(\htheta_1-\theta_1)<-1\right\} } \label{thm-3.1-2}\\
    & = \Prob{\htheta_1 < \htheta_2}\Prob{\htheta_1-\theta_1 < -1|\htheta_1 < \htheta_2}\\
    & \geq \Prob{\htheta_1 < \htheta_2}\Prob{\htheta_1-\theta_1 < -1} \label{thm-3.1-3}\\
    & = \Prob{\htheta_1 < \htheta_2}\Prob{Z > 1/\gamma} \hspace{5pt} \text{ where } Z \sim \mathcal{N}(0,1) \label{thm-3.1-4}\\
    & \geq \frac{1}{8}\exp\left(-\frac{1}{\gamma^2}\right) \label{thm-3.1-5}
\end{align}
\cref{thm-3.1-0} uses the fact that $(w-\htheta_i) > 0$, \cref{thm-3.1-1} uses $(w-\htheta_i) \leq 2$, \cref{thm-3.1-2} uses $\Prob{A} \geq \Prob{A \cap B}$ and definition of $i$. \cref{thm-3.1-3} uses the following argument. If $\theta_1 - 1 \geq \htheta_2$ then $\Prob{\htheta_1-\theta_1 < -1|\htheta_1 < \htheta_2} = 1$. If $\theta_1 - 1 < \htheta_2$ then $\Prob{\htheta_1-\theta_1 < -1|\htheta_1 < \htheta_2} = \Prob{\htheta_1-\theta_1 < -1}/\Prob{\htheta_1 < \htheta_2} \geq \Prob{\htheta_1-\theta_1 < -1}$. \cref{thm-3.1-4} uses $\htheta_1-\theta_1 \sim \mathcal{N}(0,\gamma^2)$, \cref{thm-3.1-5} uses $\Prob{\htheta_1 < \htheta_2} \geq \frac{1}{2}$ and $\Prob{Z \geq x} \geq \frac{2\exp(-x^2/2)}{\sqrt{2\pi}(\sqrt{4+x^2}+x)} \geq \frac{1}{4}\exp(-x^2)$ where $Z \sim \mathcal{N}(0,1)$ \cite{komatu1955elementary}.

In the case where $w \in (\htheta_j-2,\htheta_j]$ a similar technique can be used to lower bound $ \Prob{(w-\theta_j)^2 <(\htheta_j-\theta_j)^2} $. Thus the \text{GM-Appeal} of any local minima of the \incfl objective is lower bounded by $\frac{1}{16}\exp\left(-\frac{1}{\gamma^2}\right)$.

\section{Convergence Proof}\label{app:the1proof}
\subsection{Preliminaries}
First, we introduce the key lemmas used for the convergence analysis. 
\begin{lemma}[Bounded Dissimilarity for $\sgf(\wb)$]
With \Cref{as1} and \Cref{as4} we have the bounded dissimilarity with respect to $\sgf(\wb)$ as:
\begin{align}
    \frac{1}{M}\sum_{i=1}^M\|\nabla\widetilde{F}_i(\wb)\|^2\leq\beta'^2\|\nabla \widetilde{F}(\wb)\|^2+\kappa'^2
\end{align}
where $\beta'^2=2\beta^2,~\kappa'^2=4\beta^2 L_c^2+\kappa^2$\label{lem1}
\begin{proof}  
One can easily show that
\begin{align}
   \frac{1}{M}\sum_{i=1}^M\|\nabla\widetilde{F}_i(\wb)\|^2=\frac{1}{M}\sum_{i=1}^Mq_i(\wb)^2\|\nabla{F}_i(\wb)\|^2 \leq\frac{1}{M}\sum_{i=1}^M\|\nabla F_i(\wb)\|^2
\end{align}
due to $q_i(\wb)\leq1$. Hence we have from \Cref{as4} and Cauchy-Schwarz inequality that
\begin{align}
    \frac{1}{M}\sum_{i=1}^M\|\nabla\widetilde{F}_i(\wb)\|^2\leq \frac{1}{M}\sum_{i=1}^M\|\nabla F_i(\wb)\|^2\\
    \leq\beta^2\|\nabla F(\wb)-\nabla\sgf(\wb)+\nabla\sgf(\wb)\|^2+\kappa^2 \\ \leq 2\beta^2\|\nabla F(\wb)-\nabla\sgf(\wb)\|^2+2\beta^2\|\nabla\sgf(\wb)\|^2+\kappa^2 \label{eq-l1-0}
    \end{align}
We bound the first term in \cref{eq-l1-0} as
\begin{align}
    \|\nabla F(\wb)-\nabla\sgf(\wb)\|^2=\left\|\sum_{i=1}^M\frac{(1-q_i(\wb))}{M}\nabla F_i(\wb)\right\|^2\\
    \leq\frac{1}{M}\sum_{i=1}^M\|(1-q_i(\wb))\nabla F_i(\wb)\|^2\\
    \leq\frac{2}{M}\sum_{i=1}^M\|\nabla F_i(\wb)\|^2\leq 2L_c^2 \label{eq-l1-1}
\end{align}
where in \cref{eq-l1-1} we use $q_i(\wb)\leq1,\forall i\in[M]$ and \Cref{as1}.
Then from \cref{eq-l1-0} we have
\begin{align}
  \frac{1}{M}\sum_{i=1}^M\|\nabla\widetilde{F}_i(\wb)\|^2\leq 2\beta^2\|\nabla\sgf(\wb)\|^2+\kappa^2+4\beta^2 L_c^2
\end{align}
completing the proof. \end{proof}
\end{lemma}

\begin{lemma}[Smoothness of $\sgf(\wb)$]
If \Cref{as1} is satisfied we have that the local objectives, $\sgf_1(\wb),~...~,\sgf_M(\wb)$, are also $\widetilde{L}_s$-smooth for any $\wb$ where $\widetilde{L}_s=L_c^2/4+q_i(\bw)L_s$. \label{lem2}
\begin{proof}
Recall the definitions of $\sgf(\wb)$ below:
\begin{align}
    \sgf(\wb)=\frac{1}{M}\sum_{i=1}^M\sgf_i(\wb),~\sgf_i(\wb)\defeq\sigma(F_i(\mathbf{w})- F_i(\widehat{\mathbf{w}}_i^*)) 
\end{align}
Let $\| \hspace{1pt} \|_{op}$ denote the spectral norm of a matrix. Accordingly, with the model parameter vector $\wb\in\mathbb{R}^{d}$, we have the spectral norm of the Hessian of $\sgf_i(\wb),~\forall i\in[M]$ as:
\begin{flalign}
\begin{aligned}
    &\|\nabla^2\sgf_i(\wb)\|_{op}\\&=\|q_i(\wb)[(\nabla F_i(\wb)\nabla F_i(\wb)^{T})(1-q_i(\wb))+\nabla^2F_i(\wb)]\|_{op} \label{eq-l2-0}
    \end{aligned}
\end{flalign}
where $q_i(\wb)=\text{Sigmoid}(F_i(\wb)- F_i(\widehat{\mathbf{w}}_i^*))$ and $\nabla F_i(\wb)\in\mathbb{R}^{d\times1}$ is the gradient vector for the local objective $F_i(\wb)$ and $\nabla^2F_i(\wb)\in\mathbb{R}^{d\times d}$ is the Hessian of $F_i(\wb)$. We can bound the RHS of \cref{eq-l2-0} as follows
\begin{align}
     \|\nabla^2\sgf_i(\wb)\|_{op} = \|q_i(\bw)(1-q_i(\wb))(\nabla F_i(\wb)\nabla F_i(\wb)^{T})+q_i(\bw)\nabla^2F_i(\wb)]\|_{op} \label{eq-l2-2}\\
     \leq \|q_i(\bw)(1-q_i(\wb))(\nabla F_i(\wb)\nabla F_i(\wb)^{T})\|_{op}+\|q_i(\bw)\nabla^2F_i(\wb)\|_{op}
     \label{eq-l2-3} \\
     = q_i(\bw)(1-q_i(\wb))\|(\nabla F_i(\wb)\nabla F_i(\wb)^{T})\|_{op}+q_i(\bw)\|\nabla^2 F_i(\wb)\|_{op} \label{eq-l2-4}\\
     =q_i(\bw)(1-q_i(\wb))\|\nabla F_i(\wb)\|^2+q_i(\bw)\|\nabla^2F_i(\wb)\|_{op} \label{eq-l2-5} \\
     \leq \frac{L_c^2}{4}+q_i(\bw)L_s \label{eq-l2-6}
\end{align}
where we use triangle inequality in \cref{eq-l2-3}, and use $\|\bx\by^{T}\|_{op} = \|\bx\|\|\by\|$ in \cref{eq-l2-5}, and use $q_i(\wb) \leq 1$ along with \Cref{as1} in \cref{eq-l2-6}. Since the norm of the Hessian of $\sgf_i(\wb)$ is bounded by $\frac{L_c^2}{4}+q_i(\bw)L_s$ we complete the proof.
\end{proof} 
\end{lemma}

\comment{
\begin{lemma}[Bounded Variance for Stochastic Gradient of $\widetilde{F}_k(\wb)$] For mini-batches $\xi_k$ and $\widetilde{\xi}_k$ each independently sampled uniformly at random from $\mathcal{B}_k$ from user $k$, the variance of the stochastic gradient of $\widetilde{F}_k(\wb)$ is bounded: $\expt[\|\widetilde{q}_k(\wb_k,\widetilde{\xi}_k)\gb(\wb_k,\xi_k)-q_k(\wb_k)\nabla F_k(\wb_k)\|^2]\leq 2\sigma_g^2+2L_c^2$.
\begin{proof}
We have that
\begin{align}
\begin{aligned}
    \expt[\|\widetilde{q}_k(\wb_k,\widetilde{\xi}_k)\gb(\wb_k,\xi_k)-q_k(\wb_k)\nabla F_k(\wb_k)\|^2]\\
    =\expt[\|\widetilde{q}_k(\wb_k,\widetilde{\xi}_k)\gb(\wb_k,\xi_k)-\widetilde{q}_k(\wb_k,\widetilde{\xi}_k)\nabla F_k(\wb_k)+\widetilde{q}_k(\wb_k,\widetilde{\xi}_k)\nabla F_k(\wb_k)-q_k(\wb_k)\nabla F_k(\wb_k)\|^2]
\end{aligned}\\
\leq 2\expt[\|\widetilde{q}_k(\wb_k,\widetilde{\xi}_k)(\gb(\wb_k,\xi_k)-\nabla F_k(\wb_k))\|^2]+2\expt[\|(\widetilde{q}_k(\wb_k,\widetilde{\xi}_k)-q_k(\wb_k))\nabla F_k(\wb_k)\|^2] \label{eq-l3-1}\\
\leq 2\sigma_g^2+2\expt[(\widetilde{q}_k(\wb_k,\widetilde{\xi}_k)-q_k(\wb_k))^2\|\nabla F_k(\wb_k)\|^2] \label{eq-l3-2} \\
\leq 2\sigma_g^2+2L_c^2 \label{eq-l3-3}
\end{align}
where \cref{eq-l3-1} is due to Cauchy-Schwarz Inqeuality, \cref{eq-l3-2} is due to \Cref{as2}, and \cref{eq-l3-3} is due to \Cref{as1}. 
\end{proof} \label{lem3}
\end{lemma}
}

\subsection{Proof of \Cref{the1} -- Full Client Participation}
\comment{
\YC{[Normalizing Weight Problem:]\\
With the normalizing weights absorbed to the global learning rate we can define the effective update rule for the global model as:
\begin{align}
    \wb^{(t+1,0)}=\wb^{(t,0)}-\eta^{(t,0)}_g\eta_l\sum_{k=1^M}q_k(\wb^{(t,0)})\sum_{r=0}^{\tau-1}\gb(\wb_k^{(t,r)},\xi_k^{(t,r)})
\end{align}
where $\eta^{(t,0)}_g=\frac{\eta_g}{\sum_{k=1}^Mq_k(\wb^{(t,0)})}$ with global learning rate $\eta_g$ and local learning rate $\eta_l$. With such update rule, and $\eta_g,~\eta_l$ are set such that they satisfy the following assumptions
\begin{align}
\frac{\eta_g}{\sum_{k=1}^Mq_k(\wb^{(t,0)})}\leq\min\left\{\eta_{\text{max}},\frac{\sqrt{2}L_s}{2\widetilde{L}_sM}\right\},~\eta_l\leq\min\left\{\frac{1}{2\sqrt{2}\tau L_s},~\frac{1}{4\tau L_s\beta'}\right\}
\end{align}
we can derive the following convergence bound:
\begin{align}
\begin{aligned}
      \frac{1}{T}\sum_{t=0}^{T-1}\expt\left[\|\nabla\sgf(\wb^{(t,0)})\|^2\right]
     \leq\frac{4\left(\sgf(\wb^{(0,0)})-\sgf_{\text{inf}}\right)}{T\eta_g\eta_l\tau}+{4\overline{\eta}_g\eta_l^2\tau\sigma_g^2L_s^2M}+{16\tau^2\eta_l^2\overline{\eta}_gML_s^2\kappa'^2} \nonumber \\
     +4\eta_l\eta_{\text{max}}\widetilde{L}_s\sigma_g^2
     \end{aligned}    
\end{align}
where $\overline{\eta}_g=\eta_{\text{max}}/\eta_g$. By setting $\eta_l=\frac{1}{2\sqrt{2T}\tau L_s}$ we can further optimize the bound to decay to 0 with $T$.}}
For ease of writing, we define the following auxiliary variables for any client $i\in[M]$:
\begin{align}
    \text{Weighted Stochastic Gradient: }\hb_i^{(t,0)}\defeq q_i(\wb^{(t,0)})\sum_{r=0}^{\tau-1}\gb(\wb_i^{(t,r)},\xi_i^{(t,r)}),\\
    \text{Weighted Gradient: }\hbl_i^{(t,0)}\defeq q_i(\wb^{(t,0)})\sum_{r=0}^{\tau-1}\nabla F_i(\wb_i^{(t,r)}),\\
    \text{Normalized Global Learning Rate: }\lrg^{\ti}\defeq \lrg/\left(\sum_{i=1}^Mq_i(\wb^{(t,0)})+\epsilon\right)
\end{align}
where $\epsilon$ is a constant added to the denominator to prevent the denominator from being $0$. From \Cref{algo1} with full client participation, our proposed algorithm has the following effective update rule for the global model at the server:
\begin{align}
    \wb^{(t+1,0)}=\wb^{(t,0)}-\eta^{(t,0)}_g\eta_l\sum_{k=1}^M\hb_k^{(t,0)} \label{eq:update1}
\end{align}
With the update rule in \cref{eq:update1}, defining $\lrt^{(t,0)}\defeq\lrg^{\ti}\lrl\tau\ec$ and using \Cref{lem2} we have
\begin{flalign}
&\begin{aligned}
    \expt\left[\sgf(\wb^{(t+1,0)})\right]-\sgf(\wb^{(t,0)})\leq
    -\lrt^{(t,0)}\expt\left[\left\langle\nabla\sgf(\wb^{(t,0)}),\frac{1}{M\tau}\sum_{i=1}^M\hb_i^{(t,0)}\right\rangle\right]\\+\frac{\widetilde{L}_s(\lrt^{(t,0)})^2}{2}\expt\left[\left\|\frac{1}{M\tau}\sum_{i=1}^M\hb_i^{(t,0)}\right\|^2\right]
\end{aligned}    
\\
&\begin{aligned}
    =-\lrt^{(t,0)}\expt\left[\left\langle\nabla\sgf(\wb^{(t,0)}),\frac{1}{M\tau}\sum_{i=1}^M\left(\hb_i^{(t,0)}-\hbl_i^{(t,0)}\right)\right\rangle\right]-\lrt^{(t,0)}\expt\left[\left\langle\nabla\sgf(\wb^{(t,0)}),\frac{1}{M\tau}\sum_{i=1}^M\hbl_i^{(t,0)}\right\rangle\right]\\
    +\frac{\widetilde{L}_s(\lrt^{(t,0)})^2}{2}\expt\left[\left\|\frac{1}{M\tau}\sum_{i=1}^M\hb_i^{(t,0)}\right\|^2\right]
\end{aligned} 
\\
&\begin{aligned}
= -\frac{\lrt^{(t,0)}}{2}\left\|\nabla\sgf(\wb^{(t,0)})\right\|^2-\frac{\lrt^{(t,0)}}{2}\expt\left[\left\|\frac{1}{M\tau}\sum_{i=1}^M\hbl_i^{(t,0)}\right\|^2\right]+\frac{\lrt^{(t,0)}}{2}\expt\left[\left\|\nabla\sgf(\wb^{(t,0)})-\frac{1}{M\tau}\sum_{i=1}^M\hbl_i^{(t,0)}\right\|^2\right]\\
    +\frac{\widetilde{L}_s(\lrt^{(t,0)})^2}{2M^2\tau^2}\expt\left[\left\|\sum_{i=1}^M\hb_i^{(t,0)}\right\|^2\right] \label{eq-1-0}
    \end{aligned}
\end{flalign}
For the last term in \cref{eq-1-0}, we can bound it as
\begin{flalign}
   &\frac{\widetilde{L}_s(\lrt^{(t,0)})^2}{2M^2\tau^2}\expt\left[\left\|\sum_{i=1}^M\hb_i^{(t,0)}\right\|^2\right]\leq \frac{\widetilde{L}_s(\lrt^{(t,0)})^2}{M^2\tau^2}\sum_{i=1}^M\expt\left[\left\|\hb_i^{(t,0)}-\hbl_i^{(t,0)}\right\|^2\right]+\frac{\widetilde{L}_s(\lrt^{(t,0)})^2}{M^2\tau^2}\expt\left[\left\|\sum_{i=1}^M\hbl_i^{(t,0)}\right\|^2\right] \label{eq-2-0-0}\\
   &=\frac{\widetilde{L}_s(\lrt^{(t,0)})^2}{M^2\tau^2}\sum_{i=1}^M\expt\left[\left\|q_i(\wb^{(t,0)})\sum_{r=0}^{\tau-1}\left(\gb(\wb_i^{(t,r)},\xi_i^{(t,r)})-\nabla F_i(\wb_i^{(t,r)})\right)\right\|^2\right]+\frac{\widetilde{L}_s(\lrt^{(t,0)})^2}{M^2\tau^2}\expt\left[\left\|\sum_{i=1}^M\hbl_i^{(t,0)}\right\|^2\right]
\end{flalign}
\begin{flalign}
   &=\frac{\widetilde{L}_s(\lrt^{(t,0)})^2}{M^2\tau^2}\sum_{i=1}^Mq_i(\wb^{(t,0)})^2\sum_{r=0}^{\tau-1}\expt\left[\left\|\gb(\wb_i^{(t,r)},\xi_i^{(t,r)})-\nabla F_i(\wb_i^{(t,r)})\right\|^2\right]+\frac{\widetilde{L}_s(\lrt^{(t,0)})^2}{M^2\tau^2}\expt\left[\left\|\sum_{i=1}^M\hbl_i^{(t,0)}\right\|^2\right]\\
   &=\frac{\widetilde{L}_s(\lrt^{(t,0)})^2}{M^2\tau^2}\sum_{i=1}^Mq_i(\wb^{(t,0)})^2\tau\sgn^2+\frac{\widetilde{L}_s(\lrt^{(t,0)})^2}{M^2\tau^2}\expt\left[\left\|\sum_{i=1}^M\hbl_i^{(t,0)}\right\|^2\right]\label{eq-2-0}\\
   &\leq\frac{\widetilde{L}_s(\lrt^{(t,0)})^2\sgn^2}{M\tau}+\widetilde{L}_s(\lrt^{(t,0)})^2\expt\left[\left\|\frac{1}{M\tau}\sum_{i=1}^M\hbl_i^{(t,0)}\right\|^2\right] \label{eq-2-0-1}
\end{flalign}
where \cref{eq-2-0-0} is due to the Cauchy-Schwartz inequality and \cref{eq-2-0} is due to \Cref{as2} and \cref{eq-2-0-1} is due to $q_i(\wb)\leq1,\forall i\in[M]$. Merging \cref{eq-2-0-1} into \cref{eq-1-0} we have
\begin{align}
\begin{aligned}
 \expt\left[\sgf(\wb^{(t+1,0)})\right]-\sgf(\wb^{(t,0)})\leq -\frac{\lrt^{(t,0)}}{2}\left\|\nabla\sgf(\wb^{(t,0)})\right\|^2+\frac{\lrt^{(t,0)}}{2}\expt\left[\left\|\nabla\sgf(\wb^{(t,0)})-\frac{1}{M\tau}\sum_{i=1}^M\hbl_i^{(t,0)}\right\|^2\right]\\
    +\frac{\widetilde{L}_s(\lrt^{(t,0)})^2\sgn^2}{M\tau}+\left((\lrt^{(t,0)})^2 \widetilde{L}_s-\frac{\lrt^{(t,0)}}{2}\right)\expt\left[\left\|\frac{1}{M\tau}\sum_{i=1}^M\hbl_i^{(t,0)}\right\|^2\right] \label{eq-2-1}
    \end{aligned}
 \end{align}
Now we aim at bounding the second term in the RHS of \cref{eq-2-1} as follows:
\begin{align}
    \frac{\lrt^{(t,0)}}{2}\expt\left[\left\|\nabla\sgf(\wb^{(t,0)})-\frac{1}{M\tau}\sum_{i=1}^M\hbl_i^{(t,0)}\right\|^2\right]\\
    =\frac{\lrt^{(t,0)}}{2}\expt\left[\left\|\frac{1}{M}\sum_{i=1}^Mq_i(\wb^{(t,0)})\nabla\gf_i(\wb^{(t,0)})-\frac{1}{M\tau}\sum_{i=1}^Mq_i(\wb^{(t,0)})\sum_{r=0}^{\tau-1}\nabla F_i(\wb_i^{(t,r)})\right\|^2\right]\\
    =\frac{\lrt^{(t,0)}}{2}\expt\left[\left\|\frac{1}{M\tau}\sum_{i=1}^Mq_i(\wb^{(t,0)})\sum_{r=0}^{\tau-1}\left(\nabla\gf_i(\wb^{(t,0)})-\nabla F_i(\wb_i^{(t,r)})\right)\right\|^2\right]\\
    \leq\frac{\lrt^{(t,0)}}{2M\tau}\sum_{i=1}^Mq_i(\wb^{(t,0)})^2\sum_{r=0}^{\tau-1}\expt\left[\left\|\nabla\gf_i(\wb^{(t,0)})-\nabla F_i(\wb_i^{(t,r)})\right\|^2\right] \label{eq-3-0}\\
    =\frac{L_s^2\lrt^{(t,0)}}{2M\tau}\sum_{i=1}^Mq_i(\wb^{(t,0)})^2\sum_{r=0}^{\tau-1}\expt\left[\left\|\wb^{(t,0)}-\wb_i^{(t,r)}\right\|^2\right] \label{eq-3-1}
\end{align}
where \cref{eq-3-0} is due to Jensen's inequality and \cref{eq-3-1} is due to \Cref{lem2}. We can bound the difference of the global model and local model for any client $i\in[M]$ as follows:
\begin{align}
    \expt\left[\left\|\wb^{(t,0)}-\wb_i^{(t,r)}\right\|^2\right]=\lrl^2\expt\left[\left\|\sum_{l=0}^{r-1}\gb(\wb_i^{(t,l)},\xi_i^{(t,l)})\right\|^2\right]\\
    \leq2\lrl^2\expt\left[\left\|\sum_{l=0}^{r-1}\gb(\wb_i^{(t,l)},\xi_i^{(t,l)})-\nabla\gf_i(\wb_i^{(t,l)})\right\|^2\right]+2\lrl^2\expt\left[\left\|\sum_{l=0}^{r-1}\nabla\gf_i(\wb_i^{(t,l)})\right\|^2\right]\label{eq-2-4-0}\\
    \leq 2\lrl^2\sgn^2r+2\lrl^2\expt\left[\left\|\sum_{l=0}^{r-1}\nabla F_i(\wb_i^{(t,l)})\right\|^2\right] \label{eq-2-4}
\end{align}
where \cref{eq-2-4-0} is due to Cauchy-Schwarz inequality and \cref{eq-2-4} is due to \Cref{as2}. We bound the last term in \cref{eq-2-4} as follows:
\begin{align}
\expt\left[\left\|\sum_{l=0}^{r-1}\nabla F_i(\wb_i^{(t,l)})\right\|^2\right]\leq r\sum_{l=0}^{r-1}\expt\left[\left\|\nabla F_i(\wb_i^{(t,l)})\right\|^2\right]\leq\tau\sum_{l=0}^{\tau-1}\expt\left[\left\|\nabla F_i(\wb_i^{(t,l)})\right\|^2\right] \label{eq-4-0}\\
\leq 2\tau\sum_{l=0}^{\tau-1}\expt\left[\left\|\nabla F_i(\wb_i^{(t,l)})-\nabla F_i(\wb^{(t,0)})\right\|^2\right]+2\tau^2\expt\left[\left\|\nabla F_i(\wb^{(t,0)})\right\|^2\right] \label{eq-4-1}\\
\leq2L_s^2\tau\sum_{l=0}^{\tau-1}\expt\left[\left\|\wb_i^{(t,l)}-\wb^{(t,0)}\right\|^2\right]+2\tau^2\expt\left[\left\|\nabla F_i(\wb^{(t,0)})\right\|^2\right] \label{eq-4-2}
\end{align}
where \cref{eq-4-0} is due to Jensen's inequality, and \cref{eq-4-1} is due to Cauchy-Schwarz inequality, and \cref{eq-4-2} is due to \Cref{lem2}. Combining \cref{eq-4-2} with \cref{eq-2-4} we have that
\begin{align}
     \expt\left[\left\|\wb^{(t,0)}-\wb_i^{(t,r)}\right\|^2\right]\leq2\lrl^2\sigma_g^2r+4L_s^2\lrl^2\tau\sum_{l=0}^{\tau-1}\expt\left[\left\|\wb^{(t,0)}-\wb_i^{(t,l)}\right\|^2\right]+4\lrl^2\tau^2\expt\left[\left\|\nabla F_i(\wb^{(t,0)})\right\|^2\right] \label{eq-4-2-1}
\end{align}
Reorganizing \cref{eq-4-2-1} and taking the summation $r\in[\tau]$ on both sides we have,
\begin{align}
     (1-4L_s^2\lrl^2\tau^2)\sum_{r=0}^{\tau-1}\expt\left[\left\|\wb^{(t,0)}-\wb_i^{(t,r)}\right\|^2\right]\leq 2\lrl^2\sgn^2\sum_{r=0}^{\tau-1}r+4\lrl^2\tau^3\expt\left[\left\|\nabla F_i(\wb^{(t,0)}\right\|^2\right] \label{eq-4-3}\\
    \leq \lrl^2\sgn^2\tau^2+4\lrl^2\tau^3\expt\left[\left\|\nabla F_i(\wb^{(t,0)}\right\|^2\right] \label{eq-4-4}
\end{align}
With $\lrl\leq1/(2\sqrt{2}\tau L_s)$, we have that $1/(1-4L_s^2\lrl^2\tau^2)\leq 2$ and hence can further bound \cref{eq-4-4} as
\begin{align}
\sum_{r=0}^{\tau-1}\expt\left[\left\|\wb^{(t,0)}-\wb_i^{(t,r)}\right\|^2\right]\leq 2\lrl^2\sgn^2\tau^2+8\lrl^2\tau^3\expt\left[\left\|\nabla F_i(\wb^{(t,0)})\right\|^2\right] \label{eq-4-5}
\end{align}
Finally, plugging in \cref{eq-4-5} to \cref{eq-3-1} we have
\begin{flalign}
&\begin{aligned}
 \frac{\lrt^{(t,0)}}{2}\expt\left[\left\|\nabla\sgf(\wb^{(t,0)})-\frac{1}{M\tau}\sum_{i=1}^M\hbl_i^{(t,0)}\right\|^2\right]\\\leq\frac{L_s^2\lrt^{(t,0)}}{2M\tau}\sum_{i=1}^Mq_i(\wb^{(t,0)})^2\left(2\lrl^2\sgn^2\tau^2+8\lrl^2\tau^3\expt\left[\left\|\nabla F_i(\wb^{(t,0)})\right\|^2\right]\right) \end{aligned}\\
 &\leq L_s^2\lrt^{(t,0)}\lrl^2\sgn^2\tau+4\lrl^2\tau^2L_s^2\lrt^{(t,0)}\frac{1}{M}\sum_{i=1}^M\expt\left[\left\|\nabla F_i(\wb^{(t,0)})\right\|^2\right] \label{eq-4-5-1}\\
 &\leq L_s^2\lrt^{(t,0)}\lrl^2\sgn^2\tau+4\lrl^2\tau^2L_s^2\lrt^{(t,0)}(\beta'^2\left\|\nabla \sgf(\wb^{\ti})\right\|^2+\kappa'^2) \label{eq-4-5-2}
\end{flalign}
where \cref{eq-4-5-1} uses $q_i(\wb)\leq1,\forall i\in[M]$ and \cref{eq-4-5-2} uses \Cref{lem1}. Merging \cref{eq-4-5-2} to \cref{eq-2-1} we have
\begin{align}
\begin{aligned}
    \expt\left[\sgf(\wb^{(t+1,0)})\right]-\sgf(\wb^{(t,0)})\\
    \leq -\frac{\lrt^{(t,0)}}{2}\left\|\nabla\sgf(\wb^{(t,0)})\right\|^2+\lrt^{(t,0)}\left(\lrt^{(t,0)} \widetilde{L}_s-\frac{1}{2}\right)\expt\left[\left\|\frac{1}{M\tau}\sum_{i=1}^M\hbl_i^{(t,0)}\right\|^2\right]\\
    +\frac{\widetilde{L}_s(\lrt^{(t,0)})^2\sgn^2}{M\tau}+\lrt^{(t,0)}L_s^2\lrl^2\sgn^2\tau+4\lrt^{(t,0)}\lrl^2\tau^2L_s^2\beta'^2\left\|\nabla \sgf(\wb^{\ti}\right\|^2+4\lrt^{(t,0)}\lrl^2\tau^2L_s^2\kappa'^2 \end{aligned} \label{eq-4-5-3}
\end{align}
With $\lrl\lrg\leq1/(4\tau L_s)$ we have that $\lrt^{(t,0)} \widetilde{L}_s-\frac{1}{2}\leq-1/4$ and thus can further simplify \cref{eq-4-5-3} to
\begin{align}
\begin{aligned}
    \expt\left[\sgf(\wb^{(t+1,0)})\right]-\sgf(\wb^{(t,0)})\leq -\frac{\lrt^{(t,0)}}{2}\left\|\nabla\sgf(\wb^{(t,0)})\right\|^2+4\lrt^{(t,0)}\lrl^2\tau^2L_s^2\beta'^2\left\|\nabla \sgf(\wb^{\ti})\right\|^2 \\+\frac{\widetilde{L}_s(\lrt^{(t,0)})^2\sgn^2}{M\tau}
    +\lrt^{(t,0)}L_s^2\lrl^2\sgn^2\tau+4\lrt^{(t,0)}\lrl^2\tau^2L_s^2\kappa'^2 \end{aligned} \\
    =\lrt^{(t,0)}\left(4\lrl^2\tau^2L_s^2\beta'-\frac{1}{2}\right)\left\|\nabla\sgf(\wb^{(t,0)})\right\|^2+\frac{\widetilde{L}_s(\lrt^{(t,0)})^2\sgn^2}{M\tau} 
    +\lrt^{(t,0)}L_s^2\lrl^2\sgn^2\tau+4\lrt^{(t,0)}\lrl^2\tau^2L_s^2\kappa'^2 
\end{align}
With local learning rate $ \lrl\leq\min\{1/(4\tau L_s),1/(4\beta'\tau L_s)\}$ we have that
\begin{align}
\begin{aligned}
   \expt\left[\sgf(\wb^{(t+1,0)})\right]-\sgf(\wb^{(t,0)})\leq -\frac{\lrt^{(t,0)}}{4}\left\|\nabla\sgf(\wb^{(t,0)})\right\|^2+\frac{\widetilde{L}_s(\lrt^{(t,0)})^2\sgn^2}{M\tau} 
    +\lrt^{(t,0)}L_s^2\lrl^2\sgn^2\tau\\+4\lrt^{(t,0)}\lrl^2\tau^2L_s^2\kappa'^2
\end{aligned}
\end{align}
and we use the property of $\lrt^{(t,0)}$ that $\frac{M\tau\lrl\lrg}{M+\epsilon}\leq\lrt^{(t,0)}\leq\frac{M\tau\lrl\lrg}{\epsilon}$ to get
\begin{align}
\begin{aligned}
    \expt\left[\sgf(\wb^{(t+1,0)})\right]-\sgf(\wb^{(t,0)})\leq -\frac{M\tau\lrl\lrg}{4(M+\epsilon)}\left\|\nabla\sgf(\wb^{(t,0)})\right\|^2+\frac{\widetilde{L}_sM\tau\lrl^2\lrg^2\sgn^2}{\epsilon^2} 
    \\+\frac{M\tau^2 L_s^2\lrl^3\lrg\sgn^2}{\epsilon}+\frac{4M\lrl^3\lrg\tau^3L_s^2\kappa'^2}{\epsilon} \label{eq-4-5-4}
\end{aligned}
\end{align}
Taking the average across all rounds on both sides of \cref{eq-4-5-4} we get
\begin{align}
\begin{aligned}
     \frac{1}{T}\sum_{t=0}^{T-1}\expt\left[\|\nabla\sgf(\wb^{(t,0)})\|^2\right]
     \leq\frac{4(M+\epsilon)\left(\sgf(\wb^{(0,0)})-\sgf_{\text{inf}}\right)}{M\tau\lrl\lrg T}+\frac{16\lrl^2\tau^2L_s^2\kappa'^2(M+\epsilon)}{\epsilon}\\+\frac{4L_s^2\lrl^2\tau\sgn^2(M+\epsilon)}{\epsilon}+\frac{4\lrg\lrl\widetilde{L}_s\sgn^2(M+\epsilon)}{\epsilon^2}
     \end{aligned}\label{eq-4-5-5}
\end{align}
and prove
\begin{flalign}
\begin{aligned}
    &\min_{t\in[T]}\expt\left[\left\|\nabla\sgf(\wb^{(t,0)})\right\|^2\right]\leq \frac{1}{T}\sum_{t=0}^{T-1}\expt\left[\|\nabla\sgf(\wb^{(t,0)})\|^2\right]\leq\frac{4(M+\epsilon)\left(\sgf(\wb^{(0,0)})-\sgf_{\text{inf}}\right)}{M\tau\lrl\lrg T}
     \\&+\frac{16\lrl^2\tau^2L_s^2\kappa'^2(M+\epsilon)}{\epsilon}+\frac{4L_s^2\lrl^2\tau\sgn^2(M+\epsilon)}{\epsilon}+\frac{4\lrg\lrl\widetilde{L}_s\sgn^2(M+\epsilon)}{\epsilon^2}
\end{aligned}\label{eq-4-5-6}
\end{flalign}
Further, using $\Tilde{L}_s =\frac{L_s}{M}\sum_{k=1}^M q_k(\wb) + \frac{L_c}{4}$ and $\epsilon = \frac{ML_c}{4L_s} > 0$ from the optimal learning rate we have the bound in \cref{eq-4-5-6} to be
\begin{flalign}
\begin{aligned}
    \min_{t\in[T]}\expt\left[\left\|\nabla\sgf(\wb^{(t,0)})\right\|^2\right]&\leq\frac{(4L_s+L_c)\left(\sgf(\wb^{(0,0)})-\sgf_{\text{inf}}\right)}{L_s\tau\lrl\lrg T}
    +\frac{64\lrl^2\tau^2L_s^2\kappa'^2(4L_s+L_c)}{L_c}\\&+\frac{4L_s^2\lrl^2\tau\sgn^2(4L_s+L_c)}{L_c}+\frac{64L_s\lrg\lrl\sgn^2(L_s+L_c/4)^2}{ML_c^2}
\end{aligned}\label{eq-4-5-6-1}
\end{flalign}
By setting the global and local learning rate as $\lrg=\sqrt{\tau M}$ and $\lrl=\frac{1}{\sqrt{T}\tau}$ we can further optimize the bound as 
\begin{align}
\begin{aligned}
     \min_{t\in[T]}\expt\left[\left\|\nabla\sgf(\wb^{(t,0)})\right\|^2\right]&\leq\frac{(4L_s+L_c)\left(\sgf(\wb^{(0,0)})-\sgf_{\text{inf}}\right)}{L_s\sqrt{TM\tau}}+\frac{64L_s^2\kappa'^2(4L_s+L_c)}{L_cT}\\&+\frac{4L_s^2\sgn^2(4L_s+L_c)}{T\tau L_c}+\frac{64L_s\sgn^2(L_s+L_c/4)^2}{\sqrt{TM\tau}}
     \end{aligned}\label{eq-4-5-7}
\end{align}
completing the full client participation proof of \Cref{the1}.

\subsection{Proof of \Cref{the1} -- Partial Client Participation}
We present the convergence guarantees of \incfl~for partical client participation in this section. With partical client participation, we have the update rule in \cref{eq:update1} changed to
\begin{align}
    \wb^{(t+1,0)}=\wb^{(t,0)}-\eta^{(t,0)}_g\eta_l\sum_{k\in{\mathcal{S}^{(t,0)}}}\hb_k^{(t,0)} \label{eq:update-partial}
\end{align}
where the $m$ clients are sampled uniformly at random without replacement for $\set^{(t,0)}$ at each communication round $t$ by the server and $\eta^{(t,0)}_g=m\eta_g/(\sum_{k \in\mathcal{S}^{(t,0)}}q_k(\wb^{(t,0)})+\epsilon)$ for positive constant $\epsilon$. Then with the update rule in \cref{eq:update-partial} and \Cref{lem2}, defining $\widetilde{\lr}^{(t,0)}=\eta_g^{(t,0)}\lrl\tau m$ we have
\begin{flalign}
&\begin{aligned}
    \expt\left[\sgf(\wb^{(t+1,0)})\right]-\sgf(\wb^{(t,0)})\leq
    \expt\left[-\lrt^{(t,0)}\left\langle\nabla\sgf(\wb^{(t,0)}),\frac{1}{m\tau}\sum_{i\in\sett}\hb_i^{(t,0)}\right\rangle\right]\\+\expt\left[\frac{\widetilde{L}_s(\lrt^{(t,0)})^2}{2}\left\|\frac{1}{m\tau}\sum_{i\in\sett}\hb_i^{(t,0)}\right\|^2\right]
\end{aligned}   \label{eq-10-1} 
\end{flalign}
For the first term in the RHS of \cref{eq-10-1} we have that due to the uniform sampling of clients (see Lemma 4 in \cite{div2022parcli}), it becomes analogous to the derivation for full client participation. Hence, with the property of $\frac{m\tau\lrl\lrg}{m+\epsilon}\leq\lrt^{(t,0)}\leq\frac{m\tau\lrl\lrg}{\epsilon}$ and using the previous bounds in \cref{eq-4-5-2}, we result in the final bound for the first term in the RHS of \cref{eq-10-1} as below:
\begin{flalign}
\begin{aligned}
      \expt\left[-\lrt^{(t,0)}\left\langle\nabla\sgf(\wb^{(t,0)}),\frac{1}{m\tau}\sum_{i\in\sett}\hb_i^{(t,0)}\right\rangle\right]\leq \left(-\frac{m\tau\lrl\lrg}{m+\epsilon}+\frac{4\lrl^3\tau^3L_s^2\beta'^2\lrg m}{\epsilon}\right)\left\|\nabla\sgf(\wb^{(t,0)})\right\|^2\\+\frac{4L_s^2\tau^3\lrl^3m\lrg\kappa'^2}{\epsilon}+\frac{L_s^2\tau^2\lrl^2m\lrg\sigma_g^2}{\epsilon} \label{eq-10-1-1-1}
\end{aligned}
\end{flalign}

For the second term in the RHS of \cref{eq-10-1}, with $C=\widetilde{L}_s(m\tau\lrl\lrg/\epsilon)^2$ we have the following:
\begin{flalign}
&\begin{aligned}
    \expt\left[\frac{\widetilde{L}_s(\lrt^{(t,0)})^2}{2}\left\|\frac{1}{m\tau}\sum_{i\in\sett}\hb_i^{(t,0)}\right\|^2\right]\leq C\expt\left[\left\|\frac{1}{m\tau}\sum_{i\in\sett}(\hb_i^{(t,0)}-\overline{\hb}_i^{(t,0)})\right\|^2\right] \\
    +C\expt\left[\left\|\frac{1}{m\tau}\sum_{i\in\sett}\overline{\hb}_i^{(t,0)}\right\|^2\right]
    \end{aligned} \label{eq-10-1-1}\\
    &=\frac{C}{m^2\tau^2}\expt\left[\sum_{i\in\sett}\left\|\hb_i^{(t,0)}-\overline{\hb}_i^{(t,0)}\right\|^2\right]+C\expt\left[\left\|\frac{1}{m\tau}\sum_{i\in\sett}\overline{\hb}_i^{(t,0)}\right\|^2\right]
\\&=\frac{C}{mM\tau^2}\sum_{i=1}^M\expt\left[\left\|\hb_i^{(t,0)}-\overline{\hb}_i^{(t,0)}\right\|^2\right]+C\expt\left[\left\|\frac{1}{m\tau}\sum_{i\in\sett}\overline{\hb}_i^{(t,0)}\right\|^2\right] \label{eq-10-1-2}\\
 &\leq \frac{C\sigma_g^2}{m\tau}+C\expt\left[\left\|\frac{1}{m\tau}\sum_{i\in\sett}\overline{\hb}_i^{(t,0)}\right\|^2\right] \label{eq-10-1-3}
\end{flalign}
where \cref{eq-10-1-2} follows due to, again, the uniform sampling of clients and the rest follows identical steps for full client participation in the derivation for \cref{eq-2-0-0}. Note that
\begin{align}
    C= \left(\frac{L_s}{M}\sum_{k=1}^M q_k(\wb) + \frac{L_c}{4}\right)(m\tau\lrl\lrg/\epsilon)^2
    \leq (L_s+\frac{L_c}{4})(m\tau\lrl\lrg/\epsilon)^2 \label{eq-10-1-3-1}
\end{align}
For the second term in \cref{eq-10-1-3} we have that
\begin{align}
&\begin{aligned}
    \expt\left[\left\|\frac{1}{m\tau}\sum_{i\in\sett}\overline{\hb}_i^{(t,0)}\right\|^2\right]=\expt\left[\left\|\frac{1}{m\tau}\sum_{i\in\sett}\left(\overline{\hb}_i^{(t,0)}-\nabla\sgf_i(\wb^{(t,0)})+\nabla\sgf_i(\wb^{(t,0)})\right)\right.\right.\\\left.\left.
    -\frac{1}{\tau}\nabla\sgf(\wb^{(t,0)})+\frac{1}{\tau}\nabla\sgf(\wb^{(t,0)})\right\|^2\right]
\end{aligned}\\
&\begin{aligned}
    \leq \underbrace{3\expt\left[\left\|\frac{1}{m\tau}\sum_{i\in\sett}\left(\overline{\hb}_i^{(t,0)}-\nabla\sgf_i(\wb^{(t,0)})\right)\right\|^2\right]}_{A_1}+
    \underbrace{\frac{3}{\tau^2}\expt\left[\left\|\frac{1}{m}\sum_{i\in\sett}\nabla\sgf_i(\wb^{(t,0)})-\nabla\sgf(\wb^{(t,0)})\right\|^2\right]}_{A_2}\\+
    3\expt\left[\left\|\frac{1}{\tau}\nabla\sgf(\wb^{(t,0)})\right\|^2\right]
\end{aligned} \label{eq-10-1-4}
\end{align}
First we bound $A_1$ in \cref{eq-10-1-4} as follows:
\begin{flalign}
&\begin{aligned}
    3\expt\left[\left\|\frac{1}{m\tau}\sum_{i\in\sett}\left(\overline{\hb}_i^{(t,0)}-\nabla\sgf_i(\wb^{(t,0)})\right)\right\|^2\right]=  3\expt\left[\left\|\frac{1}{m\tau}\sum_{i\in\sett}q_i(\wb^{(t,0)})\sum_{r=0}^{\tau-1}\left(\nabla F_i(\wb_i^{(t,r)})-\nabla F_i(\wb^{(t,0)})\right)\right\|^2\right]\\
 \leq \frac{3}{m\tau}\expt\left[\sum_{i\in\sett}\sum_{r=0}^{\tau-1}\left\|\nabla F_i(\wb_i^{(t,r)})-\nabla F_i(\wb^{(t,0)})\right\|^2\right]
= \frac{3}{M\tau}\sum_{i=1}^M\sum_{r=0}^{\tau-1}\expt\left[\left\|\nabla F_i(\wb_i^{(t,r)})-\nabla F_i(\wb^{(t,0)})\right\|^2\right]
    \end{aligned} \label{eq-10-1-6}\\  
 &\begin{aligned}
\leq \frac{3L_s^2}{M\tau}\sum_{i=1}^M\sum_{r=0}^{\tau-1}\expt\left[\left\|\wb^{(t,0)}-\wb_i^{(t,r)}\right\|^2\right]
    \end{aligned} \label{eq-10-1-7}
\end{flalign}
where \cref{eq-10-1-6} is due to Jensen's inequality, $q_i(\wb)\leq 1$, and uniform sampling of clients, and \cref{eq-10-1-7} is due to \Cref{as1}. Using \cref{eq-2-1} we have already derived, bound \cref{eq-10-1-7} further to:
\begin{align}
     3\expt\left[\left\|\frac{1}{m\tau}\sum_{i\in\sett}\left(\overline{\hb}_i^{(t,0)}-\nabla\sgf_i(\wb^{(t,0)})\right)\right\|^2\right]\leq {6L_s^2\lrl^2\sigma_g^2\tau}+\frac{24L_s^2\lrl^2\tau^2}{M}\sum_{i=1}^M\expt\left[\left\|\nabla F_i(\wb^{(t,0)})\right\|^2\right]\\
     \leq {6L_s^2\lrl^2\sigma_g^2\tau}+{24L_s^2\lrl^2\tau^2}(\beta'^2\|\nabla \sgf(\wb^{(t,0)}\|^2+\kappa'^2) \label{eq-10-1-8}
\end{align}
where \cref{eq-10-1-8} is due to \Cref{lem1}.

Next we bound $A_2$ as follows:
\begin{flalign}
&\begin{aligned}
  \frac{3}{\tau^2}\expt\left[\left\|\frac{1}{m}\sum_{i\in\sett}\nabla\sgf_i(\wb^{(t,0)})-\nabla\sgf(\wb^{(t,0)})\right\|^2\right]\\=    \frac{3(M-m)}{\tau^2mM(M-1)}\sum_{i=1}^M\expt\left[\left\|\nabla\sgf_i(\wb^{(t,0)})-\nabla\sgf(\wb^{(t,0)})\right\|^2\right] 
  \end{aligned} \label{eq-10-1-9} \\
&\begin{aligned}
  = \frac{3(M-m)}{\tau^2mM(M-1)}\sum_{i=1}^M\left\|\nabla q_i(\wb^{(t,0)})\gf_i(\wb^{(t,0)})-\frac{1}{M}\sum_{i=1}^Mq_i(\wb^{(t,0)})\nabla\gf_i(\wb^{(t,0)})\right\|^2
  \end{aligned} \label{eq-10-2-0} \\
  &\begin{aligned}
  \leq \frac{6(M-m)}{\tau^2mM(M-1)}\sum_{i=1}^M\left(\left\|\nabla\gf_i(\wb^{(t,0)})\right\|^2+
  \left\|\frac{1}{M}\sum_{i=1}^Mq_i(\wb^{(t,0)})\nabla\gf_i(\wb^{(t,0)})\right\|^2\right)
  \end{aligned} \label{eq-10-2-1} \\
  &\begin{aligned}
  \leq \frac{12(M-m)L_c^2}{\tau^2m(M-1)}
  \end{aligned} \label{eq-10-2-2}
\end{flalign}
where \cref{eq-10-1-9} is due to the variance under uniform sampling without replacement (see Lemma 4 in \cite{div2022parcli}) and \cref{eq-10-2-1} is due to the Cauchy-Schwarz inequality and \cref{eq-10-2-2} is due to \Cref{as1}.

Mering the bounds for $A_1$ and $A_2$ to \cref{eq-10-1-4} we have that
\begin{flalign}
&\begin{aligned}
     \expt\left[\left\|\frac{1}{m\tau}\sum_{i\in\sett}\overline{\hb}_i^{(t,0)}\right\|^2\right]\leq {6L_s^2\lrl^2\sigma_g^2\tau}+{24L_s^2\lrl^2\tau^2}\beta'^2\|\nabla \sgf(\wb^{(t,0)})\|^2\\ +24L_s^2\lrl^2\tau^2\kappa'^2+\frac{12(M-m)L_c^2}{\tau^2m(M-1)}+3\expt\left[\left\|\frac{1}{\tau}\nabla\sgf(\wb^{(t,0)})\right\|^2\right]
\end{aligned}\\
&\begin{aligned}
     =\left(24L_s^2\lrl^2\tau^2\beta'^2+\frac{3}{\tau^2}\right)\|\nabla \sgf(\wb^{(t,0)})\|^2+6L_s^2\lrl^2\tau(\sigma_g^2+4\tau\kappa'^2)+\frac{12(M-m)L_c^2}{\tau^2m(M-1)}\label{eq-10-2-3}
\end{aligned}
\end{flalign}
Then we can plug in \cref{eq-10-2-3} back to \cref{eq-10-1-3} and plugging in \cref{eq-10-1-1-1} to \cref{eq-10-1}, we can derive the bound in \cref{eq-10-1} as
\begin{flalign}
\begin{aligned}
    &\expt\left[\sgf(\wb^{(t+1,0)})\right]-\sgf(\wb^{(t,0)})\\
    &\leq
   \left(-\frac{m\tau\lrl\lrg}{m+\epsilon} +\frac{4\lrl^3\lrg\tau^3L_s^2\beta'^2m}{\epsilon}+\nu\left(\tau\lrl\lrg\right)^2(24L_s^2\lrl^2\tau^2\beta'^2+3)\right)\left\|\nabla\sgf(\wb^{(t,0)})\right\|^2
   +\left(\tau\lrl\lrg\right)^2\nu\frac{\sgn^2}{m\tau}\\
   &+\left(\tau\lrl\lrg\right)^2\nu\left(6L_s^2\lrl^2\tau(\sigma_g^2+4\tau\kappa'^2)+\frac{12(M-m)L_c^2}{\tau^2m(M-1)}\right)+\frac{4L_s^2\tau^3\lrl^3m\lrg\kappa'^2}{\epsilon}+\frac{L_s^2\tau^2\lrl^2m\lrg\sigma_g^2}{\epsilon}
\end{aligned}
\end{flalign}
where $\nu=L_s+L_c/4$. With $\lrl\leq 1/4\beta'\tau L_s,~\epsilon=m$, and $\lrg\lrl\leq\frac{1}{9\tau\nu}$, 
we can further bound above as
\begin{flalign}
\begin{aligned}
    &\expt\left[\sgf(\wb^{(t+1,0)})\right]-\sgf(\wb^{(t,0)})\leq
   -\frac{\lrl\lrg\tau}{4}\left\|\nabla\sgf(\wb^{(t,0)})\right\|^2+\left(\tau\lrl\lrg\right)^2\nu\frac{\sgn^2}{m\tau}\\
   &+\left(\tau\lrl\lrg\right)^2\nu\left(6L_s^2\lrl^2\tau(\sigma_g^2+4\tau\kappa'^2)+\frac{12(M-m)L_c^2}{\tau^2m(M-1)}\right)+4L_s^2\tau^3\lrl^3\lrg\kappa'^2+{L_s^2\tau^2\lrl^2\lrg\sigma_g^2}
\end{aligned} \label{eq-10-3-1}
\end{flalign}
Taking the average across all rounds on both sides of \cref{eq-10-3-1} and rearranging the terms we get
\begin{flalign}
\begin{aligned}
 &\frac{1}{T}\sum_{t=0}^{T-1}\expt\left[\|\nabla\sgf(\wb^{(t,0)})\|^2\right]
     \leq \frac{4\left(\sgf(\wb^{(0,0)})-\sgf_{\text{inf}}\right)}{T\lrl\lrg\tau}+4\sgn^2\lrl\left(\frac{\lrg\nu}{m}+\frac{2L_s^2\lrl\tau}{3}+L_s^2\tau\right)\\
     &+\frac{80L_s^2\lrl^2\tau^2\kappa'^2}{3}+\frac{48\lrl\lrg\nu(M-m)L_c^2}{\tau m(M-1)}
\end{aligned}
\end{flalign}
With the small enough learning rate $\lrl=1/(\sqrt{T}\tau)$ and $\lrg=\sqrt{\tau m}$ one can prove that
\begin{flalign}
\begin{aligned}
 \min_{t\in[T]}\expt\left[\left\|\nabla\sgf(\wb^{(t,0)})\right\|^2\right]
     \leq \frac{4\left(\sgf(\wb^{(0,0)})-\sgf_{\text{inf}}\right)+4\sgn^2\nu}{\sqrt{T\tau m}}+
     \frac{4\sgn^2L_s^2}{\sqrt{T}}+\frac{8\sgn^2L_s^2}{3\tau T}\\
     +\frac{80L_s^2\kappa'^2}{T}+\frac{48\nu(M-m)L_c^2\sqrt{\tau}}{\sqrt{T m}}
\end{aligned}\\
\begin{aligned}
= \mathcal{O}\left(\frac{\sigma_g^2}{\sqrt{T\tau m}}\right)+
    \mathcal{O}\left(\frac{\sigma_g^2}{\tau T}\right)+
\mathcal{O}\left(\frac{\kappa'^2}{T}\right)+
\mathcal{O}\left(\frac{\sqrt{\tau}}{\sqrt{Tm}}\right)
\end{aligned}
\end{flalign}
completing the proof for \Cref{the1} for partial client participation. 
\comment{
\subsection{Proof of \Cref{the2}}
We first derive the relation between $F(\wb)$ and $\sgf(\wb)$ and to prove the convergence of our algorithm for $F(\wb)$:\\
\begin{align}
    \nabla F(\wb)-\nabla\sgf(\wb)=\sum_{i=1}^M\frac{1-q_i(\wb)}{M}\nabla F_i(\wb)=\sum_{i=1}^M\frac{1-q_i(\wb)}{\sqrt{M}}\times\frac{\nabla F_i(\wb)}{\sqrt{M}}
\end{align}
With the Cauchy-Schwartz Inequality and \Cref{lem1} we have that
\begin{align}
    \left\|\nabla F(\wb)-\nabla\sgf(\wb)\right\|^2=\left\|\sum_{i=1}^M\frac{1-q_i(\wb)}{\sqrt{M}}\times\frac{\nabla F_i(\wb)}{\sqrt{M}}\right\|^2\leq\left[\sum_{i=1}^M\frac{(1-q_i(\wb))^2}{M}\right]\left[\frac{1}{M}\sum_{i=1}^M\|\nabla F_i(\wb)\|^2\right]\\
    \leq \beta'^2\|\nabla\sgf(\wb)\|^2+\kappa'^2
\end{align}
Note that
\begin{align}
    \|\nabla F(\wb)\|^2\leq 2\left\|\nabla F(\wb)-\nabla\sgf(\wb)\right\|^2+2\|\nabla \sgf(\wb)\|^2\leq2(\beta'^2+1)\|\nabla\sgf(\wb)\|^2+2\kappa'^2
\end{align}
Hence we have that
\begin{align}
    \min_{t\in[T]}\|\nabla F(\wb^{(t,0)})\|^2\leq\frac{1}{T}\sum_{t=0}^{T-1}\|\nabla F(\wb^{(t,0)})\|^2\\
    \leq 2(\beta'^2+1)\frac{1}{T}\sum_{t=0}^{T-1}\|\nabla\sgf(\wb^{(t,0)})\|^2+2\kappa'^2\\
    =2(2\beta^2+1)\frac{1}{T}\sum_{t=0}^{T-1}\|\nabla\sgf(\wb^{(t,0)})\|^2+8\beta^2 L_c^2+2\kappa^2
\end{align}
completing the proof of \Cref{the2}.}

\section{Simulation Details for \Cref{fig-1a}}

For the mean estimation simulation for \Cref{fig:mean}(a), we set the true means for the two clients as $\theta_1=0,~\theta_2=2\gamma_G$ where $\gamma_G \in[0,\sqrt{20}]$. The simulation was perfomed using NumPy \cite{harris2020array} and SciPy \cite{virtanen2020scipy}. The empirical means $\widehat{\theta}_1$ and $\widehat{\theta}_2$ are sampled from the distribution $\mathcal{N}(\theta_1,~1)$ and $\mathcal{N}(\theta_2,~1)$ respectively where the number of samples are assumed to be identical for simplicity. For local training we assume clients set their local models as their local empirical means which is analogous to clients performing a large number of local SGD steps to obtain the local minima of their empirical loss. For the global objective (standard FL, \incfl (ReLU), \incfl) a local minima is found using the  \texttt{scipy.optimize} function in the SciPy package. For each $\gamma_G^2 \in[0,\sqrt{20}]$, the average \text{GM-Appeal} is calculated over 10000 runs for each global objective.

\section{Experiment Details and Additional Results}
All experiments are conducted on clusters equipped with one NVIDIA TitanX GPU. The algorithms are implemented in PyTorch 1. 11. 0. All experiments are run with 3 different random seeds and the average performance with the standard deviation is shown. The code used for all experiments is included in the supplementary material.

\subsection{Experiment Details}\label{app:ed}


For FMNIST, for the results in \Cref{fig:dnn}, \Cref{tab:unseen}, and \Cref{tab:local}, the data is partitioned into 5 clusters where 2 labels are assigned for each cluster with no labels overlapping across clusters. For the other FMNIST results and EMNIST, we use the Dirichlet distribution~\cite{hsu2019noniid} to partition the data with $\alpha=0.5,~0.05$ respectively. Clients are randomly assigned to each cluster, and within each cluster, clients are homogeneously distributed with the assigned labels. For the Sent140 dataset, clients are naturally partitioned with their twitter IDs. The data of each client is partitioned to $60\%:40\%$ for training and test data ratio unless mentioned otherwise.

\paragraph{Obtaining $\widehat{\wb}_i,~i\in[M]$ for \incfl Results in \Cref{sec:exp}.} In \incfl, we use $\widehat{\wb}_i,~i\in[M]$ to calculate the aggregating weights (see \Cref{algo1}). For all experiments with \incfl, we obtain $\widehat{\wb}_i,~i\in[M]$ at each client by each client taking 100 local SGD steps on its local dataset with its own separate local model before starting federated training. We use the same batch-size and learning rate used for the local training at clients done after we start the federated training (line 8-9 in \Cref{algo1}). The specific values are mentioned in the next paragraph.

\paragraph{Local Training and Hyperparameters.} For all experiments, we do a grid search over the required hyperparameters to find the best performing ones. Specifically, we do a grid search over the learning rate: $\lrl\lrg\in\{0.1, 0.05, 0.01, 0.005, 0.001\}$, batchsize: $b\in\{32, 64, 128\}$, and local iterations: $\tau\in\{10, 30, 50\}$ to find the hyper-parameters with the highest test accuracy for each benchmark. For all benchmarks we use the best hyper-parameter for each benchmark after doing a grid search over feasible parameters referring to their source codes that are open-sourced. For a fair comparison across all benchmarks we do not use any learning rate decay or momentum.

\paragraph{DNN Experiments.} For FMNIST and EMNIST, we train a deep multi-layer perceptron network with 2 hidden layers of units $[64,30]$ with dropout after the first hidden layer where the input is the normalized flattened image and the output is consisted of 10 units each of one of the 0-9 labels. For Sent140, we train a deep multi-layer perceptron network with 3 hidden layers of units $[128,86,30]$ with pre-trained 200D average-pooled GloVe embedding \cite{pennington2014glove}. The input is the embedded 200D vector and the output is a binary classifier determining whether the tweet sentiment is positive or negative with labels 0 and 1 respectively. All clients have at least 50 data samples. To demonstrate further heterogeneity across clients' data we perform label flipping to $30\%$ of the clients that are uniformly sampled without replacement from the entire number of clients.

\subsection{Additional Experimental Results}\label{app:er}

\begin{figure*}[!t]
\centering
\begin{subfigure}{0.31\textwidth}
\centering
\includegraphics[width=1\textwidth]{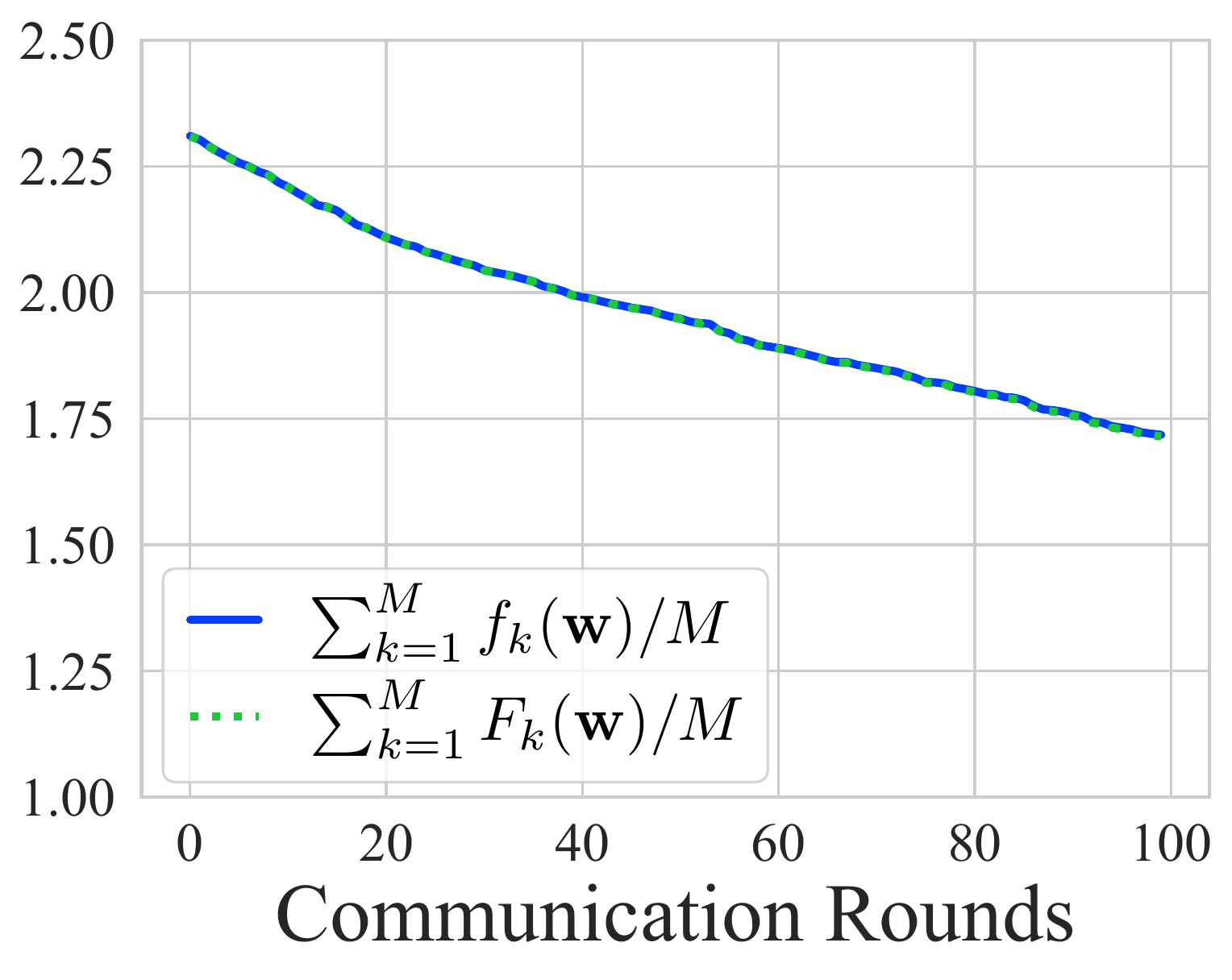} \caption{FMNIST} 
\end{subfigure} 
\begin{subfigure}{0.31\textwidth}
\centering
\includegraphics[width=1\textwidth]{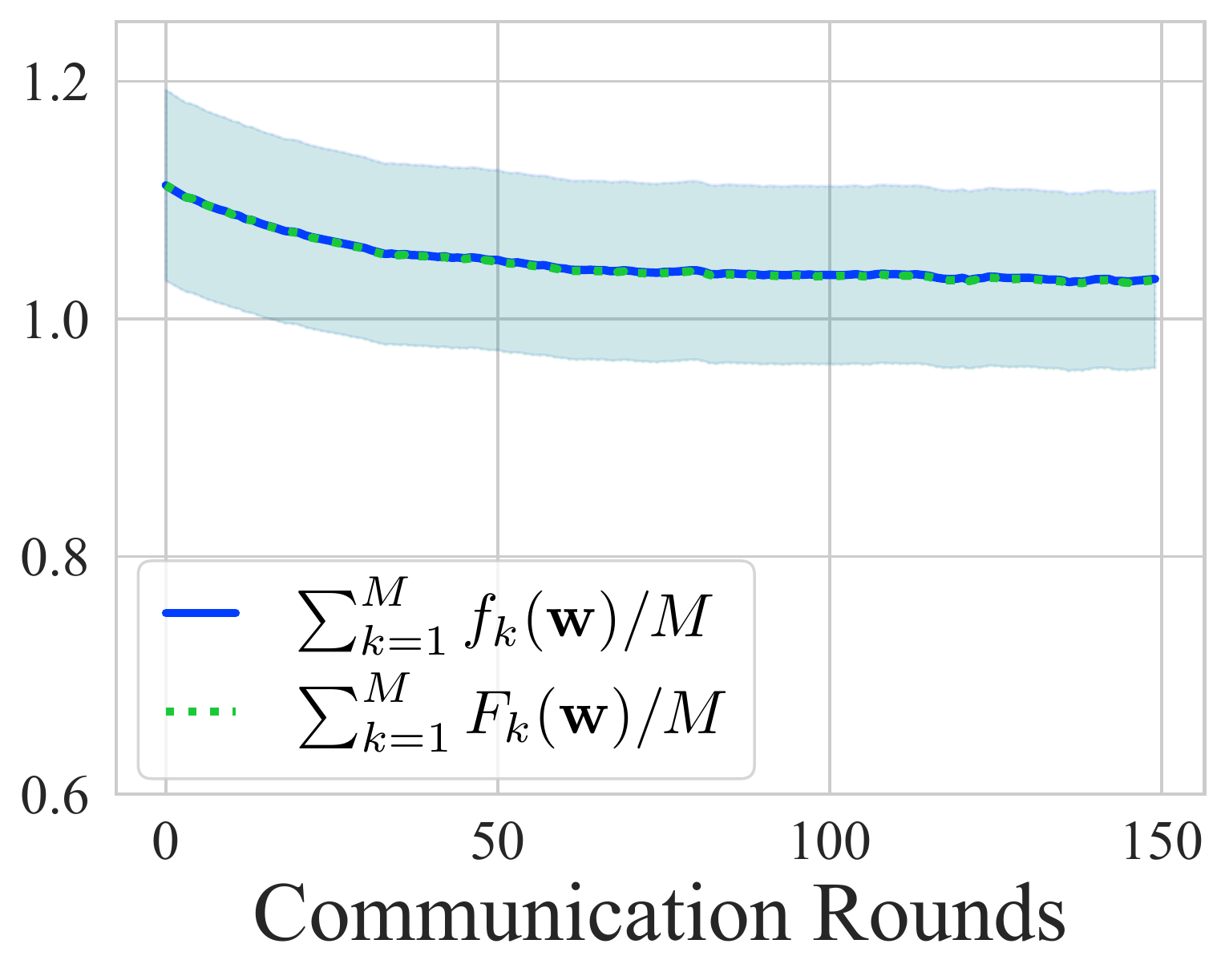} \caption{Sent140} 
\end{subfigure} 
 \caption{Comparison of the average of the true local losses across all clients ($\textstyle{\sum\nolimits_{k=1}^Mf_k(\mathbf{w})/M}$) and the empirical local losses across all clients ($\textstyle{\sum\nolimits_{k=1}^MF_k(\mathbf{w})/M}$) where the former is calculated on the test dataset and the latter is calculated on the training dataset for the global model $\wb$. We show that the average of the true local losses is nearly identical to the average empirical local loss across all clients empirically validating our relaxation of replacing $f_k(\wb)$ with $F_k(\wb)$.} \label{fig:abl} 
\end{figure*}

\paragraph{Ablation Study on $f_k(\wb)\approx F_k(\wb)$.} One of the two key relaxations we use for \incfl~(see \Cref{sec:proposed}) is that we replace $f_k(\wb)-f_k(\widehat{\wb}_k)$ with $F_k(\wb)-F_k(\widehat{\wb}_k)$. In other words, we replace the true loss $f_\ic(\wb)=\expt_{\xi\sim\mathcal{D}_\ic}[\ell(\wb,\xi)]$ with the empirical loss $F_\ic(\wb)=\frac{1}{|\bdat|}\sum_{\xi \in \bdat} \ell(\wb,\xi)$ for all clients $k\in[M]$. We have used the likely conjecture that the global model $\wb$ is trained on the data of all clients, making it unlikely to overfit to the local data of any particular client, leading to $f_k(\wb) \approx F_k(\wb)$. We show in \Cref{fig:abl} that this is indeed the case. For all DNN experiments, we show that the average true local loss across all clients, i.e., $\textstyle{\sum\nolimits_{k=1}^Mf_k(\mathbf{w})/M}$ is nearly identical to the average empirical local loss across all clients, i.e., $\textstyle{\sum\nolimits_{k=1}^MF_k(\mathbf{w})/M}$ given the training of the global model $\wb$ throughout the communication rounds. This empirically validates our relaxation of the true local losses to the empirical local losses.

\paragraph{Ablation Study on the Number of Local Steps $\tau_l$ to train $\widehat{\mathbf{w}}_k,k\in[M]$.} We conduct an additional ablation study where we vary the number of local steps to obtain $\widehat{\mathbf{w}}_k,k\in[M]$ for clients as shown in \Cref{tab:localiter}. Despite that a smaller number of local steps can lead to underfitting and a larger number of local steps can lead to overfitting, we show that all methods' \text{GM-Appeal}s do not vary much by the different number of local steps used for training.

\begin{table*}[!t]  \centering
\setlength\tabcolsep{4pt}
 \label{tab:localiter} 
\begin{tabular}{@{}lcccccc@{}} 
\toprule 
& \multicolumn{3}{c}{\vpr} & \multicolumn{3}{c}{Preferred-Model Test Acc.} \\ \cmidrule(lr){2-4} \cmidrule(lr){5-7}
& $\tau_l=50$ & $\tau_l=100$ & $\tau_l=150$ & $\tau_l=50$ & $\tau_l=100$ & $\tau_l=150$ \\ \hline
 FedAvg & $0.01~{\scriptstyle (\pm 0.01)}$ & $0.01~{\scriptstyle (\pm 0.0)}$ & $0.01~{\scriptstyle (\pm 0.0)}$ & $98.56~{\scriptstyle (\pm 0.08)}$ & $98.72~{\scriptstyle (\pm 1.02)}$ & $98.75~{\scriptstyle (\pm 1.28)}$ \\  \hline
   FedProx & $0.01~{\scriptstyle (\pm 0.0)}$ & $0.01~{\scriptstyle (\pm 0.0)}$ & $0.01~{\scriptstyle (\pm 0.01)}$ & $98.56~{\scriptstyle (\pm 1.15)}$ & $98.72~{\scriptstyle (\pm 1.08)}$ & $98.75~{\scriptstyle (\pm 1.05)}$ \\  \hline
\incfl & $\mathbf{0.55}~{\scriptstyle (\pm 0.0)}$  & $\mathbf{0.55}~{\scriptstyle (\pm 0.0)}$ & $\mathbf{0.55}~{\scriptstyle (\pm 0.0)}$ &  $\mathbf{98.64}~{\scriptstyle (\pm 1.03)}$ & $98.77~{\scriptstyle (\pm 1.01)}$ & $\mathbf{98.77}~{\scriptstyle (\pm 1.01)}$\\
 \bottomrule   
\end{tabular} 
\caption{\vpr and preferred-model test accuracy for \incfl and different baselines with FMNIST for varying number of local steps $\tau_l$ to obtain $\widehat{\wb}_k,k\in[M]$ where $T=200$ is the total number of communication rounds for training the global model.}
\end{table*}

\paragraph{Preferred-model Test Accuracy for the Local-Tuning Results in \Cref{tab:local}.}
In \Cref{tab:local}, we have shown how \incfl can largely increase the \vpr compared to the other baselines even when jointly used with local-tuning. In \Cref{tab:localPRA}, we show the corresponding preferred-model test accuracies. We show that for the seen clients that were active during training, \incfl~achieves at least the same or higher preferred-model test accuracy than the other methods for all the different datasets. Hence, the clients are able to also gain from \incfl~by achieving the highest accuracy in average with their preferred models (either global model or solo-trained local model). For the unseen clients with FMNIST, FedProx achieves a slightly higher preferred-model test accuracy ($+0.05$) than \incfl~but with a much lower \text{GM-Appeal} of 0.46 (see \Cref{tab:local}) as \incfl's \text{GM-Appeal} is 0.56. For the other datasets with unseen clients, \incfl~achieves at least the same or higher preferred-model test accuracy than the other methods. This demonstrates that \incfl~consistently largely improves the \text{GM-Appeal} compared to the other methods while losing very little, if any, in terms of the preferred-model test accuracy. \vspace{-1em}

\begin{table*}[!t] \centering
\setlength\tabcolsep{1.5pt}
\begin{tabular}{@{}lcccc@{}} 
\toprule 
& \multicolumn{2}{c}{Seen Clients} & \multicolumn{2}{c}{Unseen Clients} \\ \cmidrule(lr){2-3} \cmidrule(lr){4-5}
& {FMNIST} & {Sent140} & {FMNIST} &  {Sent140} \\ \hline
 FedAvg & $99.37~{\scriptstyle (\pm 0.24)}$ & $55.71~{\scriptstyle (\pm 0.46)}$ & $99.50~{\scriptstyle (\pm 0.02)}$ & $58.79~{\scriptstyle (\pm 0.67)}$ \\  \hline
   FedProx & $99.35~{\scriptstyle (\pm 0.23)}$  & $55.75~{\scriptstyle (\pm 0.80)}$ & $\mathbf{99.55}~{\scriptstyle (\pm 0.09)}$ & $58.82~{\scriptstyle (\pm 0.72)}$ \\  \hline
  PerFedAvg & $99.20~{\scriptstyle (\pm 0.25)}$  & $55.74~{\scriptstyle (\pm 0.80)}$ & $98.98~{\scriptstyle (\pm 0.55)}$ & $58.82~{\scriptstyle (\pm 0.72)}$\\  \hline
  MW-Fed & $99.27~{\scriptstyle (\pm 0.39)}$ & $55.06~{\scriptstyle (\pm 0.38)}$ & $99.47~{\scriptstyle (\pm 0.08)}$ & $57.36~{\scriptstyle (\pm 0.71)}$\\  \hline
\incfl & $\mathbf{99.40}~{\scriptstyle (\pm 0.30)}$ & $\mathbf{55.82}~{\scriptstyle (\pm 0.82)}$ &  $99.50~{\scriptstyle (\pm 0.02)}$ & $\mathbf{58.88}~{\scriptstyle (\pm 0.77)}$\\
 \bottomrule
\end{tabular}\caption{Preferred-model test accuracy with the locally-tuned models with 5 local steps from the final global model for seen clients' and unseen clients' test data (the corresponding \text{GM-Appeal} is in \Cref{tab:local}).} \label{tab:localPRA}  
\end{table*}

\begin{table*}[!t]  \centering
\setlength\tabcolsep{1.5pt}

\begin{tabular}{@{}lccccc@{}} 
\toprule 
& \multicolumn{2}{c}{\vpr} & \multicolumn{2}{c}{Preferred-Model Test Acc.} \\ \cmidrule(lr){2-3} \cmidrule(lr){4-5}
& {FMNIST} & {Sent140} & {FMNIST} & {Sent140} \\ \hline
q-FFL ($q=1$) & $0.03~{\scriptstyle (\pm 0.01)}$ & $0.09~{\scriptstyle (\pm 0.06)}$ & $99.24~{\scriptstyle (\pm 0.05)}$ &  $53.10~{\scriptstyle (\pm 2.63)}$ \\  \hline
q-FFL ($q=10$) & $0.0~{\scriptstyle (\pm 0.0)}$ & $0.09~{\scriptstyle (\pm 0.0)}$ & $98.90~{\scriptstyle (\pm 0.01)}$  & $52.71~{\scriptstyle (\pm 1.40)}$ \\  \hline
\incfl & $\mathbf{0.55}~{\scriptstyle (\pm 0.0)}$   & $\mathbf{0.41}~{\scriptstyle (\pm 0.07)}$ &  $\mathbf{99.29}~{\scriptstyle (\pm 0.03)}$ & $\mathbf{53.93}~{\scriptstyle (\pm 1.87)}$\\
 \bottomrule   
\end{tabular} 
\caption{\vpr and preferred-model test accuracy for the seen clients' test data with the final global models trained via \incfl and q-FFL~\cite{li2019fair} which aims in improving fairness. The baseline q-FFL with large $q$, e.g. $q=10$, emulates the behavior of another well-known algorithm for improving fairness named AFL~\cite{mohri2019agnostic}.  \label{tab:qfflseen} }
\end{table*}
\paragraph{Comparison with Algorithms for Fairness} Fair FL methods~\cite{li2019fair,mohri2019agnostic} aim in training a global model that yields small variance across the clients' test accuracies. These methods may satisfy the worst performing clients, but potentially at the cost of causing dissatisfaction from the best performing clients. We show in \Cref{tab:qfflseen} that the common fair FL methods are indeed not effective in improving the overall clients' \vpr. We see that the fair FL methods achieve a \vpr lower than 0.01 for all datasets while \incfl achieves at least 0.40 for all datasets. Moreover, the preferred-model test accuracy is also higher for \incfl compared to the fair FL methods. This underwelming performance of fair FL methods in \vpr can be due to the fact that fair FL methods try to find the global model that performs well, in overall, over \textit{all} clients which results in failing to satisfy \textit{any} client. 

\paragraph{Ablation Study on $f_k(\wb)\approx F_k(\wb)$.} One of the two key relaxations we use for \incfl~(see \Cref{sec:proposed}) is that we replace $f_k(\wb)-f_k(\widehat{\wb}_k)$ with $F_k(\wb)-F_k(\widehat{\wb}_k)$. In other words, we replace the true loss $f_\ic(\wb)=\expt_{\xi\sim\mathcal{D}_\ic}[\ell(\wb,\xi)]$ with the empirical loss $F_\ic(\wb)=\frac{1}{|\bdat|}\sum_{\xi \in \bdat} \ell(\wb,\xi)$ for all clients $k\in[M]$. We have used the likely conjecture that the global model $\wb$ is trained on the data of all clients, making it unlikely to overfit to the local data of any particular client, leading to $f_k(\wb) \approx F_k(\wb)$. We show in \Cref{fig:abl} that this is indeed the case. For all DNN experiments, we show that the average true local loss across all clients, i.e., $\textstyle{\sum\nolimits_{k=1}^Mf_k(\mathbf{w})/M}$ is nearly identical to the average empirical local loss across all clients, i.e., $\textstyle{\sum\nolimits_{k=1}^MF_k(\mathbf{w})/M}$ given the training of the global model $\wb$ throughout the communication rounds. This empirically validates our relaxation of the true local losses to the empirical local losses.

\begin{wrapfigure}{r}{0.25\textwidth} 
\centering \vspace{-2em}
    \includegraphics[width=1\textwidth]{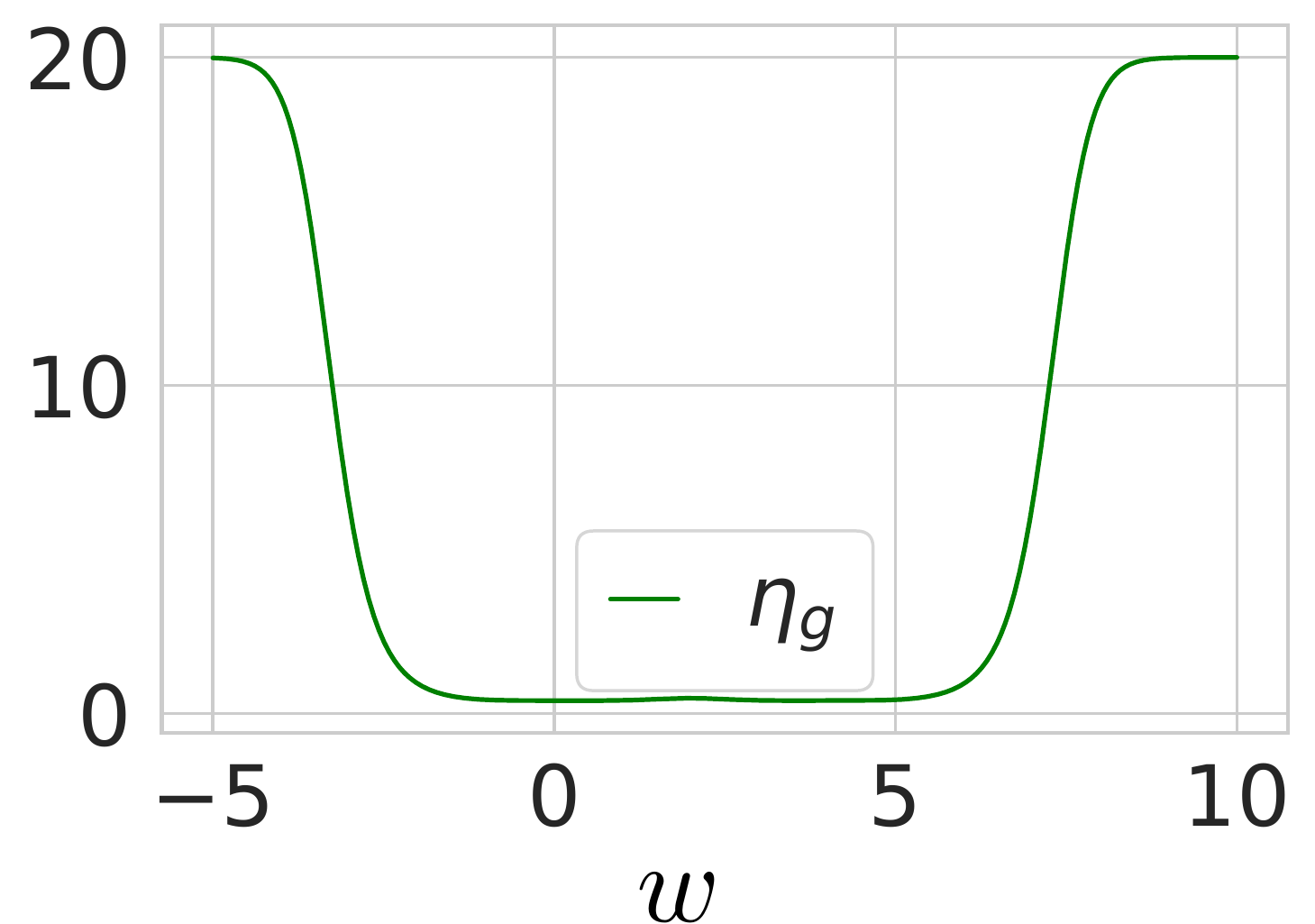}
    \caption{Behavior of Theoretical Learning Rate of \incfl~for the mean estimation example in Fig. 1(b). As expected from the theoretical learning rate formula, we see a higher learning rate in regions where the function is flat.}
    \label{fig:lrtheo} 
\end{wrapfigure}
\textbf{\incfl's Theoretical Learning Rate Behavior for Fig. 1 (b)}. 
Here, we provide a plot of \incfl's theoretical learning rate for the mean estimation example in Fig. 1(b) in \Cref{fig:lrtheo} to show how the learning rate changes for different regions of the model. We show this plot as a proof of concept on the adaptive learning rate we discuss in Section 4. For the sigmoid function which is used for our \incfl~objective, using a global notion of smoothness can cause gradient descent to be too slow since global smoothness is determined by behavior at $w=0$ where $w$ is the model. In this case, it is better to use a local estimate of smoothness in the flat regions where $|w|>>0$. Recall that $\nabla^2 \sigma (w) = \sigma(w)(1-\sigma(w))(1-2\sigma(w)) < \sigma(x)(1-\sigma(w))$ and therefore setting the learning rate proportional to $\frac{1}{\sigma(w)(1-\sigma(w))}$ can increase the learning rate in flat regions where $\sigma(w)$ is close to 1 or 0. Following a similar argument, we can show that the learning rate in our objective should be proportional to $1/\left(\sum_{i=1}^M \sigma(F_i(w)-F_i(\hat{w}^*) (1- \sigma(F_i(w)-F_i(\hat{w}^*))\right)$.   

\end{document}